\pdfoutput=1
\documentclass[11pt]{article}
\usepackage[a4paper,margin=1in]{geometry}
\usepackage[T1]{fontenc}
\usepackage[utf8]{inputenc}
\usepackage{lmodern,microtype}
\usepackage{amsmath,amssymb,amsthm,mathtools,bm}
\usepackage{booktabs,tabularx,array,longtable}
\usepackage{enumitem,graphicx,pdflscape,float}
\usepackage{multirow}
\usepackage[numbers,sort&compress]{natbib}
\usepackage[table]{xcolor}
\usepackage[normalem]{ulem}
\usepackage{tikz,xspace,hyperref}
\usepackage{pgfplots}\pgfplotsset{compat=1.18}
\usepgfplotslibrary{groupplots}
\usepgfplotslibrary{fillbetween}
\definecolor{AGIdleFill}{HTML}{E0EBF7}
\definecolor{AGIdleEdge}{HTML}{4472C4}
\definecolor{AGPickFill}{HTML}{CFE8D0}
\definecolor{AGPickEdge}{HTML}{3D7A44}
\definecolor{AGHandoffFill}{HTML}{DCD6F2}
\definecolor{AGHandoffEdge}{HTML}{5A4A9C}
\definecolor{AGAlignFill}{HTML}{FFC882}
\definecolor{AGAlignEdge}{HTML}{A06A20}
\definecolor{AGAssembleFill}{HTML}{FF8C6E}
\definecolor{AGAssembleEdge}{HTML}{A03A20}
\definecolor{AGDeliverFill}{HTML}{7FD88F}
\definecolor{AGDeliverEdge}{HTML}{1E6B2C}
\definecolor{AGShelfFill}{HTML}{FFDC96}
\definecolor{AGSubassembly}{HTML}{1E8F89}
\definecolor{AGRaw}{HTML}{C9962A}
\definecolor{AGRawAlt}{HTML}{8C5AC8}
\definecolor{AGSelect}{HTML}{D81B60}
\definecolor{AGDark}{HTML}{555555}
\definecolor{AGOtherFill}{HTML}{F2E2B8}
\definecolor{AGOtherEdge}{HTML}{8A6D1F}
\definecolor{AGFinal}{HTML}{268934}
\definecolor{vlabelblue}{HTML}{0D47A1}
\definecolor{vlabelred}{HTML}{B71C1C}
\definecolor{vlabelgreen}{HTML}{1B5E20}
\definecolor{vlabelviolet}{HTML}{4A148C}
\definecolor{vlabelteal}{HTML}{00695C}
\definecolor{vlabelorange}{HTML}{E65100}
\definecolor{vlabelgray}{HTML}{424242}

\definecolor{AGTableGray}{HTML}{E1E1E1}  

\definecolor{AGQMIX}{HTML}{F9DEAC}      
\definecolor{AGIPPO}{HTML}{F99B1C}      
\definecolor{AGMAPPO}{HTML}{7A5100}     
\definecolor{AGGreedy}{HTML}{C8C8C8}    
\definecolor{AGParallel}{HTML}{969696}  
\definecolor{AGDarkGray}{HTML}{4B4B4B}

\colorlet{ovStruct}{AGParallel}      
\colorlet{ovStructLight}{AGGreedy}   
\colorlet{ovGeom}{AGMAPPO}           
\colorlet{ovInput}{AGIPPO}           
\colorlet{ovSoft}{AGQMIX}            
\colorlet{ovTopology}{AGDarkGray}

\definecolor{AGProdInFill}{HTML}{F9DEAC}
\definecolor{AGProdInDraw}{HTML}{F9BC64}
\definecolor{AGProdOutFill}{HTML}{B87114}
\definecolor{AGProdOutDraw}{HTML}{7A5100}
\definecolor{AGCellFill}{HTML}{FAFAFA}
\definecolor{AGCellDraw}{HTML}{E1E1E1}

\pgfplotsset{
  ag/bar/qmix/.style={draw=AGQMIX, fill=AGQMIX!65},
  ag/bar/ippo/.style={draw=AGIPPO, fill=AGIPPO!65},
  ag/bar/mappo/.style={draw=AGMAPPO, fill=AGMAPPO!65},
  ag/bar/greedy/.style={draw=AGGreedy, fill=AGGreedy!65},
  ag/bar/parallel/.style={draw=AGParallel, fill=AGParallel!65}
}

\pgfplotsset{
  ag/qmix/.style={color=AGQMIX},
  ag/ippo/.style={color=AGIPPO},
  ag/mappo/.style={color=AGMAPPO},
  ag/greedy/.style={color=AGGreedy},
  ag/parallel/.style={color=AGParallel}
}

\hypersetup{
  hidelinks,
  pdftitle={AssemblyGrid v1: A Benchmark for Multi-Robot Production with Temporary Coalitions, Local Information, and Geometric Constraints},
  pdfauthor={Fouad Bahrpeyma; David Heik; Dirk Reichelt}
}
\usetikzlibrary{positioning,arrows.meta,calc,fit,backgrounds,shapes.geometric}
\usetikzlibrary{decorations.pathreplacing,patterns}
\tikzset{
  agbox/.style={rectangle, rounded corners=2pt, draw=black!55, fill=black!6,
                align=center, inner sep=3pt, font=\footnotesize},
  agbad/.style={rectangle, rounded corners=2pt, draw=red!65!black, fill=red!7,
                align=center, inner sep=3pt, font=\footnotesize},
  aggood/.style={rectangle, rounded corners=2pt, draw=teal!70!black, fill=teal!9,
                align=center, inner sep=3pt, font=\footnotesize},
  agnew/.style={rectangle, rounded corners=2pt, draw=violet!65!black, fill=violet!8,
                align=center, inner sep=3pt, font=\footnotesize},
  agarrow/.style={-{Latex[length=2.2mm]}, draw=black!65, thick},
  agthin/.style={-{Latex[length=1.8mm]}, draw=black!45},
  agtie/.style={draw=black!35, dashed, thin},
  aghead/.style={font=\scriptsize\bfseries, text=black!70},
  agnote/.style={font=\scriptsize, text=black!70, align=left},
  agbar/.style={draw=none, fill=teal!55!black},
  agbarbad/.style={draw=none, fill=red!55!black},
  agbarref/.style={draw=none, fill=black!35},
}
\newcolumntype{Y}{>{\raggedright\arraybackslash}X}
\newcolumntype{C}{>{\centering\arraybackslash}X}
\newcommand{\AG}{\textsc{AssemblyGrid}\ v1\xspace}
\newcommand{\Robots}{\mathcal{U}}
\title{AssemblyGrid v1: A Benchmark for Multi-Robot Production with Temporary Coalitions, Local Information, and Geometric Constraints}
\author{%
Fouad Bahrpeyma \quad David Heik \quad Dirk Reichelt\\[0.55em]
\small Faculty of Computer Science/Mathematics\\
\small Hochschule f\"ur Technik und Wirtschaft Dresden -- University of Applied Sciences\\
\small Dresden, Germany\\[0.35em]
\scriptsize \href{mailto:Bahrpeyma@ieee.org}{Bahrpeyma@ieee.org} \quad
\href{mailto:david.heik@htw-dresden.de}{david.heik@htw-dresden.de} \quad
\href{mailto:dirk.reichelt@htw-dresden.de}{dirk.reichelt@htw-dresden.de}
}
\date{}
\begin{document}
\maketitle

\begin{abstract}
Flexible robotic production requires decisions about process progression, material routing, resource assignment, temporary cooperation, and simultaneous execution to be made jointly, because each decision can affect the feasibility of the others. This becomes especially challenging under decentralized control, where each robot acts from bounded local information while overall progress depends on collective decisions, shared resources, material state, and workspace compatibility. Such settings are relevant not only to manufacturing control, but also to cooperative multi-agent decision making under partial observability, local coordination, temporary cooperation, and resource contention. These properties closely match the class of problems addressed by MARL, where decentralized agents must learn from interaction how locally informed decisions contribute to long-term production progress under interdependent actions and an evolving system state.
This paper introduces \textsc{AssemblyGrid} v1, a reproducible benchmark for repeated multi-robot production that combines explicit process progression, decentralized observations, material transfer, temporary multi-robot coalitions, productive concurrency, and geometry-dependent feasibility within one task-level formulation. The benchmark suite comprises \textsc{Flow}, \textsc{Coalition}, and \textsc{Concurrency} workload families, each with three scenario levels. Task success and evaluation measures are defined independently of learning reward and solution method, allowing MARL and other learning-based and non-learning methods to address the same production problem.

\textsc{AssemblyGrid} v1 is evaluated through executable conformance checks, dedicated mechanism studies, and algorithmic evaluations. The algorithmic evaluation includes a privileged centralized reference, structured decentralized controllers, and MARL methods including IPPO, MAPPO, and QMIX. Experimental results confirm the existence of feasible centralized and decentralized control solutions, while the MARL results show that decentralized policies can learn effective production behavior from local observations and actions.

\end{abstract}
\section{Introduction}

Modern production systems are moving away from fixed production structures toward configurations in which capacity, routing, and resource assignments can be adapted to changing production requirements. Reconfigurable and flexible manufacturing systems enable such adaptation through modular, scalable, and convertible system designs \citep{koren1999rms,elmaraghy2005flex}. Matrix-structured manufacturing extends this flexibility by introducing redundant workstations and alternative routing possibilities \citep{schoenemann2015matrix,wang2025matrix}, while line-less assembly and swarm-production concepts further relax fixed material paths and resource assignments \citep{huttemann2019lineless,schmitt2021lineless,avhad2023swarm,avhad2024swarm}. As these alternatives increase, the central challenge shifts from providing flexibility to controlling it: production decisions become increasingly state dependent and interdependent, since the choice of what to execute, where material should move, which resources should participate, and what can proceed concurrently must be resolved together. 
In a flexible robotic workcell, this control problem involves four closely related decisions: which product operation to execute, where its material should be routed, which resources should perform the operation, and whether the resulting assignment is geometrically feasible. These decisions are inherently coupled. The selected operation determines the required material and process state, resource assignment affects which activities can proceed concurrently, and workspace geometry may render an otherwise valid process and resource assignment infeasible. Similar dependencies among routing, resource allocation, look-ahead, layout, and material supply have been identified in matrix-production control and integrated production-system planning \citep{nielsen2023matrixcontrol,li2026matrixreview,schumacher2026matrixplanning}.

These dependencies make task-level controller design a central challenge, since task progress, resource use, concurrency, and geometric feasibility must be considered jointly. In this paper, a \emph{controller} refers to any method that selects task-level actions for the robots, including learned policies, planners, schedulers, optimization methods, and heuristics. Benchmarks can support the development and evaluation of such controllers by defining the underlying decision problem reproducibly without prescribing how it should be solved. Multi-agent reinforcement learning (MARL) is particularly relevant in this setting because multiple interacting agents must learn from experience how locally informed decisions contribute to the common production objective. In the cooperative setting considered here, all agents contribute to this shared production task. This task-level cooperation is distinct from a \emph{coalition}, which here denotes a temporary subset of robots that jointly executes a particular operation and dissolves when that operation ends.

The Multi-Agent Particle Environment (MPE) \citep{lowe2017multiagent} provides compact continuous multi-agent coordination tasks and became a widely used testbed for decentralized policies trained with centralized information. The StarCraft Multi-Agent Challenge (SMAC) \citep{samvelyan2019starcraft} became a standard cooperative benchmark for centralized training with decentralized execution, and SMACv2 later extended this benchmark with procedural generation and stronger partial observability to reduce dependence on fixed scenario regularities \citep{ellis2023smacv2}. Hanabi \citep{bard2020hanabi} represents cooperative inference under imperfect information and conventions; Google Research Football \citep{kurach2020football} provides a physical team-game environment; and MAgent \citep{zheng2018magent} addresses large-population multiagent interaction. Closer to logistics and structured task execution, Level-Based Foraging (LBF) \citep{christianos2020shared} includes tasks requiring simultaneous participation by several agents, whereas Multi-Robot Warehouse (RWARE) \citep{papoudakis2021benchmarking} combines local observations, delivery objectives and congestion. Flatland \citep{mohanty2020flatland} represents railway routing under operational conflicts; Overcooked-AI \citep{carroll2019overcooked}, together with later variants by Ruhdorfer et al. \citep{ruhdorfer2024ogc} and Gessler et al. \citep{gessler2025overcookedv2}, combines sequential subtasks, blocking and division of labour; and Melting Pot \citep{leibo2021meltingpot} evaluates behavior across changing social configurations.

A complementary body of infrastructure research addresses reproducibility and implementation. PettingZoo standardizes multiagent environment interfaces \citep{terry2021pettingzoo}; the Vectorized Multi-Agent Simulator (VMAS) \citep{bettini2022vmas} and Gigastep \citep{lechner2023gigastep} provide vectorized physical and high-throughput environments; and JaxMARL \citep{rutherford2024jaxmarl}, MARLlib \citep{hu2023marllib}, and BenchMARL \citep{bettini2024benchmarl} provide reusable evaluation infrastructure across algorithms and environments. Together, these benchmarks and frameworks cover partial observability, coordination, routing, congestion, conventions, division of labour and large-scale interaction. Repeated typed material transformation, temporary operation-specific robot coalitions, bounded decentralized execution and geometry-dependent admissibility are typically distributed across separate benchmark traditions instead of being combined within one production problem definition.

The remaining mechanisms are developed more directly in neighboring research areas. Multi-robot task allocation was formalized early in \citep{gerkey2004mrta} and later synthesized in \citep{nunes2017taxonomy}, while market-based and other decentralized mechanisms provide established solution families \citep{dias2006market,chakraa2023mrta,alirezazadeh2024survey}. Coalition-formation methods, such as those studied in \citep{vig2006coalition} and, more recently in \citep{arjun2025coalition} and \citep{verma2025cfhmrta}, extend this setting to tasks for which the participating robot set is itself part of the decision problem. Manufacturing MARL represents production processes through scheduling, transport and resource-allocation decisions, as surveyed by Bahrpeyma et al. \citep{bahrpeyma2022smartfactory} and Xu et al. \citep{xu2025fsspsurvey}, including distributed factory scheduling \citep{zhou2021scheduling}, joint layout and scheduling \citep{kaven2024layout}, dynamic flexible job-shop scheduling \citep{burggraef2022dfjsp,zhang2023deepmag,zhang2024dfjsp}, factory-scale control \citep{jang2025factorywide}, size generalization \citep{bhatta2025trainsmall}, breakdown response \citep{lv2025breakdown}, and knowledge-guided control \citep{qin2025kgmarl}.

Collaborative assembly and task-and-motion planning address another part of the problem by representing workspace interference, synchronization, shared manipulation and collision-aware motion. Representative studies include collaborative assembly and synchronization work by Boschetti et al. \citep{boschetti2021collab}, human-robot collaboration analysis by Keshvarparast et al. \citep{keshvarparast2024cobot}, collaborative manipulation by Feng et al. \citep{feng2020collabmanip}, assembly-oriented coordination by Marvel et al. \citep{marvel2018assembly}, assembly-task reviews by Jiang et al. \citep{jiang2022assemblyreview}, and task-and-motion planning formulations by Chen et al. \citep{chen2022tamp}. Recent embodied systems such as RAMP \citep{collins2024ramp}, RoboFactory \citep{qin2025robofactory}, the large-scale collaborative setting of Brown et al. \citep{brown2025largescale}, APEX-MR (multi-robot asynchronous planning and execution) \citep{huang2025apexmr}, and Fabrica \citep{tian2025fabrica} further incorporate compositional assembly constraints, collaborative transport, and asynchronous or contact-rich execution. These studies address physical and cooperative execution at a substantially higher level of physical detail, although evaluation generally centers on individual assembly tasks or manipulation episodes instead of continuing production with repeated products. Recent reviews of MARL for industrial cooperative manipulation, such as Huertos et al. \citep{huertos2026industrialmarl}, also identify standardized industrial benchmarks and reproducibility as continuing research needs.

The gap addressed here is the integration of mechanisms that are typically represented in separate problem formulations or at different abstraction levels. Scheduling models repeated process progression and production objectives; task-allocation research represents assignment and coalition formation; collaborative assembly represents geometric and physical feasibility; and cooperative MARL provides established formulations for decentralized decision making under partial observability. What remains less commonly represented is their combination within a single benchmark definition in which typed material progression, temporary local coalitions, productive concurrency and geometric admissibility are simultaneously part of the decision problem, with task outcomes defined independently of the controller and its learning reward, as summarized in Table~\ref{tab:litfamilies}.

\begin{table*}[t]
\centering
\caption{Cross-literature mechanism matrix.}
\label{tab:litfamilies}
\scriptsize
\renewcommand{\arraystretch}{1.10}
\setlength{\tabcolsep}{3.0pt}
\begin{tabularx}{\textwidth}{>{\raggedright\arraybackslash}p{0.22\textwidth}CCCCCC}
\toprule
Family & Repeated process / material state & Temporary multi-robot coalition & Bounded local decentralized execution & Task and motion / geometry coupling & Production-rate outcome & Reusable benchmark \\
\midrule
Generic cooperative MARL (MPE, SMAC/SMACv2, Hanabi) \citep{lowe2017multiagent,samvelyan2019starcraft,ellis2023smacv2,bard2020hanabi} & -- & $\circ$ & $\checkmark$ & $\circ$ & -- & $\checkmark$ \\
Logistics and routing MARL (LBF, RWARE, Flatland) \citep{christianos2020shared,papoudakis2021benchmarking,mohanty2020flatland} & $\circ$ & $\circ$ & $\checkmark$ & $\circ$ & $\circ$ & $\checkmark$ \\
Overcooked family \citep{carroll2019overcooked,ruhdorfer2024ogc,gessler2025overcookedv2} & $\circ$ & -- & $\checkmark$ & $\circ$ & $\circ$ & $\checkmark$ \\
Multi-robot task allocation / coalition formation \citep{gerkey2004mrta,nunes2017taxonomy,vig2006coalition} & $\circ$ & $\checkmark$ & $\circ$ & $\circ$ & $\circ$ & $\circ$ \\
Reconfigurable / matrix / line-less / swarm production \citep{koren1999rms,schoenemann2015matrix,huttemann2019lineless,avhad2023swarm} & $\checkmark$ & $\circ$ & $\circ$ & $\circ$ & $\checkmark$ & $\circ$ \\
Manufacturing MARL / scheduling \citep{bahrpeyma2022smartfactory,zhou2021scheduling,kaven2024layout} & $\checkmark$ & $\circ$ & $\checkmark$ & $\circ$ & $\checkmark$ & $\circ$ \\
Task and motion planning / collaborative assembly \citep{chen2022tamp,boschetti2021collab,jiang2022assemblyreview} & $\circ$ & $\checkmark$ & $\circ$ & $\checkmark$ & $\circ$ & $\circ$ \\
RAMP / RoboFactory / Fabrica \citep{collins2024ramp,qin2025robofactory,tian2025fabrica} & $\circ$ & $\circ$ & $\circ$ & $\checkmark$ & -- & $\checkmark$ \\
\rowcolor{AGTableGray!45}\AG & $\checkmark$ & $\checkmark$ & $\checkmark$ & $\circ$ (AssemblyGrid v1 abstract proxy) & $\checkmark$ & $\checkmark$ \\
\bottomrule
\end{tabularx}
\vspace{-1mm}
\raggedright\footnotesize $\checkmark$ primary mechanism\par
\vspace{0mm}
\raggedright\footnotesize $\circ$ present in some members or at a coarser abstraction\par
\vspace{-1mm}
\raggedright\footnotesize -- generally outside the modeled object\par
\end{table*}

Table~\ref{tab:litfamilies} summarizes how the relevant mechanisms are distributed across neighboring benchmark and problem families, and positions AssemblyGrid v1 relative to them. 
Hence, AssemblyGrid v1 is organized around six design principles: (1) some operations involve cooperation among multiple robots; (2) products progress through explicit recipe-defined production operations, thereby keeping material and process state explicit; (3) the robot group required for one operation should be temporary and operation-specific instead of fixed across the production process; (4) information locality and manipulation locality should be represented separately; (5) geometric admissibility should constrain otherwise valid task and resource choices; and (6) benchmark outcomes should be defined independently of controller-specific objectives, thereby allowing learning, planning, scheduling, optimization and heuristic methods to be studied on the same production problem.
AssemblyGrid v1 combines these elements in a controller-independent benchmark for continuing multi-robot production under decentralized local information, with explicit recipe and material progression, temporary multi-robot cooperation, and task-level geometric constraints. Its official suite organizes workload variation into \textsc{Flow}, \textsc{Coalition}, and \textsc{Concurrency}, each evaluated under a common benchmark definition. The paper contributes this production task and scenario suite together with a reproducible evaluation framework that uses conformance checks, mechanism-level validation, and reference controllers to demonstrate benchmark executability.
The remainder of the paper follows the benchmark from definition to validation and scope. Section~\ref{sec:environment} defines the environment and decentralized decision problem, Section~\ref{sec:benchmarkdef} specifies benchmark instances, workload families, evaluation measures and benchmark requirements, and Section~\ref{sec:evalresults} presents the experimental validation and analysis. Section~\ref{sec:discussion} then discusses scope, limitations and extensions, and Appendix~\ref{app:formal-spec} provides the complete formal benchmark definition.

\section{The AssemblyGrid v1 environment}
\label{sec:environment}
\begin{figure}[H]
\centering
\begin{minipage}[b]{0.49\textwidth}
\centering
\includegraphics[width=\textwidth]{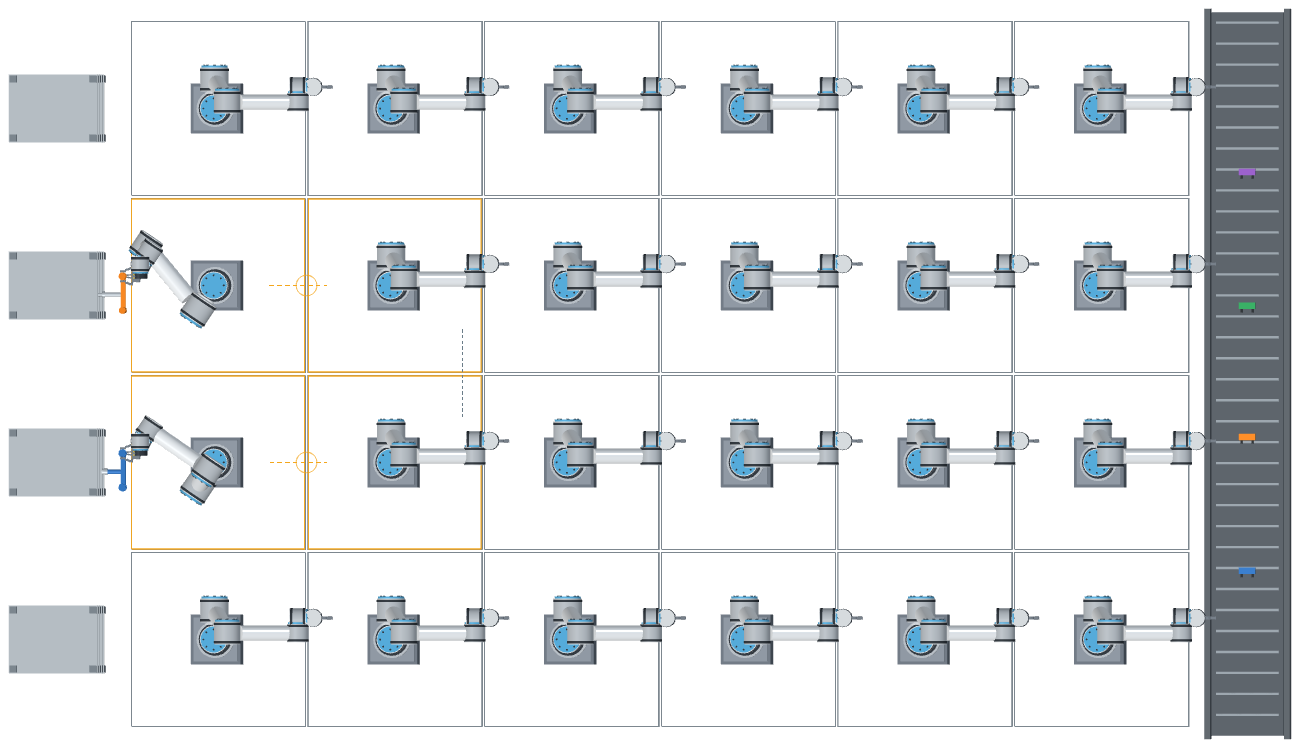}\\[1.5mm]
{\footnotesize (a) top view}
\end{minipage}\hfill
\begin{minipage}[b]{0.49\textwidth}
\centering
\includegraphics[width=\textwidth]{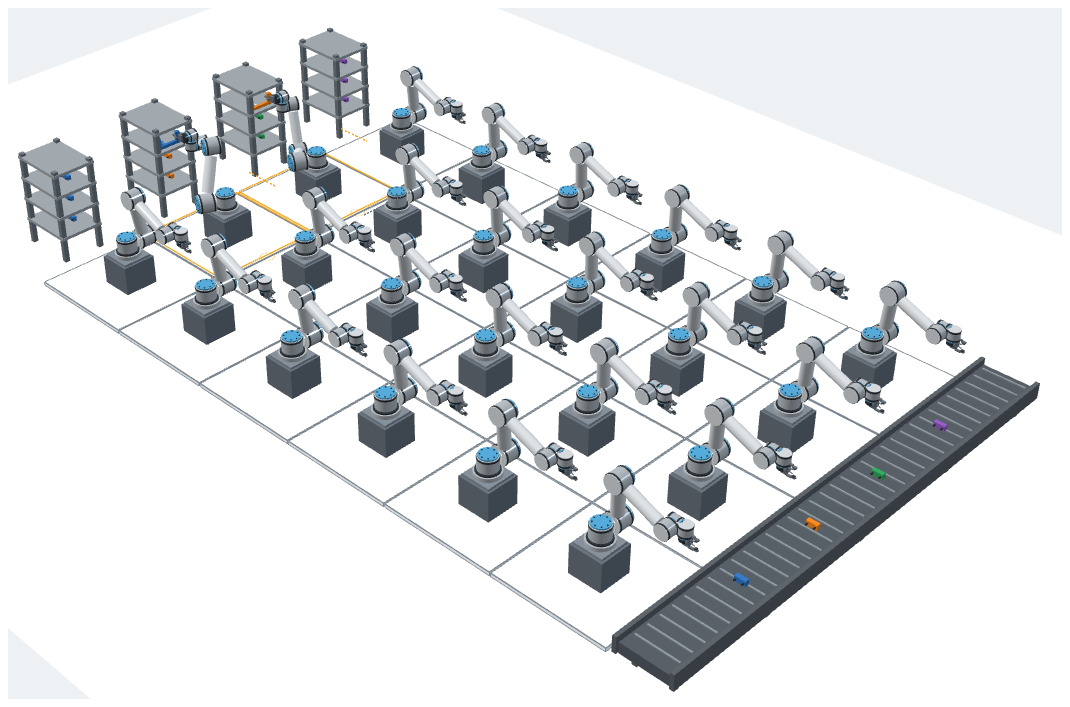}\\[1.5mm]
{\footnotesize (b) isometric view}
\end{minipage}
\caption{Architectural illustration of the AssemblyGrid v1 task-level production cell (illustrative, while geometry remains abstract).}
\label{fig:grid_architecture}
\end{figure}

Figure~\ref{fig:grid_architecture} illustrates the AssemblyGrid v1 production cell, where fixed-base manipulators occupy a regular lattice, raw materials enter from the left, products progress through recipe-defined operations, and completed products leave on the right. Overlapping workspaces determine where robots can interact, form temporary coalitions, and interfere with concurrent activities. Each robot is paired one-to-one with a task-level agent, and thus the benchmark connects physical interaction and material flow directly to recipe requirements, local actions, coalition formation, and decentralized decision making.

\subsection{Physical organization and local interaction}

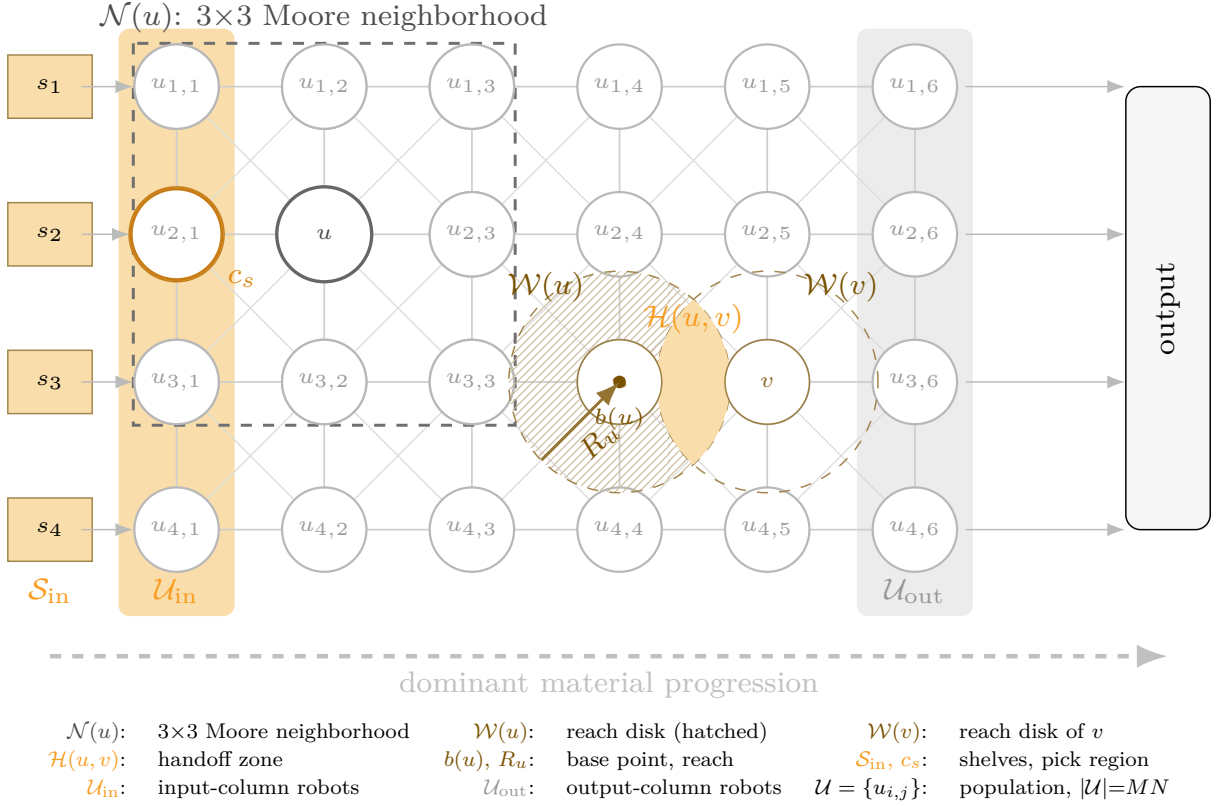
\begin{figure}[H]
\centering
\resizebox{\linewidth}{!}{%
\begin{tikzpicture}[>=Latex, scale=1.0]

  \def\sp{1.4}

  \def\yPortLabels{-6.18}
  \def\yBarBottom{-6.42}
  \def\yPopulation{-6.5}
  \def\yFlow{-6.80}

  \colorlet{ovGridMain}{ovStruct!48}
  \colorlet{ovGridDiag}{ovStruct!34}
  \colorlet{ovNodeStroke}{ovStruct!68}
  \colorlet{ovNodeText}{ovStruct!94}
  \colorlet{ovFocusStroke}{ovTopology!85}
  \colorlet{ovFocusText}{ovTopology}

  \fill[ovSoft, rounded corners=3pt]
    ({\sp*1-0.55}, {-\sp*1+0.55})
    rectangle
    ({\sp*1+0.55}, {\yBarBottom});

  \fill[ovStructLight!35, rounded corners=3pt]
    ({\sp*6-0.55}, {-\sp*1+0.55})
    rectangle
    ({\sp*6+0.55}, {\yBarBottom});

  \foreach \i in {1,...,4}{%
    \foreach \j in {1,...,5}{%
      \pgfmathtruncatemacro{\jj}{\j+1}%
      \draw[ovGridMain, line width=0.45pt]
        ({\sp*\j},{-\sp*\i}) --
        ({\sp*\jj},{-\sp*\i});%
    }%
  }%

  \foreach \i in {1,...,3}{%
    \pgfmathtruncatemacro{\ii}{\i+1}%
    \foreach \j in {1,...,6}{%
      \draw[ovGridMain, line width=0.45pt]
        ({\sp*\j},{-\sp*\i}) --
        ({\sp*\j},{-\sp*\ii});%
    }%
  }%

  \foreach \i in {1,...,3}{%
    \foreach \j in {1,...,5}{%
      \pgfmathtruncatemacro{\ii}{\i+1}%
      \pgfmathtruncatemacro{\jj}{\j+1}%

      \draw[ovGridDiag, line width=0.35pt]
        ({\sp*\j},{-\sp*\i}) --
        ({\sp*\jj},{-\sp*\ii});%

      \draw[ovGridDiag, line width=0.35pt]
        ({\sp*\jj},{-\sp*\i}) --
        ({\sp*\j},{-\sp*\ii});%
    }%
  }%

  \foreach \i in {1,...,4}{%
    \foreach \j in {1,...,6}{%
      \node[
        circle,
        draw=ovNodeStroke,
        fill=white,
        minimum size=8mm,
        inner sep=0pt,
        line width=0.6pt,
        font=\tiny\color{ovNodeText}
      ] (g-\i-\j)
      at ({\sp*\j},{-\sp*\i})
      {$u_{\i,\j}$};%
    }%
  }%

  \draw[dashed, thick, ovTopology!80]
    ($(g-1-1.north west)+(-0.12,0.12)$)
    rectangle
    ($(g-3-3.south east)+(0.12,-0.12)$);

  \node[
    font=\scriptsize,
    ovTopology
  ] at (2.8,-0.75)
  {$\mathcal{N}(u)$: $3{\times}3$ Moore neighborhood};

  \foreach \i in {1,2,3}{
    \foreach \j in {1,2,3}{

      \ifnum\i=2
        \ifnum\j=2
        \else
          \node[
            circle,
            draw=ovNodeStroke,
            fill=white,
            minimum size=8mm,
            inner sep=0pt,
            line width=0.6pt,
            font=\tiny\color{ovNodeText}
          ] at ({\sp*\j},{-\sp*\i})
          {$u_{\i,\j}$};
        \fi
      \fi

      \ifnum\i=1
        \node[
          circle,
          draw=ovNodeStroke,
          fill=white,
          minimum size=8mm,
          inner sep=0pt,
          line width=0.6pt,
          font=\tiny\color{ovNodeText}
        ] at ({\sp*\j},{-\sp*\i})
        {$u_{\i,\j}$};
      \fi

      \ifnum\i=3
        \node[
          circle,
          draw=ovNodeStroke,
          fill=white,
          minimum size=8mm,
          inner sep=0pt,
          line width=0.6pt,
          font=\tiny\color{ovNodeText}
        ] at ({\sp*\j},{-\sp*\i})
        {$u_{\i,\j}$};
      \fi
    }
  }

  \node[
    circle,
    draw=ovFocusStroke,
    fill=white,
    minimum size=9mm,
    inner sep=0pt,
    line width=0.9pt,
    font=\tiny\color{ovFocusText}
  ] at ({\sp*2},{-\sp*2}) {$u$};

  \fill[
    pattern=north east lines,
    pattern color=ovGeom!40
  ] ({\sp*4},{-\sp*3}) circle (1.05);

  \draw[ovGeom!70, dashed]
    ({\sp*4},{-\sp*3}) circle (1.05);

  \draw[ovGeom!70, dashed]
    ({\sp*5},{-\sp*3}) circle (1.05);

  \begin{scope}
    \clip ({\sp*4},{-\sp*3}) circle (1.05);
    \clip ({\sp*5},{-\sp*3}) circle (1.05);

    \fill[ovSoft]
      ({\sp*4-1},{-\sp*3-1})
      rectangle
      ({\sp*5+1},{-\sp*3+1});
  \end{scope}

  \node[
    circle,
    draw=ovGeom!75,
    fill=white,
    minimum size=8mm,
    inner sep=0pt,
    font=\tiny\color{ovGeom!85}
  ] at ({\sp*4},{-\sp*3}) {$u$};

  \node[
    circle,
    draw=ovGeom!75,
    fill=white,
    minimum size=8mm,
    inner sep=0pt,
    font=\tiny\color{ovGeom!85}
  ] at ({\sp*5},{-\sp*3}) {$v$};

  \fill[ovGeom]
    ({\sp*4},{-\sp*3}) circle (0.06);

  \draw[
    ->,
    ovGeom!80,
    line width=0.8pt
  ]
    ({\sp*4-0.74},{-\sp*3-0.74}) --
    ({\sp*4},{-\sp*3})
    node[
      font=\scriptsize,
      midway,
      sloped,
      below,
      ovGeom
    ] {$R_u$};

  \node[
    font=\scriptsize,
    ovGeom
  ] at ({\sp*4-0.7},{-\sp*3+0.9})
  {$\mathcal{W}(u)$};

  \node[
    font=\scriptsize,
    ovGeom
  ] at ({\sp*5+0.7},{-\sp*3+0.9})
  {$\mathcal{W}(v)$};

  \node[
    font=\scriptsize,
    ovInput
  ] at ({\sp*4.5},{-\sp*3+0.6})
  {$\mathcal{H}(u,v)$};

  \node[
    font=\scriptsize,
    ovGeom
  ] at ({\sp*4},{-\sp*3-0.35})
  {\tiny $b(u)$};

  \node[
    font=\scriptsize,
    ovInput
  ] at ({\sp*1},{\yPortLabels})
  {$\mathcal{U}_{\mathrm{in}}$};

  \node[
    font=\scriptsize,
    ovStruct
  ] at ({\sp*6},{\yPortLabels})
  {$\mathcal{U}_{\mathrm{out}}$};

  \foreach \i in {1,...,4}{
    \node[
      rectangle,
      draw=ovGeom!70,
      fill=ovSoft,
      minimum width=8mm,
      minimum height=6mm,
      font=\tiny
    ] at (0.2,{-\sp*\i})
    {$s_\i$};
  }

  \node[
    font=\scriptsize,
    ovInput
  ] at (0.2,\yPortLabels)
  {$\mathcal{S}_{\mathrm{in}}$};

  \draw[
    ovInput!80!black,
    fill=ovSoft,
    fill opacity=0.55
  ] ({\sp*1},{-\sp*2}) circle (0.45);

  \node[
    circle,
    draw=ovInput!80!black,
    fill=white,
    minimum size=8.6mm,
    inner sep=0pt,
    line width=0.8pt,
    font=\tiny\color{ovNodeText}
  ] (inputrobot)
  at ({\sp*1},{-\sp*2})
  {$u_{2,1}$};

  \foreach \i in {1,3,4}{
    \draw[->,ovStruct!65]
      (0.5,{-\sp*\i}) --
      (g-\i-1.west);
  }

  \draw[->,ovStruct!65]
    (0.5,{-\sp*2}) --
    (inputrobot.west);

  \node[
    font=\scriptsize,
    ovInput!90!black
  ] at ({\sp*1+0.62},{-\sp*2-0.42})
  {$c_s$};

  \node[
    draw,
    rounded corners,
    minimum width=8mm,
    minimum height=42mm,
    fill=ovStructLight!15
  ] (out) at (10.8,-3.5) {};

  \node[
    font=\scriptsize,
    rotate=90
  ] at (10.8,-3.5)
  {output};

  \foreach \i in {1,...,4}{
    \draw[->,ovStruct!65]
      ([xshift=2pt]g-\i-6.east) --
      (out.west |- g-\i-6.east);
  }

  \draw[
    ->,
    ovStruct!60,
    very thick,
    dashed
  ]
  (0.2,\yFlow) --
  (10.8,\yFlow)
    node[
      font=\scriptsize,
      midway,
      below
    ] {dominant material progression};

\end{tikzpicture}%
}

\vspace{0.2em}

{\centering\scriptsize
\begin{tabular}{rl@{\hspace{1.5em}}rl@{\hspace{1.5em}}rl}

\textcolor{ovTopology}{$\mathcal{N}(u)$}: &
$3{\times}3$ Moore neighborhood &
\textcolor{ovGeom}{$\mathcal{W}(u)$}: &
reach disk (hatched) &
\textcolor{ovGeom}{$\mathcal{W}(v)$}: &
reach disk of $v$ \\[1pt]

\textcolor{ovInput}{$\mathcal{H}(u,v)$}: &
handoff zone &
\textcolor{ovGeom}{$b(u),\,R_u$}: &
base point, reach &
\textcolor{ovInput}{$\mathcal{S}_{\mathrm{in}},\,c_s$}: &
shelves, pick region \\[1pt]

\textcolor{ovInput}{$\mathcal{U}_{\mathrm{in}}$}: &
input-column robots &
\textcolor{ovStruct}{$\mathcal{U}_{\mathrm{out}}$}: &
output-column robots &
$\mathcal{U}=\{u_{i,j}\}$: &
population, $|\mathcal{U}|{=}MN$ \\

\end{tabular}\par
}
\caption{Physical organization and local interaction in AssemblyGrid v1.}
\label{fig:grid_overview}
\end{figure}

The logical representation of AssemblyGrid v1 is shown in Figure~\ref{fig:grid_overview}.

\paragraph{Robot grid and local topology.} 
For an $M\times N$ grid, the robot population is
\begin{equation}
\mathcal U=\{u_{i,j}:1\le i\le M,\;1\le j\le N\},
\qquad |\mathcal U|=MN,
\end{equation}
where $M$ and $N$ are the numbers of grid rows and columns, $u_{i,j}$ is the fixed robot at grid position $(i,j)$, and $\mathcal U$ is the complete robot population.
AssemblyGrid v1 distinguishes three forms of locality: observation locality $\mathcal G_{\mathrm{obs}}$, bilateral-handoff locality $\mathcal G_{\mathrm{handoff}}$, and cooperative-manipulation locality $\mathcal G_{\mathrm{manip}}$. These relations share the same Moore-radius-one topology in the default AssemblyGrid v1 configuration while remaining semantically distinct, since information access, material transfer and joint manipulation are different benchmark relations. Each robot's local interaction set is its closed Moore neighborhood $\mathcal N(u_{i,j})$ (Appendix~\ref{app:relations}). As illustrated in Figure~\ref{fig:grid_overview}, a robot away from the grid boundary has eight direct neighbors. Under Moore-radius-one manipulation locality, every mutually adjacent coalition is contained within at most a $(2\times2)$ local block. Thus $k_{\mathrm{topo}}^{\max}=\omega(\mathcal G_{\mathrm{manip}})$, which equals four for the official AssemblyGrid v1 grids; smaller coalitions may occupy any admissible subset of such a block.
\paragraph{Abstract geometric feasibility.}
\label{sec:motionfeas}
AssemblyGrid v1 uses a lightweight task-space geometry model to decide whether local robot interactions are spatially admissible. Figure~\ref{fig:grid_overview} illustrates the key quantities: robot $u$ has an abstract reachable workspace $\mathcal W(u)$ of radius $R_u$ about its base $b(u)$, and neighboring robots can hand off only inside the overlap of their reachable workspaces. Cooperative operations assign role-specific targets around a common task location; concurrent activities must additionally satisfy reach, clearance, and workspace-conflict constraints. Approach and retreat distances contribute to the abstract motion delay. Geometric distances are expressed in normalized task-space units; the formal workspace relations and geometry-profile ownership are specified in Appendix~\ref{app:parameters}, while the active profile or named variant is part of the benchmark-instance definition.

\paragraph{Scope of the geometry model.}
The geometry model represents spatial admissibility at the task level. Reach, workspace overlap, clearance, and interference determine whether handoffs, cooperative operations, and concurrent activities are admissible and contribute to their abstract execution time. Robot-specific kinematics, trajectory planning, grasping, and dynamics lie outside this abstraction and may be incorporated through higher-detail extensions.

\paragraph{Input, internal flow, and output.}
As shown in Figure~\ref{fig:grid_overview}, Raw material enters along the left boundary through $\mathcal S_{\mathrm{in}}$ and is picked by the left-column robots $\mathcal U_{\mathrm{in}}$; finished products leave through the right-column robots $\mathcal U_{\mathrm{out}}$ (Appendix~\ref{app:relations}). Material transfer between robots occurs through local bilateral handoff.

\subsection{Product recipes and operation requirements}
\label{sec:recipes}

A product progresses according to its recipe without following a predefined robot route. The recipe specifies the production operations that must be completed, the order in which they may occur, the material each operation consumes or produces, and any coalition or manipulation roles required for execution. The recipe separates \emph{what must be produced} from \emph{which robots perform each operation and where}.

\paragraph{Material and product state.}
For every material item $q$, representing a part or subassembly, the environment records its location, holder when applicable, abstract task pose, process status, and associated product. In the default setting a robot holds at most one item, $\sum_q \mathbf 1\{\mathrm{holder}_t(q)=u\}\le1$, where $\mathrm{holder}_t(q)$ denotes the robot holding item $q$ at decision epoch $t$, when one exists.

\paragraph{Recipe graph.}
Each product type $r\in\mathcal R$ is defined by a recipe structure $\Gamma_r=(\mathcal T_r,\mathcal E_r,\mathcal X_r)$, whose precedence component $(\mathcal T_r,\mathcal E_r)$ is a directed acyclic graph (DAG). Here $\mathcal T_r$ contains the recipe operations, $\mathcal E_r$ contains mandatory predecessor relations, and $\mathcal X_r$ records mutually exclusive operation alternatives. A \emph{sequence} enforces a declared order, an \emph{AND} relation requires all predecessor branches, and an \emph{exclusive OR (XOR)} relation selects one alternative and closes its mutually exclusive siblings for that product unit. Thus sequence and AND are encoded by the mandatory predecessor structure, while XOR is represented explicitly by $\mathcal X_r$.
For product unit $z$ and recipe operation $\tau$, the pair $(z,\tau)$ denotes that particular operation for that product and is referred to as an \emph{operation occurrence}; its readiness indicator $\mathrm{ready}_t(z,\tau)\in\{0,1\}$ (Appendix~\ref{app:rules}) states whether the recipe precedence permits it to proceed. Readiness alone does not guarantee execution: the required material, a suitable robot group, roles, site and auxiliary resources, reach, clearance, and compatibility with simultaneous work must also be satisfied.
Successful completion applies the recipe-defined material transition $\Delta_\tau$: required inputs are consumed or incorporated and the resulting part or subassembly state is created or updated atomically. Applying $\Delta_\tau$ atomically makes material conservation, subassembly provenance, joins, and terminal product formation explicit without a second progress mechanism separate from the recipe. To make these requirements explicit, every recipe operation records what it consumes, how successful completion transforms the material state, how many robots it needs, the roles required, which exclusive resources it uses, and how long productive processing takes; its geometric requirements and work location follow from the coalition and the active geometry profile. We write this compactly as

\begin{equation}
\tau=
(
\mathrm{type}_\tau,
\mathrm{inputs}_\tau,
\Delta_\tau,
k_\tau^{\min},
k_\tau^{\max},
\mathrm{roles}_\tau,
\mathrm{res}_\tau,
d_\tau
),
\end{equation}

where $\Delta_\tau$ is the recipe-defined material-state transition and the remaining fields are summarized in Table~\ref{tab:operation_fields}.

\begin{table}[H]
\centering
\caption{Recipe-operation fields.}
\label{tab:operation_fields}
\small
\begin{tabularx}{0.94\linewidth}{p{0.25\linewidth}Y}
\toprule
Symbol & Meaning \\
\midrule
$\mathrm{type}_\tau$ & operation type \\
$\mathrm{inputs}_\tau$ & required input material or subassemblies \\
$\Delta_\tau$ & material-state transition applied atomically on successful completion \\
$[k_\tau^{\min},k_\tau^{\max}]$ & admissible coalition-size interval \\
$\mathrm{roles}_\tau$ &  ordered manipulation roles and any material-role bindings used by the canonical assignment; AssemblyGrid v1 instantiates exactly one role per participating robot \\
$\mathrm{geom}_\tau$ & derived, not declared by the recipe: geometric requirements fixed by the active geometry profile \\
$\mathrm{sites}_\tau$ & derived, not declared by the recipe: the common target point of the input holders \\
$\mathrm{res}_\tau$ & additional exclusive resources \\
$d_\tau$ & nominal productive processing duration \\
\bottomrule
\end{tabularx}
\end{table}

\paragraph{Recipe-defined cooperation.}
Each operation declares the coalition sizes that are semantically valid:
\begin{equation}
k_\tau^{\min}
\le |C|
\le
\min\!\left(k_\tau^{\max},k_{\mathrm{topo}}^{\max}\right),
\end{equation}
where $C$ is a concrete coalition and $k_{\mathrm{topo}}^{\max}=\omega(\mathcal G_{\mathrm{manip}})$ is the clique-number bound imposed by the active manipulation topology. For the official AssemblyGrid v1 Moore-radius-one grids, this value is four. By default $k_\tau^{\min}=k_\tau^{\max}$, and the recipe therefore requests an exact size; a wider declared interval remains capped by the active topology. When $k_\tau^{\min}<k_\tau^{\max}$ and explicit roles are used, the AssemblyGrid v1 role declaration covers the maximum admissible team. Material-bound roles are assigned first to the robots holding their associated inputs, and the remaining participating robots, ordered row-major, receive the remaining declared roles in recipe order; a realized size-$k$ candidate therefore instantiates exactly $k$ roles.
The productive duration $d_\tau$ is fixed with respect to coalition size; cooperation therefore represents operation requirements such as holding, stabilizing, aligning, inserting, or fastening. Geometry may nevertheless add approach and retreat time. Before evaluation, recipes are validated for schema correctness, precedence consistency, material semantics, coalition-size compatibility, and realizability under the declared benchmark constraints, with the complete validation, topology, transition, and reporting definitions given in Appendix~\ref{app:formal-spec}.

\subsection{Local decisions and temporary coalitions}

Time is represented discretely, with decision epoch $t$ denoting the beginning of simulation tick $t$ and each tick serving as the benchmark unit of simulated duration. Available robots select task actions at these decision epochs, while committed activities may remain active across multiple ticks. At each decision epoch, an available robot selects among feasible local actions. A ready operation occurrence $(z,\tau)$ may be executable by more than one admissible robot set. Each feasible combination of the operation and a participating robot set is represented as an \emph{execution candidate} (hereafter, candidate refers to an execution candidate). The identity of an execution candidate is given by the canonical key $\kappa(c)=(z,\tau,\operatorname{sort}_{\mathrm{rm}}(U_c))$, where $U_c$ is the proposed robot set and $\operatorname{sort}_{\mathrm{rm}}$ denotes row-major sorting. For a fixed benchmark instance and decision state, AssemblyGrid v1 uses a fixed rule to decide how it is carried out in the environment: Material-bound roles are assigned to the robots holding the required inputs, and the remaining roles are assigned to the other robots in the proposed set in row-major order. The operation site is computed from the relevant robot base positions, and the active geometry profile determines the geometric assignment. Thus, each candidate key corresponds to a single concrete realization. These attributes therefore specify one concrete realization without creating a second execution candidate identity for the same key.
The task-level action types are
\begin{equation}
\mathcal A^u=
\{
\mathrm{idle},
\mathrm{pick}(q),
\mathrm{handoff}(v,q),
\mathrm{receive}(v,q),
\mathrm{support}(c),
\mathrm{deliver}(z)
\},
\end{equation}
where $\mathcal A^u$ is the task-level action set of robot $u$, $q$ is a locally accessible material item, $v$ is a direct handoff neighbor, $c$ is an execution candidate, and $z$ is a product unit ready for delivery. By selecting $\mathrm{support}(c)$, robot $u$ chooses to participate in the particular operation execution represented by candidate $c$. Recipe-specific operations such as inspection or assembly are represented through candidate selection, without separate action types.
Each agent thus selects a task-level activity, while the environment evaluates the geometry and timing associated with the selected option under the active benchmark profile.

\paragraph{Bilateral handoff.}
A handoff occurs only when two neighboring robots make matching choices in the same decision epoch: one robot chooses $\mathrm{handoff}(v,q)$ and the other chooses $\mathrm{receive}(u,q)$. The material can move only through the shared reachable region $\mathcal H(u,v)$ illustrated in Figure~\ref{fig:grid_overview} (Appendix~\ref{app:rules}). Requiring both robots to choose the transfer preserves handoff as a coordinated decision and prevents the environment from executing the transfer automatically.

\paragraph{Dynamic coalition formation.}
\label{sec:teamformation}
Each eligible agent may select $\mathrm{support}(c)$ for one candidate, while admission occurs only when every required robot supports the same option and the environment confirms that the combined execution is feasible. The resulting admitted activity $e=e(c)$ is executed by the temporary coalition $C(e)$, whose participating robots remain committed to their assigned roles until the activity completes, after which they return to the available resource pool. Support is a pre-admission choice, whereas commitment begins only after admission.

\paragraph{Execution-candidate participation lifecycle.}
By selecting $\mathrm{support}(c)$, robot $u$ records its intention to participate in execution candidate $c$; the record may persist across decision epochs, and the required robots need not select the candidate simultaneously. A robot can support at most one candidate at a time. If support is created at decision epoch $t_0$, it is eligible for admission at epochs $t_0,\ldots,t_0+3$: admission is checked before the clock advances, and expiration is applied afterward; consequently, the first expiry occurs at $t_0+4$. Re-selecting the same candidate refreshes neither the timeout nor the one-epoch switching lock; switching to a different candidate is permitted from $t_0+1$. A fully supported candidate that is still blocked at $t_0+3$ remains unadmitted and expires at $t_0+4$ unless it becomes invalid earlier. Accordingly, pending-support state retains the candidate identity and the timing information needed to determine its remaining lifetime and switching eligibility.

\paragraph{Coordination cue.}
Coordination in AssemblyGrid v1 is informed by an environment-mediated support-fraction cue without introducing an explicit messaging channel. For each execution candidate available to an agent, the observation includes a \emph{support fraction}: the number of robots required by the candidate whose support is already recorded, divided by the number of robots the candidate requires. Observations at decision epoch $t$ are formed from the state before the actions selected at $t$ are applied. The cue indicates how much pending participation has accumulated for that specific execution candidate without revealing the complete support state. 

\paragraph{Task selection and geometric execution.}
\label{sec:motionexecution}
Each eligible robot selects its own task-level action. When the required members support the same execution candidate, the environment evaluates its joint compatibility before admission. If candidate $c$ is admitted, the resulting activity $e=e(c)$ has realized duration
\begin{equation}
D(e)=
D_{\mathrm{approach}}(e)
+
D_{\mathrm{process}}(e)
+
D_{\mathrm{retreat}}(e),
\end{equation}
where $D_{\mathrm{approach}}(e)$ and $D_{\mathrm{retreat}}(e)$ are the abstract approach and retreat delays, respectively, and $D_{\mathrm{process}}(e)=d_\tau$ is the productive processing duration for an activity executing operation $\tau$ in AssemblyGrid v1. Decomposing execution into approach, process, and retreat keeps the action space at task level while allowing reach, clearance, and concurrent workspace conflicts to affect feasibility and resource occupancy. Formally, a robot with base point $b_u$ and reach radius $r_u$ can serve a target point $x$ only if $\lVert x-b_u\rVert_2\le r_u$. Each activity $e$ is represented by the planar approach and retreat segments $\mathcal S(e)$ of its members and its target point $x(e)$. With $\operatorname{dist}(s,s')$ the minimum Euclidean distance between two segments, zero when they intersect, two activities $e$ and $e'$ conflict geometrically if $\min_{s\in\mathcal S(e),\,s'\in\mathcal S(e')}\operatorname{dist}(s,s')<c(e,e')$ or $\lVert x(e)-x(e')\rVert_2<c(e,e')$, where $c(e,e')=\rho_{\mathrm{conf}}+\max_{u\in C(e)\cup C(e')}c_u$ combines the workspace-conflict radius with the largest robot clearance $c_u$ among the members of both activities.

A fully supported candidate is admitted only if its robots, required material, role requirements, operation site, exclusive auxiliary resources, and abstract geometric predicates are jointly compatible. Overlap between possible local coalition footprints is not itself a conflict. Two admitted or executing activities conflict if they share an exclusive robot or auxiliary resource, or if their simultaneous abstract motion segments or target workspaces violate the active geometry setting's clearance rule.

\paragraph{Action masking.}
Per-agent masks remove only hard impossibilities that can be determined from the agent's permitted local information, such as an unavailable material pick, an unreachable local target under the active geometry setting, or an action outside the robot's static capability. Let $m_t^u$ denote the local action mask supplied to agent $u$ at decision epoch $t$. The mask is limited to locally detectable impossibilities; matching choices by other agents and hidden remote conflicts remain unresolved until the joint transition. Coalition agreement, matching handoff choices, remote contention, and conflicts between otherwise legal local actions are resolved by the joint transition model.

\paragraph{Candidate exposure boundary.}
Candidates may be generated from the global state, but robot $u$ sees only those it can locally take part in. Remote conflicts remain hidden and are handled only during joint admission.

\subsection{Decentralized information model}
\label{sec:obscontract}

Each agent selects its action from local execution information, while the next production state depends on the combined actions of all agents; the formal model therefore separates a global state and transition process from the action and observation spaces available to each agent. Denoting the AssemblyGrid v1 environment by $\mathcal E_{\mathrm{AG}}$, we represent it as the decentralized partially observable controlled process

\begin{equation}
\mathcal E_{\mathrm{AG}}
=
\left\langle
\mathcal U,
\mathcal S,
\{\mathcal A^u\},
P,
\{\mathcal O^u\},
O
\right\rangle,
\end{equation}
where $\mathcal U$ is the robot population, $\mathcal S$ is the global state space, $\{\mathcal A^u\}$ are the per-agent action spaces, $P$ is the joint transition model, $\{\mathcal O^u\}$ are the local observation spaces, and $O$ is the observation function (Appendix~\ref{app:notation}). The recipe family, feasibility rules, timing model, success condition, and termination rule complete the environment definition.

At decision epoch $t$, agent $u$ receives execution information
\begin{equation}
I_t^u=(o_t^u,m_t^u),
\end{equation}
where $o_t^u\in\mathcal O^u$ is its local observation and $m_t^u$ is the local action mask defined above. A decentralized controller can be represented by a collection of agent policies $\Pi=\{\pi^u\}_{u\in\mathcal U}$. For a history-dependent controller,
\begin{equation}
a_t^u\sim\pi^u(\cdot\mid h_t^u),
\qquad
h_t^u=(I_0^u,a_0^u,\ldots,a_{t-1}^u,I_t^u),
\end{equation}
while a memoryless controller uses $a_t^u\sim\pi^u(\cdot\mid I_t^u)$. The independently selected agent actions form the joint action
\begin{equation}
\mathbf a_t=(a_t^u)_{u\in\mathcal U}
\in
\prod_{u\in\mathcal U}\mathcal A^u,
\qquad
s_{t+1}\sim P(\cdot\mid s_t,\mathbf a_t).
\end{equation}
Thus multiple decision makers select their own action components, while their consequences are coupled through one environment transition. In this paper, a \emph{policy} is an agent-level action-selection mapping used by a controller.

\paragraph{Reward independence.}
$\mathcal E_{\mathrm{AG}}$ is specified independently of reward and discounting. Reward functions and discount factors are introduced only by learning methods that require them. Reinforcement-learning algorithms may add a learning reward $R$ and discount factor $\gamma$ to instantiate a cooperative decentralized partially observable Markov decision process (Dec-POMDP) \citep{oliehoek2016decpomdp} on top of $\mathcal E_{\mathrm{AG}}$ when the reward is common to all agents, or a partially observable stochastic game when it contains agent-specific terms, while planners, schedulers, optimization methods, and heuristic controllers use $\mathcal E_{\mathrm{AG}}$ directly. Benchmark outcomes are defined and scored through common production and task-success measures independently of accumulated learning reward. Learning return may additionally be reported as a method-specific optimization diagnostic, without becoming part of the benchmark task definition. Methods with different internal objectives therefore remain comparable under the same benchmark instances and metrics. The corresponding global state $s_t$ carries material and product state, recipe progress, robot holdings, pending candidate-support records and their lifecycle timing, coalition commitments, resources, abstract geometric state, and benchmark time, as detailed in Appendix~\ref{app:entities}.

\paragraph{Local observation principle.}
The decentralized observation $o_t^u$ contains the task information locally available to agent $u$, including the agent's own state, nearby robot and material state, and the execution candidates in which the agent can participate. The complete execution information $I_t^u$ additionally contains the local action mask $m_t^u$. Table~\ref{tab:observation_contract} summarizes the boundary directly: an agent sees local task state, its available execution options, its role and local geometric information, and the aggregate support for those options, while it does not see the global production state, remote contention, detailed intentions of other agents, or explicit messages.

\begin{table}[H]
\centering
\caption{Decentralized execution information in AssemblyGrid v1.}
\label{tab:observation_contract}
\small
\setlength{\tabcolsep}{6pt}
\begin{tabularx}{0.98\linewidth}{@{}Y Y@{}}
\toprule
Available to a policy for agent $u$ & Not exposed to agent $u$ \\
\midrule
$\bullet$\hspace{0.3em}Own task-level state and current commitment & $\bullet$\hspace{0.3em}Global benchmark state \\
$\bullet$\hspace{0.3em}Task-level state of local neighbors & $\bullet$\hspace{0.3em}State of remote robots \\
$\bullet$\hspace{0.3em}Locally visible material and transfer opportunities & $\bullet$\hspace{0.3em}Remote material and remote resource contention \\
$\bullet$\hspace{0.3em}Eligible local execution candidates and their defining attributes & $\bullet$\hspace{0.3em}Global joint action or global feasibility oracle \\
$\bullet$\hspace{0.3em}Own candidate role, expected duration, and locally computable reach margin & $\bullet$\hspace{0.3em}Raw robot configurations and low-level trajectories \\
$\bullet$\hspace{0.3em}Local action mask for hard impossibilities detectable from permitted local information & $\bullet$\hspace{0.3em}Masking based on hidden nonlocal conflicts \\
$\bullet$\hspace{0.3em}Aggregate support fraction for each candidate & $\bullet$\hspace{0.3em}Explicit supporter list or other agents' pending intentions \\
$\bullet$\hspace{0.3em}Recipe information needed to interpret local choices & $\bullet$\hspace{0.3em}Free-form or addressable messages \\
\bottomrule
\end{tabularx}
\end{table}

The canonical operation occurrence and proposed robot set identify the same execution candidate for every eligible agent. The associated operation site and geometric assignment are deterministic candidate attributes, while role information is presented from each agent's own perspective. Each eligible agent must nevertheless be able to distinguish the candidate options available to it. The local-information restriction governs action selection during evaluation, while training may use additional global information provided that the resulting execution policy still selects actions from the permitted local information $I_t^u$. Section~\ref{sec:evalresults} uses this distinction to separate centralized training from controllers that use privileged information during execution. Memory over an agent's own execution-information and action history preserves the same information boundary, since that history contains only information already available to the agent. Recurrent or history-dependent policies are therefore permitted within the decentralized information structure.

\section{Benchmark design, scenario suite, and measurement}
\label{sec:benchmarkdef}

This section defines the benchmark instances, scenario suite, task-success conditions, and evaluation measures used for reproducible comparison. The official suite contains the \textsc{Flow}, \textsc{Coalition}, and \textsc{Concurrency} workload families, while geometric admissibility constrains all three and is varied separately in matched controls.

\subsection{Benchmark instances and reproducibility}
\label{sec:instances}

\paragraph{Benchmark, runtime, method, and evaluation parameters.}
A benchmark instance is one concrete production problem. Settings that can change which actions are possible, how the production state evolves, what information agents receive, or what counts as task success are benchmark-defining. These include the grid and interaction topology, recipe, robot capabilities, geometry settings, material and resource structure, demand process, work-in-progress (WIP) bound, and initial condition. AssemblyGrid v1 also fixes runtime semantics shared by the official scenarios, including the tick and transition order and the coalition-support switching and timeout rules. These runtime semantics are benchmark controlled and cannot be altered by a controller during evaluation; changing them requires an explicitly named benchmark or runtime variant. Method-level quantities such as network architecture, optimizer, learning reward, and learning rate may differ between controllers without changing the benchmark task. Evaluation-level quantities such as the declared horizon, dataset split, logging, and plotting procedure belong to the evaluation protocol and must be held fixed within a reported comparison. Some quantities are derived not chosen. In particular, the topological coalition limit is
\begin{equation}
 k_{\mathrm{topo}}^{\max}=\omega(\mathcal G_{\mathrm{manip}})
\end{equation}
where $\mathcal G_{\mathrm{manip}}$ is the direct-manipulation graph and $\omega(\mathcal G_{\mathrm{manip}})$ is its clique number. The topological coalition limit is therefore determined by the manipulation graph and is never a free parameter. Since geometry changes which commitments are feasible, changing the geometry setting also changes the benchmark instance.

\paragraph{Geometry settings.}
A \emph{geometry profile} is a named set of spatial and timing assumptions that AssemblyGrid v1 applies together whenever an activity is checked for geometric admissibility. The profile determines how robot reach relates to grid spacing, the clearance required between simultaneous workspaces, the approach and retreat timing, and the threshold used to flag workspace conflicts. Keeping these settings together under one named profile makes the geometric assumptions visible and consistent across evaluations. A \emph{geometry-control variant} is an explicitly declared change to one or more of these settings for a matched study; it does not silently redefine the workload family.

\paragraph{Scenarios, instances, and random seeds.}
The evaluation hierarchy is \emph{workload family $\rightarrow$ difficulty level $\rightarrow$ official scenario $\rightarrow$ benchmark instance $\rightarrow$ evaluation execution}. An official scenario gives the complete benchmark configuration for one family and difficulty level. A \emph{benchmark instance} is one concrete production problem generated from that scenario. Its generation random seed determines any scenario-level randomized choices. If the environment contains stochastic evolution, an execution random seed determines that realization during evaluation. The algorithm or training random seed controls controller construction, such as initialization or exploration, and is not part of the benchmark instance. Exact reproducibility metadata and instance fingerprints are specified in Appendix~\ref{app:reproducibility}.

\paragraph{Feasibility.}
A benchmark instance is \textbf{feasible} when at least one successful execution exists independently of the evaluated policy; the material-reachability, handoff-chain, coalition-size, operation-site, and resource conditions this requires are stated in Appendix~\ref{app:constraints}. An instance that violates them is invalid. Task success is defined in Section~\ref{sec:tasksuccess}.

\subsection{The AssemblyGrid v1 scenario suite}
\label{sec:suite}

The official AssemblyGrid v1 suite contains three workload families with different purposes (Table~\ref{tab:suites}). \textsc{Flow} emphasizes recipe depth and material progression, \textsc{Coalition} emphasizes temporary multi-robot operations, and \textsc{Concurrency} emphasizes simultaneous productive work under increasing production pressure, with each family providing easy, medium, and hard scenarios. These labels encode predefined family-specific workload configurations and are compared within a workload family; they do not imply a monotonic ordering of every controller outcome. Scaling, topology transfer, failures, communication interventions, and robot-specific physics remain outside the AssemblyGrid v1 suite evaluated in this study. All nine official scenario definitions are released as runnable benchmark configurations from which conformance-checked benchmark instances are generated using declared instance seeds, and the MARL-controller characterization in this paper covers all three workload families at every difficulty level. The non-learning executability validation is presented in detail for the \textsc{Coalition} family, which most directly exercises temporary multi-robot cooperation together with local information, material progression, and geometric feasibility.

\begin{table}[H]
\centering
\caption{Official AssemblyGrid v1 workload families.}
\label{tab:suites}
\small
\begin{tabularx}{\textwidth}{p{0.15\textwidth}p{0.29\textwidth}Y}
\toprule
Family & Primary workload characteristic & Benchmark question \\
\midrule
\textsc{Flow} & Sequential production depth under increasing system load & Can controllers route material and respect increasingly deep recipe precedence without a coalition requirement? \\
\textsc{Coalition} & Temporary role-constrained coalition formation and increasing coalition structure & Can local robots form, maintain, and dissolve temporary coalitions as the required cooperative structure becomes more demanding? \\
\textsc{Concurrency} & Productive overlap under increasing production pressure & Can controllers exploit independent ready work while avoiding congestion and unnecessary blocking? \\
\bottomrule
\end{tabularx}
\end{table}

All nine official scenarios use the same default geometry setting, with reach, clearance, motion delay, and workspace-conflict checks remaining active. The matched geometry study changes only the workspace-conflict radius and keeps the production scenario, underlying generated production realizations, generation and execution seeds, controller, and all non-geometric settings fixed; the resulting geometry settings are matched instance variants. The matched design isolates geometric interference from workload and controller changes, with the corresponding results reported in Section~\ref{sec:mechanismevidence}. Benchmark instances are generated deterministically from their declared random seeds. Conditional on the environment version, scenario configuration, instance-generation seed, execution seed, and joint action sequence, AssemblyGrid v1 transitions are deterministic. The determinism claim concerns environment transitions; exact bitwise reproduction of neural-network training can additionally depend on hardware, accelerator kernels, framework versions, and deterministic-computation settings. The controller experiments use separate fixed random-seed sets for training, validation, and final testing, as specified in Section~\ref{sec:evalresults}.

\subsection{Difficulty levels: easy, medium, and hard}
\label{sec:difficultylevels}

\begin{figure}[H]
\centering
\begin{tikzpicture}[
  flowop/.style={
    circle,
    draw=AGParallel!85,
    fill=white,
    minimum size=5.6mm,
    inner sep=0pt,
    font=\scriptsize,
    text=black!70
  },
  op/.style={
    circle,
    draw=AGParallel!85,
    fill=white,
    minimum size=6.5mm,
    inner sep=0pt,
    font=\scriptsize,
    text=black!70
  },
  opc/.style={
    circle,
    draw=AGProdOutDraw,
    fill=AGProdInFill,
    minimum size=6.5mm,
    inner sep=0pt,
    font=\scriptsize,
    text=AGProdOutDraw
  },
  prodin/.style={
    fill=AGProdInFill!50,
    draw=AGProdInDraw!50,
    rounded corners=2pt
  },
  prodindot/.style={
    fill=AGProdInDraw!75!black
  },
  prodout/.style={
    fill=AGProdInFill,
    draw=AGProdInDraw,
    rounded corners=2pt
  },
  prodoutdot/.style={
    fill=AGProdInDraw!60!black
  },
  flowar/.style={
    -{Latex[length=1.6mm]},
    draw=AGParallel!80
  },
  cellbox/.style={
    draw=AGCellDraw,
    fill=AGCellFill,
    rounded corners=2pt
  },
  lvl/.style={
    font=\scriptsize\bfseries,
    text=AGProdOutDraw
  },
  fam/.style={
    font=\footnotesize\bfseries,
    text=black!85,
    anchor=east,
    align=right,
    text width=1.75cm
  },
  focus/.style={
    font=\tiny,
    text=black!60,
    anchor=east,
    align=right,
    text width=1.75cm
  },
  note/.style={
    font=\tiny,
    text=black!65,
    align=center
  },
  x=1cm, y=1cm
]

\def\colA{1.6}
\def\colB{6.0}
\def\colC{10.5}
\def\labelx{-0.75}
\def\labelxC{-0.75}
\def\labelwidthC{2.35cm}

\foreach \cx in {\colA,\colB,\colC}{
  \draw[cellbox] (\cx-1.95, 0.15) rectangle (\cx+1.95,-1.75);
  \draw[cellbox] (\cx-1.95,-1.95) rectangle (\cx+1.95,-3.85);
  \draw[cellbox] (\cx-1.95,-4.05) rectangle (\cx+1.95,-5.95);
}

\node[lvl] at (\colA,0.75) {EASY};
\node[lvl] at (\colB,0.75) {MEDIUM};
\node[lvl] at (\colC,0.75) {HARD};

\node[fam]   at (\labelx,-0.58) {\textsc{Flow}};
\node[focus] at (\labelx,-0.94) {recipe depth};

\node[fam]   at (\labelx,-2.55) {\textsc{Coalition}};
\node[focus] at (\labelx,-2.91) {coalition structure};

\node[
  font=\footnotesize\bfseries,
  text=black!85,
  anchor=east,
  align=right,
  text width=\labelwidthC
] at (\labelxC,-4.78) {\textsc{Concurrency}};

\node[
  font=\tiny,
  text=black!60,
  anchor=east,
  align=right,
  text width=\labelwidthC
] at (\labelxC,-5.14) {production pressure};

\node[flowop] (f1) at (\colA,-0.55) {1};
\node[note] at (\colA,-1.38) {1 unary operation};

\node[flowop] (m1) at ({\colB-0.85},-0.55) {1};
\node[flowop] (m2) at ({\colB},-0.55) {2};
\node[flowop] (m3) at ({\colB+0.85},-0.55) {3};
\draw[flowar] (m1) -- (m2);
\draw[flowar] (m2) -- (m3);
\node[note] at (\colB,-1.38) {3 unary operations};

\node[flowop] (h1) at ({\colC-1.60},-0.55) {1};
\node[flowop] (h2) at ({\colC-0.80},-0.55) {2};
\node[flowop] (h3) at ({\colC},-0.55) {3};
\node[flowop] (h4) at ({\colC+0.80},-0.55) {4};
\node[flowop] (h5) at ({\colC+1.60},-0.55) {5};
\draw[flowar] (h1) -- (h2);
\draw[flowar] (h2) -- (h3);
\draw[flowar] (h3) -- (h4);
\draw[flowar] (h4) -- (h5);
\node[note] at (\colC,-1.38) {5 unary operations};

\draw[prodin] ({\colA-0.55},-2.98) rectangle ({\colA+0.55},-2.28);
\foreach \x in {{\colA-0.25},{\colA+0.25}}
  \fill[prodindot] (\x,-2.63) circle (2.1pt);
\node[note] at (\colA,-3.55) {one 2-robot operation};

\draw[prodin] ({\colB-1.50},-2.98) rectangle ({\colB-0.40},-2.28);
\foreach \x in {{\colB-1.20},{\colB-0.70}}
  \fill[prodindot] (\x,-2.63) circle (2.1pt);

\draw[prodout] ({\colB+0.40},-2.98) rectangle ({\colB+1.50},-2.28);
\foreach \x in {{\colB+0.70},{\colB+1.20}}
  \fill[prodoutdot] (\x,-2.63) circle (2.1pt);

\draw[flowar] ({\colB-0.40},-2.63) -- ({\colB+0.40},-2.63);
\node[note] at (\colB,-3.55) {2 robots, then 2 to 4 robots};

\draw[prodin] ({\colC-1.20},-2.98) rectangle ({\colC-0.10},-2.28);
\foreach \x in {{\colC-0.90},{\colC-0.40}}
  \fill[prodindot] (\x,-2.63) circle (2.1pt);

\draw[prodout] ({\colC+0.30},-3.03) rectangle ({\colC+1.20},-2.13);
\foreach \x/\y in {
  {\colC+0.53}/-2.40,
  {\colC+0.97}/-2.40,
  {\colC+0.53}/-2.76,
  {\colC+0.97}/-2.76}
  \fill[prodoutdot] (\x,\y) circle (2.1pt);

\draw[flowar] ({\colC-0.10},-2.63) -- ({\colC+0.30},-2.63);
\node[note] at (\colC,-3.55) {2 robots, then 4 robots};

\foreach \c/\wip/\release in {\colA/2/10, \colB/5/5, \colC/10/2} {
  \node[op] (a\wip) at ({\c-0.38},-4.52) {a};
  \node[op] (b\wip) at ({\c-0.38},-5.20) {b};
  \node[opc] (j\wip) at ({\c+0.52},-4.86) {$\bowtie$};
  \draw[flowar] (a\wip) -- (j\wip);
  \draw[flowar] (b\wip) -- (j\wip);
}

\node[note] at (\colA,-5.71) {WIP 2, release every 10 ticks};
\node[note] at (\colB,-5.71) {WIP 5, release every 5 ticks};
\node[note] at (\colC,-5.71) {WIP 10, release every 2 ticks};

\end{tikzpicture}
\caption{Difficulty levels across the three workload families. The figure highlights each family's primary progression; complete scenario levels may change several parameters together, as listed in Table~\ref{tab:cellparameters}.}
\label{fig:difficulty}
\end{figure}
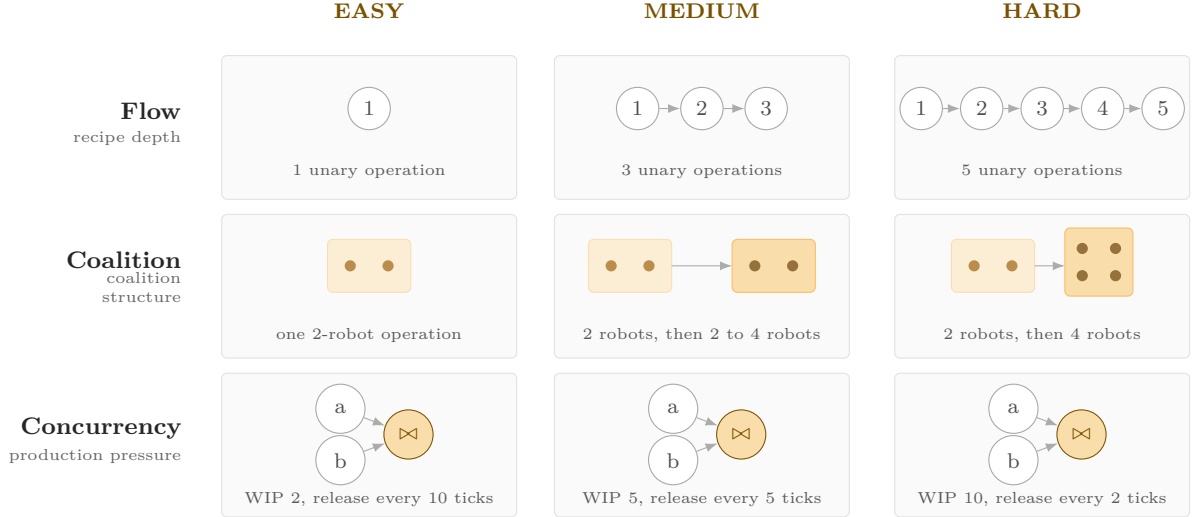

Within each family, easy, medium, and hard define difficulty levels. A level may change several parameters when those parameters jointly create the intended workload structure; easy-to-hard differences therefore describe complete scenarios. Figure~\ref{fig:difficulty} explains the main change within each family, including the WIP pressure used in the \textsc{Concurrency} family, while Table~\ref{tab:cellparameters} gives the complete scenario settings.

Across the official suite, all nine scenarios use the default normalized workspace-conflict radius $\rho_{\mathrm{conf}}=0.25$; geometry-control variants are declared separately, while Table~\ref{tab:cellparameters} reports the remaining workload parameters that define each scenario. Keeping this geometry setting fixed across the workload suite allows differences among \textsc{Flow}, \textsc{Coalition}, and \textsc{Concurrency} to be interpreted without simultaneously changing the common spatial-interference layer.

\begin{table}[H]
\centering
\caption{Official AssemblyGrid v1 scenarios.}
\label{tab:cellparameters}
\footnotesize
\setlength{\tabcolsep}{3.2pt}
\begin{tabularx}{\textwidth}{@{}llcccYcc@{}}
\toprule
Family & Difficulty & Grid & WIP & Release & Recipe & $|\mathcal T_r|$ & $\max_\tau k_\tau^{\min}$ \\
\midrule
\textsc{Flow} & easy   & $4\times5$ & 3  & 8  & single unary operation & 1 & 1 \\
               & medium & $5\times6$ & 5  & 5  & 3-stage serial & 3 & 1 \\
               & hard   & $6\times8$ & 10 & 2  & 5-stage serial & 5 & 1 \\
\addlinespace
\textsc{Coalition} & easy   & $5\times6$ & 4 & 7 & single 2-robot operation & 1 & 2 \\
               & medium & $5\times6$ & 4 & 5 & fixed 2-robot operation, then variable 2-to-4-robot operation & 2 & 2 \\
               & hard   & $6\times8$ & 1 & 8 & 2-robot then 4-robot operations & 2 & 4 \\
\addlinespace
\textsc{Concurrency} & easy   & $5\times6$ & 2  & 10 & two unary branches with 2-robot join & 3 & 2 \\
               & medium & $5\times6$ & 5  & 5  & two unary branches with 2-robot join & 3 & 2 \\
               & hard   & $6\times8$ & 10 & 2  & two unary branches with 2-robot join & 3 & 2 \\
\bottomrule
\end{tabularx}
\end{table}

\noindent\textbf{\textsc{Flow}.} Sequential production is the primary challenge. The recipe grows from one unary operation to three and then five unary operations, the grid and WIP limit increase, and products are released more frequently. The family therefore tests routing and precedence handling under progressively larger serial-flow scenarios.

\noindent\textbf{\textsc{Coalition}.} Temporary multi-robot cooperation is the primary challenge. The easy scenario requires one fixed two-robot operation. The medium scenario uses a fixed two-robot predecessor followed by a variable-arity operation that admits coalitions of two to four robots. Because productive duration does not depend on coalition size and skill semantics are inactive, a third or fourth robot does not bring any benefit, and thus this operation also tests whether controllers avoid committing more robots than it requires. Roles within a coalition are assigned by the environment's canonical rule so that controllers can select the coalition composition without separately deciding the robot-to-role assignment. The hard scenario requires a fixed four-robot operation after a two-robot predecessor. The hard scenario uses a WIP cap of one, thereby making the stricter minimum coalition requirement the primary workload change; the complete scenario also changes the grid size as reported in Table~\ref{tab:cellparameters}.

\noindent\textbf{\textsc{Concurrency}.} The same branching recipe is used in all three scenarios: two independent unary branches followed by a pairwise join. Increasing the WIP limit and product release frequency exposes more simultaneous ready work, and the hard scenario also enlarges the grid. The family therefore provides progressively higher-load scenarios in which productive overlap and contention become more prominent. Because WIP, release interval, and, for the hard level, grid size change together, the family is a workload progression.

\subsection{Task success and evaluation modes}
\label{sec:tasksuccess}

A reproducible evaluation needs three distinct definitions: when an individual product counts as successful, how the evaluation is bounded, and how performance is measured. Product success is defined as a state condition for one product, while the evaluation mode determines whether the system is observed over a continuing time horizon or until a fixed batch is finished; the performance measures used to compare controllers are defined separately in Section~\ref{sec:evaluation}.

\paragraph{Task success.}
A product unit $z$ is successful once its terminal recipe output has been produced and delivered through an admissible output interface. The terminal output can be created only after the required recipe path has satisfied its precedence, material-consumption, and exclusive-choice rules; delivery of that output therefore suffices to certify product completion.

\paragraph{Evaluation modes.}
AssemblyGrid v1 supports two evaluation modes:
\begin{itemize}[leftmargin=*,itemsep=1pt]
\item \textbf{Continuing production}: products may continue to arrive during a fixed evaluation horizon $T$. Production rate, offered-demand service, and the fraction of admitted products completed by $T$ are therefore natural outcomes; a single makespan for the continuing stream is not defined. The horizon is an evaluation-protocol parameter and is not part of the environment dynamics. Since all nine official configurations use the same value, it is listed among the shared constants in Table~\ref{tab:abstractv1}.

\item \textbf{Finite batch}: a fixed product set $\mathcal Z$ is released according to the benchmark instance and the evaluation follows that batch  until all products are delivered or a declared finite-batch timeout $T_B$ is reached. Product completion times $C_z$ and the batch makespan $C_{\max}$ are then appropriate outcome measures for completed batches; an incomplete batch is reported as timed out together with its delivered and unfinished counts, and $C_{\max}$ is left undefined for that execution.
\end{itemize}
The per-product success condition is the same in both modes; only the evaluation boundary changes.

\paragraph{Common evaluation framework.}
AssemblyGrid v1 defines a common set of production outcomes and mechanism diagnostics for all controllers evaluated on the benchmark. Productive concurrency can improve these outcomes by increasing useful parallel activity, while the outcomes themselves remain the primary performance objectives.

\subsection{Benchmark measures}
\label{sec:evaluation}

Benchmark evaluation should address two questions: how much production was achieved, and which mechanisms enabled or limited that production; AssemblyGrid v1 therefore separates \emph{outcome measures} from \emph{mechanism diagnostics}. Delivered products and throughput are the primary continuing-production outcomes. Demand-facing companions report how many scheduled release opportunities are admitted or lost and what fraction of offered demand is ultimately delivered. Admitted-product completion, mean WIP, and restricted mean flow time are secondary outcome measures; candidate admission, coalition formation, blocking, concurrency, utilization, occupancy, and geometry quantities are mechanism diagnostics. Unless stated otherwise, $T$ denotes the evaluation horizon in ticks, with the complete metric definitions given in Appendix~\ref{app:metric-register}.

\paragraph{Outcome measures.}
Let $\mathcal O_T$ be the set of scheduled release opportunities at or before horizon $T$, let $\mathcal Z_T=\{z:a_z\le T\}$ be the product units actually admitted at those opportunities, and let $\mathcal D_T=\{z\in\mathcal Z_T:C_z\le T\}$ be those delivered at or before the horizon. In continuing production,
\begin{equation}
\mathrm{Throughput}(T)=\frac{|\mathcal D_T|}{T},
\qquad
\mathrm{Completion}_{\mathrm{adm}}(T)=\frac{|\mathcal D_T|}{|\mathcal Z_T|}\quad\text{for }|\mathcal Z_T|>0.
\end{equation}
To make admission selectivity explicit, define
\begin{equation}
\mathrm{ReleaseAcceptance}(T)=\frac{|\mathcal Z_T|}{|\mathcal O_T|},
\qquad
\mathrm{DemandFulfillment}(T)=\frac{|\mathcal D_T|}{|\mathcal O_T|}
\quad\text{for }|\mathcal O_T|>0,
\end{equation}
where $N_{\mathrm{lost}}(T)=|\mathcal O_T|-|\mathcal Z_T|$ is the number of scheduled releases not admitted. Throughput measures delivered products per tick and is the principal continuing-production measure. $\mathrm{Completion}_{\mathrm{adm}}(T)$ is explicitly the admitted-product completion fraction, not a demand-fulfillment rate. A scheduled release is admitted only when the WIP cap and raw-material availability permit it, otherwise that release opportunity is lost. A controller that clears work faster can therefore admit more products, which changes $|\mathcal Z_T|$. Because the release schedule is fixed, these quantities follow from the delivered count, the admitted-product completion fraction, and $|\mathcal O_T|$. They prevent a high admitted-product completion fraction from being interpreted as high service of offered demand. Within one official scenario, $|\mathcal O_T|$ is fixed by the common release schedule and evaluation horizon, hence demand fulfillment is a constant rescaling of delivered count and need not duplicate every controller table. Cross-scenario interpretation must additionally account for differences in release process, WIP limit, recipe structure, and workload. The same conditioning on admitted products applies to mean WIP and restricted mean flow time below. Let
\begin{equation}
\mathrm{WIP}(t)=|\mathcal Z_t\setminus\mathcal D_t|,
\qquad
\overline{\mathrm{WIP}}(T)=\frac{1}{T}\sum_{t=1}^{T}\mathrm{WIP}(t),
\end{equation}
where $\mathcal Z_t$ is the set of products admitted by decision epoch $t$ and $\mathcal D_t$ is the subset already delivered. Mean WIP is reported for continuing-production evaluations, and the time-resolved WIP trajectory may additionally be reported when transient congestion or release dynamics are under study.

To account for products that remain unfinished at the horizon, the restricted mean flow time is
\begin{equation}
\mathrm{RMFT}(T)=\frac{1}{|\mathcal Z_T|}\sum_{z\in\mathcal Z_T}\left(\widehat C_z(T)-a_z\right),
\end{equation}
where $a_z$ is the release epoch of product $z$, $C_z$ is its delivery epoch when $C_z\le T$, and $\widehat C_z(T)=\min(C_z,T)$ for a delivered product and $\widehat C_z(T)=T$ otherwise. A product released exactly at $T$ belongs to $\mathcal Z_T$ and contributes zero censored age if unfinished, while a product delivered exactly at $T$ belongs to $\mathcal D_T$. The difference $C_z-a_z$ counts elapsed environment transitions. In finite-batch mode, a completed batch has makespan $C_{\max}=\max_{z\in\mathcal Z}C_z$; timed-out incomplete batches have no reported makespan (Appendix~\ref{app:metric-register}).

\paragraph{Mechanism diagnostics.}
Let $N_{\mathrm{adm}}(T)$ be the number of execution candidates that acquire all required support and are admitted by horizon $T$, and let $N_{\mathrm{fail}}(T)$ be the number of nonempty candidate-support attempts that expire or become invalid before admission. Candidates of every arity contribute, including singleton candidates. If all recorded support for an attempt is withdrawn before admission, the attempt is treated as abandoned and is excluded from both counts; an attempt that retains nonempty support and subsequently expires or becomes invalid contributes to $N_{\mathrm{fail}}(T)$. A nonempty support attempt that is still active when the evaluation horizon is reached is right-censored: it is counted in neither $N_{\mathrm{adm}}(T)$ nor $N_{\mathrm{fail}}(T)$ and is therefore excluded from the admission-success denominator. Because abandoned and right-censored attempts are excluded, this is a resolved admission success rate and is read together with candidate switching and admission latency:
\begin{equation}
\mathrm{AdmSuccess}(T)=
\frac{N_{\mathrm{adm}}(T)}{N_{\mathrm{adm}}(T)+N_{\mathrm{fail}}(T)}
\end{equation}
where the rate is defined only when $N_{\mathrm{adm}}(T)+N_{\mathrm{fail}}(T)>0$. Admission failures per evaluation horizon are reported as $N_{\mathrm{fail}}(T)$, and admission latency is the mean number of ticks from the first recorded support for an execution candidate to its successful admission, averaged over admitted activities. The coalition success rate $\mathrm{CoalSuccess}(T)$, coalition-formation failures, and coalition-formation latency are the same quantities restricted to candidates with $|U_c|\ge2$. Every operation in the official \textsc{Coalition} recipes requires at least two robots, thus in that family the admission and coalition quantities coincide and are reported under the coalition names; \textsc{Concurrency} results report the admission quantities, and neither is reported for \textsc{Flow}, whose official recipes are entirely unary. Geometry diagnostics include geometry-blocked activities and task-level approach/retreat delay. For an admitted operation $e$, this delay is $D_{\mathrm{approach}}(e)+D_{\mathrm{retreat}}(e)$ ticks and may be accumulated or averaged as a geometry diagnostic. A blocked-activity event is counted when an individually admissible activity with a valid abstract-motion assignment fails to start at a decision epoch because an already committed activity holds a robot it needs, has exhausted a finite-capacity resource it needs, or occupies an overlapping workspace. Picks, bilateral handoffs, deliveries and operation admissions are all counted. Admissions within one epoch are arbitrated in a fixed candidate order, and the counting rule follows that arbitration: for an operation admission, a shared robot, shared exclusive resource or workspace overlap against an operation admitted earlier in the same epoch is treated as simultaneous arbitration and is not counted, whereas exhaustion of a finite-capacity resource is counted even when the exhausting activity was admitted in the same epoch. Logistics activities are counted against every committed activity, including reservations created earlier in the same epoch. Geometry-blocked activities are the subset in which abstract workspace overlap is the only cause, and are reported separately from the robot-blocked and resource-blocked subsets. The count is per blocked activity and start attempt, not per robot pair or conflicting reservation; therefore the same activity blocked over several epochs can contribute more than one event. Additional diagnostics are listed in Appendix~\ref{app:metric-register}.

The remaining diagnostics used explicitly in Section~\ref{sec:evalresults} are defined as follows. Let $B_t$ be the number of robots busy in persistent physical or logistics activity at sampled decision epoch $t$, and let $\mathcal T_{\mathrm{samp}}$ be the set of sampled ticks. For the canonical evaluations reported here, $\mathcal T_{\mathrm{samp}}=\{0,\ldots,T-1\}$, with these state diagnostics are sampled once per evaluation tick; any alternative diagnostic subsampling must be declared explicitly. Robot occupancy is
\begin{equation}
\mathrm{Occ}=\frac{1}{|\mathcal T_{\mathrm{samp}}|}\sum_{t\in\mathcal T_{\mathrm{samp}}}\frac{B_t}{MN},
\end{equation}
where $MN$ is the total number of robots; accordingly, $\mathrm{Occ}\in[0,1]$ is the time-average fraction of robots that are busy. For blocked-ready work, let $R_t$ be the set of logically ready operation samples at tick $t$, and let $F_t\subseteq R_t$ be those with at least one currently feasible coalition assignment. The ready-operation feasible-coalition fraction and blocked-ready-operation rate are
\begin{equation}
\mathrm{RF}=\frac{\sum_{t\in\mathcal T_{\mathrm{samp}}}|F_t|}{\sum_{t\in\mathcal T_{\mathrm{samp}}}|R_t|},
\qquad
\mathrm{BR}=1-\mathrm{RF},
\end{equation}
where the common denominator must be nonzero. Thus, $\mathrm{RF}$ is the fraction of ready-operation samples for which at least one feasible coalition is available, whereas $\mathrm{BR}$ is the fraction for which none is available. These quantities describe access to a feasible coalition when work is ready; they are not measures of reward, controller failure, or production success. A controller can therefore sustain productive output even when $\mathrm{BR}$ is high, as temporary infeasibility may affect some ready-operation occurrences while a feasible subset continues to progress.

Finally, define a no-progress indicator for tick $t$ to be one when WIP is nonzero, no progress event occurs during that tick, and no recipe-operation activity, whether singleton or cooperative, remains active. A progress event is a successful pick, handoff, candidate admission, recipe-operation completion, or delivery. The reported \emph{operational no-progress-tick} diagnostic is the sum of this indicator over the evaluation horizon. The metric is therefore interpreted as an inactivity diagnostic: intentional waiting, temporary idleness, or controller choices may satisfy the condition even when a feasible continuation exists.
 Due-date tardiness, value-weighted throughput, backlog age, workload balance, and compute cost remain optional extensions or instrumentation and never replace the benchmark outcome measures.

\subsection{Productive concurrency}

Since concurrency is useful only when it advances production, AssemblyGrid v1 distinguishes robot occupancy from productive operation concurrency: a system may exhibit high occupancy while making little or no product progress. Table~\ref{tab:concurrencydistinctions} summarizes the distinct interpretations.

\begin{table}[H]
\centering
\caption{Occupancy, visited-state productive-concurrency capacity, and realized productive concurrency.}
\label{tab:concurrencydistinctions}
\small
\begin{tabularx}{\textwidth}{p{0.22\textwidth}p{0.34\textwidth}Y}
\toprule
Quantity & Question answered & Interpretation \\
\midrule
Robot occupancy & How many robots are busy? & Resource-use intensity \\
 Visited-state productive-concurrency capacity & How many mutually compatible recipe-operation activities could productively occupy the next interval from the current state? & Simultaneous productive opportunity in the state reached by the controller \\
 Realized productive concurrency & How many concurrent activities actually advance product progress? & Simultaneous useful progress \\
\bottomrule
\end{tabularx}
\end{table}

\paragraph{State-dependent concurrency capacity.}
Productive concurrency capacity is state-dependent as the set of operations that can proceed simultaneously changes as products progress, robots become occupied or available, materials move, and resource or geometric conflicts appear or disappear. The coalition-size bound applies only to a single operation and therefore does not determine this system-wide capacity. At any decision epoch, the achievable concurrency is determined by the ready operations and the combinations of them that can execute together without conflict. Because the state $s_t$ is itself reached under a controller's preceding decisions, averages of this capacity along an execution are \emph{visited-state} quantities.

\paragraph{Concurrency capacity.}
At decision epoch $t$, the canonical productive-concurrency capacity is the largest set of mutually compatible recipe-operation activities that could occupy the next execution interval while preserving work already in progress. Let $\mathcal E_t^{\mathrm{run}}$ be the committed recipe-operation activities already executing and let $\mathcal C_t^{\mathrm{new}}$ be the individually admissible execution candidates in the current state. Nonproductive logistics activities may limit which productive activities can run, but they do not count toward the concurrency capacity. Each candidate $c\in\mathcal C_t^{\mathrm{new}}$ corresponds to the prospective activity $e(c)$ that would be created if the candidate were admitted. The activities considered in the capacity calculation are therefore
\begin{equation}
\mathcal E_t=\mathcal E_t^{\mathrm{run}}\cup\{e(c):c\in\mathcal C_t^{\mathrm{new}}\}.
\end{equation}
Every member of $\mathcal E_t$ is a recipe-operation activity whose successful completion advances product progress. Let $\mathfrak I_t$ be the conflict-free subsets of $\mathcal E_t$ that contain every activity in $\mathcal E_t^{\mathrm{run}}$. The canonical count capacity and its realized utilization are therefore
\begin{equation}
K_{\mathrm{prod}}^*(s_t)=\max_{I\in\mathfrak I_t}|I|,
\qquad
U_{\mathrm{prod}}^{(0)}(t)=\frac{K_{\mathrm{prod},t}^{\pi}}{K_{\mathrm{prod}}^*(s_t)}\ \ \text{when }K_{\mathrm{prod}}^*(s_t)>0,
\end{equation}
where $K_{\mathrm{prod},t}^{\pi}$ is the number of recipe-operation activities actually executing in that interval under policy $\pi$. The statewise quantity $K_{\mathrm{prod}}^*(s_t)$ is controller-independent once $s_t$ is fixed, but its execution-average is generally controller-dependent because different controllers visit different states. Accordingly, cross-controller differences in mean capacity are interpreted descriptively. Two capacity activities are incompatible when they share an exclusive robot or resource, are alternative realizations of the same operation occurrence, or violate the active geometric constraints. Compatibility is therefore not a purely pairwise relation: a resource with finite capacity above one imposes a cardinality constraint on the whole selected set, which the reference implementation enforces against the accumulated selection. Implementation details for the capacity calculation, including exact-search limits and approximation handling, are given in Appendix~\ref{app:feasibility}. Mutually exclusive alternative routes of the same product are also incompatible, ensuring that the capacity does not count routes that cannot be admitted simultaneously.

\paragraph{Productive concurrency versus parallelism.}
\label{sec:productive}
A recipe-operation activity is \textbf{productive} when its completion advances the recipe state of at least one product. Once admitted, such an operation remains in this productive class throughout its committed approach, process, and retreat interval; these phases form one indivisible task-level operation in AssemblyGrid v1. Picking, carrying, positioning, and handoff enable production but do not themselves count as productive activities (Appendix~\ref{app:feasibility}). The utilization ratio is evaluated only when $K_{\mathrm{prod}}^*(s_t)>0$; states with no productive opportunity are excluded from the average and are not assigned a perfect score.

\subsection{Conformance checks}
\label{sec:conformance}

Conservative algorithm-independent lower bounds (robot-time, critical-path, and resource) and the eleven benchmark invariants that an implementation consistent with the benchmark must preserve in every reachable state are given in Appendix~\ref{app:formal-spec} (Appendix~\ref{app:invariants} for the invariants). Section~\ref{sec:evalresults} reports the executable checks that exercise these requirements.

\section{Benchmark validation}
\label{sec:evalresults}

This section tests whether the reference implementation realizes the mechanisms of Sections~\ref{sec:environment} and~\ref{sec:benchmarkdef} and whether the official scenarios support productive control under the benchmark rules and information structure.

\subsection{Evaluation protocol}

The evidence has three levels: conformance checks test the declared rules, controlled mechanism studies test the intended effects, and reference controllers characterize productive execution under privileged centralized, structured decentralized, and learned decentralized control. Together, these experiments establish benchmark conformance, mechanism behavior, and executability across distinct controller classes. The experiments address four validation questions.

\begin{itemize}[leftmargin=*,itemsep=2pt]
\item \textbf{Recipe and material semantics.} Does the reference implementation correctly realize material conservation, sequence, AND, XOR, joins, bilateral handoffs, and terminal delivery under the published recipe definition?
\item \textbf{Temporary coalition coordination.} Under the declared decentralized observation interface, including the support-fraction cue, do multi-robot operation requirements produce observable coalition-formation behavior and failed formation attempts?
\item \textbf{Productive concurrency.} Does the benchmark distinguish concurrent \emph{productive} work from robot busyness in a matched deterministic control?
\item \textbf{Geometric feasibility.} Does the abstract geometry layer classify reach, clearance, and simultaneous-segment conflicts as specified in deterministic controlled cases?
\end{itemize}

Conformance and mechanism checks use deterministic (fixed and reproducible) test cases together with the conformance checks included in the release. The four non-learning reference controllers are evaluated on the same 50 held-out final-test instances used for learned-policy evaluation; these are unseen realizations of the same scenario generator, not transfers to new layouts, recipes or scales,. The reported MARL campaign comprises exactly $3\times3\times3\times10=270$ training runs: independent proximal policy optimization (IPPO), multi-agent proximal policy optimization (MAPPO), and QMIX, a monotonic value-function factorisation method, on the three workload families at three difficulty levels with ten independent algorithm seeds per method and scenario. All MARL methods start from random initialization without demonstrations, behavior cloning, or pretrained weights and train for 1{,}536{,}000 environment steps. Periodic training checkpoints were not retained; accordingly, each policy is evaluated at its final training checkpoint on the 50 held-out final-test instances; no post-hoc checkpoint selection is performed, and final-test measurements are not used during training. The exact instance indices and optimization records are provided in Appendix~\ref{app:reproducibility}. Table~\ref{tab:evidencethreshold} summarizes the criteria used to assess the campaign.

\begin{table}[H]
\centering
\caption{Benchmark-validation criteria.}
\label{tab:evidencethreshold}
\small
\begin{tabularx}{\columnwidth}{p{0.31\columnwidth}Y}
\toprule
Validation criterion & Requirement \\
\midrule
Benchmark conformance & Formal invariants, recipe semantics, interfaces, and metric boundaries pass executable checks \\
Privileged productive-execution evidence & The centralized method achieves positive mean deliveries in all nine official scenarios and at least one delivery on every final-test instance of the three \textsc{Coalition} scenarios using global information and centralized action selection \\
Structured decentralized control & Greedy and Parallel-aware achieve productive execution through decentralized benchmark interfaces on the held-out final-test set \\
Learned decentralized control & Representative MARL controllers are trained through the common decentralized execution interface with ten independent seeds per scenario and held-out final-test evaluation \\
Mechanism evidence & Controlled evidence exercises recipe/material, coalition, productive-concurrency, and geometry semantics; workload-level coalition results use the declared support-fraction cue \\
Reproducibility record & Benchmark version, benchmark-instance definition, random seeds, information access, budget, reward definition, and abstraction level are recorded \\
\bottomrule
\end{tabularx}
\end{table}

The non-learning controllers serve different purposes. Random masked provides a simple feasible-action baseline, Greedy aims to assess if basic decentralized execution is possible, Parallel-aware favors actions that support concurrent progress, and the Centralized controller uses global information to show what can be achieved with centralized decisions. These controllers draw on common ideas from multi-robot task allocation, decentralized allocation, and dispatching \citep{gerkey2004mrta,dias2006market,nunes2017taxonomy}. Table~\ref{tab:controllerdefs} summarizes their roles, available information, and selection rules.

\begin{table}[H]
\centering
\caption{Non-learning validation references.}
\label{tab:controllerdefs}
\small
\begin{tabularx}{\textwidth}{p{0.17\textwidth}p{0.13\textwidth}YY}
\toprule
Reference & Information & Task and operation selection & Routing and tie handling \\
\midrule
Random masked & local & uniform sample from the currently legal per-agent masked actions & seeded random choice \\
Greedy \citep{gerkey2004mrta,nunes2017taxonomy} & local & continue the pending candidate unless another enabled candidate has strictly higher visible support; otherwise select the highest-support enabled candidate & move material only along routes that reduce distance to the deterministic staging or output target \\
Parallel-aware \citep{dias2006market,nunes2017taxonomy} & local & same operation logic as Greedy & among equally progressing routes, prefer the destination with lower locally observed congestion \\
Centralized & privileged & select a compatible set from the globally admissible execution candidates using weighted independent-set selection & robots outside selected coalitions use the local parallel-aware logistics rule \\
\bottomrule
\end{tabularx}
\end{table}

The three MARL methods share the same decentralized execution information $I_t^u$ and action masks. IPPO uses a local actor and a local observation-based critic \citep{dewitt2020ippo,schulman2017ppo}. MAPPO uses the same local actor interface and a privileged agent-conditioned global-state critic during training \citep{yu2022mappo}. QMIX uses decentralized local action values combined through a monotonic global-state-conditioned mixer during training \citep{rashid2018qmix}. MAPPO and QMIX therefore use centralized information only during training, whereas the Centralized method uses privileged information for action selection during evaluation. All three MARL methods use the same training budget and the same team-level reward terms, with method-specific handling of the support credit described below; detailed network, optimizer, replay, and update settings are reported in Appendix~\ref{app:reproducibility}, Table~\ref{tab:marlconfig}.

Training rewards belong to the algorithm layer and do not enter the benchmark measures. The three MARL methods use the same team-level reward terms, with method-specific handling of the support credit. The team reward is
\begin{equation}
\begin{aligned}
r_t^{\mathrm{team}}={}&
0.10\,\Delta p_t
+2.0\,\Delta \mathrm{coal}_t
+3.0\,\Delta \mathrm{op}_t
+11.0\,\Delta \mathrm{deliv}_t \\
&-0.25\,\Delta \mathrm{fail}_t
-0.05\,\Delta \mathrm{churn}_t
-0.001,
\end{aligned}
\label{eq:trainingreward}
\end{equation}
where $\Delta p_t$ counts progress events (successful picks, handoffs, candidate admissions, operation completions, and deliveries), $\Delta\mathrm{coal}_t$ counts candidate admissions, which are coalition formations when $|U_c|\ge2$, $\Delta\mathrm{op}_t$ counts completed operations, $\Delta\mathrm{deliv}_t$ counts deliveries, $\Delta\mathrm{fail}_t$ counts failed admissions, $\Delta\mathrm{churn}_t$ counts candidate switching, and every agent receives the same $r_t^{\mathrm{team}}$. Let $c_{t,u}^{\mathrm{sup}}\in\{0,1\}$ equal one when robot $u$ held no pending support before the step and holds pending support after it, so that a switch between candidates and a support admitted within the same tick both give zero. The training rewards are then
\begin{equation}
r_{t,u}^{\mathrm{IPPO/MAPPO}}=r_t^{\mathrm{team}}+0.05\,c_{t,u}^{\mathrm{sup}},\qquad
r_t^{\mathrm{QMIX}}=r_t^{\mathrm{team}}+0.05\,\frac{1}{|\mathcal U|}\sum_{u\in\mathcal U}c_{t,u}^{\mathrm{sup}},
\end{equation}
where the support credit is agent-specific for IPPO and MAPPO, and a shared team quantity for QMIX. For IPPO and MAPPO it is therefore agent-specific reward shaping; it is small relative to the delivery coefficient and does not change the benchmark objective. The delivery coefficient is the sum of two implementation terms: the algorithm-layer environment reward, which contributes $1.0$ per delivered product since every official recipe declares unit product value and that profile assigns zero weight to its tardiness, motion-energy, safety, and WIP penalty terms, plus a $10.0$ milestone term added by the training runner. Delivery has the largest weight; admission and support terms are smaller and cease to accrue once an activity is committed. The reward is used for policy optimization, while evaluation follows the benchmark measures defined in Section~\ref{sec:evaluation}. The MARL results are therefore conditional on this reward specification.

Each independently trained policy is treated as one MARL-reference replicate. Each policy contributes one mean over its 50 final-test instances, and MARL-reference summaries and their uncertainty are computed across the ten independent training replicates for each method and scenario ($n=10$). By contrast, uncertainty for the deterministic non-learning methods is computed across the 50 benchmark instances, not across training seeds; these two uncertainty sources are therefore reported and interpreted separately, following the empirical-practice recommendations of Henderson et al. \citep{henderson2018matters}, Colas et al. \citep{colas2018seeds}, and Agarwal et al. \citep{agarwal2021precipice}. All valid MARL runs, including zero-delivery outcomes, are retained in the aggregate summaries. Runtime or infrastructure failures are reported separately from valid algorithmic outcomes.

\subsection{Validation of benchmark mechanisms}

\label{sec:mechanismevidence}

The mechanism-level experiments connect the formal benchmark definition to observable benchmark behavior before the workload results are interpreted. They proceed from deterministic recipe and material checks to temporary coalition coordination, then to a matched productive-concurrency control and deterministic geometry cases.

\medskip\noindent\textbf{Recipe and material semantics.}
The conformance evaluation examines whether the implementation preserves the benchmark semantics defined in the specification. The checks cover recipe execution and material provenance, bilateral transfer and exclusivity, coalition formation, information and action restrictions, concurrent execution, geometry constraints, and the separation of training rewards from benchmark reporting. Controlled recipe traces further verify serial, AND, XOR, and joining structures, including correct handoffs and terminal product formation only after a valid recipe path is completed. Appendix Table~\ref{tab:invariant-check-map} links the main benchmark invariants to their corresponding executable checks.

\medskip\noindent\textbf{Temporary coalition formation.}

Deterministic controls realize two-, three-, and four-robot coalitions and preserve pairwise locality, exact matching of the selected execution candidate, the canonical role assignment, and immutable membership during execution. As an interface-feasibility check, all 20 trials at each tested coalition size form in one tick both with and without the aggregate support-fraction cue. These controls show that the coalition mechanism itself does not require the cue in the constructed cases. They do not establish the workload-level causal effect of removing the cue under candidate competition.

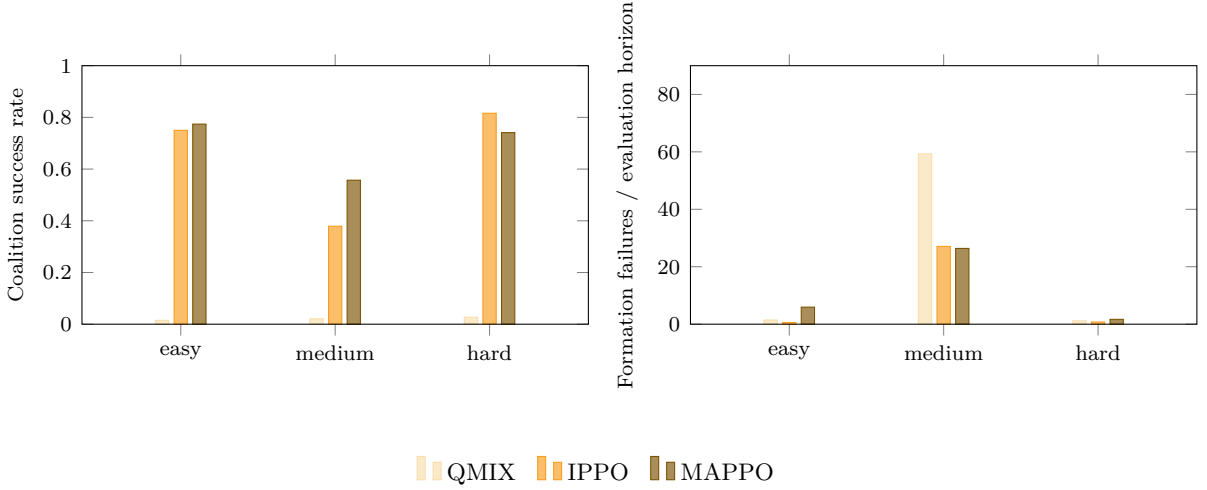
\begin{figure}[H]
\centering
\begin{tikzpicture}
\begin{groupplot}[
  group style={group size=2 by 1, horizontal sep=1.35cm},
  width=0.52\linewidth, height=5.0cm,
  every axis/.append style={ybar,bar width=5pt,clip=true,axis on top},
  symbolic x coords={easy, medium, hard}, xtick=data,
  enlarge x limits=0.32, ymin=0,
  scaled y ticks=false,
  yticklabel style={/pgf/number format/fixed, font=\scriptsize},
  tick label style={font=\scriptsize}, label style={font=\scriptsize},
]
\nextgroupplot[
  ymax=1,
  ylabel={Coalition success rate},
  legend to name=coaldiaglegend,
  legend columns=3,
  legend style={draw=none, font=\footnotesize, /tikz/every even column/.append style={column sep=6pt}}
]
\addplot+[ag/bar/qmix] coordinates {(easy,0.015) (medium,0.021) (hard,0.027)};
\addplot+[ag/bar/ippo] coordinates {(easy,0.750) (medium,0.379) (hard,0.816)};
\addplot+[ag/bar/mappo] coordinates {(easy,0.774) (medium,0.557) (hard,0.741)};
\legend{QMIX, IPPO, MAPPO}

\nextgroupplot[
  ymax=90,
  ylabel={Formation failures / evaluation horizon}
]
\addplot+[ag/bar/qmix] coordinates {(easy,1.45) (medium,59.31) (hard,1.14)};
\addplot+[ag/bar/ippo] coordinates {(easy,0.59) (medium,27.09) (hard,0.75)};
\addplot+[ag/bar/mappo] coordinates {(easy,5.93) (medium,26.38) (hard,1.69)};
\end{groupplot}
\node[anchor=north, yshift=-0.55cm] at (current bounding box.south) {\pgfplotslegendfromname{coaldiaglegend}};
\end{tikzpicture}
\caption{Coalition-formation diagnostics across difficulty levels.}
\label{fig:coaldiag}
\end{figure}

The official \textsc{Coalition} scenarios characterize temporary multi-robot coordination under the canonical AssemblyGrid v1 observation interface, which includes the support-fraction cue. Figure~\ref{fig:coaldiag} therefore reports the two MARL-method diagnostics most directly tied to coalition formation: success rate and failed formation attempts. The broader MARL-reference tables retain a common diagnostic core across workload families, thereby enabling blocking, inactivity, production, and robot use to be compared under one reporting convention; coalition-specific quantities are emphasized only where the corresponding recipes contain multi-robot operations. 

The mechanism diagnostics indicate that coalition behavior differs materially across the three scenarios under the declared support-fraction observation. The \textsc{medium} scenario produces many more failed formation attempts for IPPO and MAPPO than \textsc{easy} or \textsc{hard}, whereas formation success is not monotone in coalition arity. In particular, the WIP cap of one in \textsc{hard} reduces candidate contention despite its fixed four-robot terminal operation, and \textsc{medium} permits a variable two-to-four-robot second operation. QMIX exhibits very low coalition success throughout. 

\medskip\noindent\textbf{Productive concurrency.}

A deterministic matched control checks the productive-concurrency semantics independently of controller learning. The operation multiset, durations, precedence, release time, and total productive work are fixed. The serial schedule executes two independent alignment operations consecutively and completes at $C_{\max}=11$, whereas the productive-parallel schedule allows those same independent alignment operations to overlap and completes at $C_{\max}=8$, a reduction of 3 ticks or 27.3\%. Because the only change in this control is whether the independent productive operations may overlap, the result verifies that the benchmark distinguishes useful simultaneous progress from robot busyness. It does not by itself establish that the three official \textsc{Concurrency} scenarios isolate concurrency as a single causal factor.

\medskip\noindent\textbf{Geometric feasibility.}

The abstract geometry model is checked through three deterministic cases: a two-robot coalition targeting the midpoint between its members is accepted with minimum reach margin $1.0$ and realized duration $D_{\mathrm{approach}}+D_{\mathrm{process}}+D_{\mathrm{retreat}}=4+3+1=8$ ticks. A singleton target at distance $2.000$ against a reach of $1.500$ is rejected. Two crossing nominal approach segments are reported as conflicting, whereas two separated segments are not. Together, the deterministic cases demonstrate that the abstract geometry layer classifies the declared task-level reach and conflict conditions as specified.

\medskip\noindent\textbf{Matched geometry sensitivity.}
\label{sec:geometry-sensitivity}

A matched intervention then evaluates whether the geometry layer changes the production problem in addition to classifying isolated geometric cases. The production configuration is fixed to a $5\times6$ grid with a two-robot supported-insertion recipe, WIP cap 4, product release every 5 ticks, and a 400-tick evaluation horizon. For each of 50 random seeds, the same base production realization is evaluated under three matched geometry variants, $\rho_{\mathrm{conf}}\in\{0.12,0.25,0.40\}$, with the controller and all remaining settings held fixed. Since the geometry profile is instance-defining, these are matched variants. Increasing $\rho_{\mathrm{conf}}$ enlarges the keep-clear region used for simultaneous abstract motions and therefore changes workspace contention under the declared task-level geometry model.

The graded response for Greedy is reported in Table~\ref{tab:geometry-sensitivity}, where the capacity and realized-concurrency entries are arithmetic means of $K_{\mathrm{prod}}^*(s_t)$ and $K_{\mathrm{prod},t}^{\pi}$ over the states visited at sampled decision epochs. The utilization entry is the mean of $K_{\mathrm{prod},t}^{\pi}/K_{\mathrm{prod}}^*(s_t)$ over exact opportunity states with positive productive-concurrency capacity, consistent with Section~\ref{sec:productive}. Because the controller is fixed within this matched intervention, changes in its mean visited-state capacity are interpreted across geometry variants for that controller, not as a controller-independent scenario constant. As $\rho_{\mathrm{conf}}$ rises, geometry-blocked activities increase while productive-concurrency capacity, realized concurrency, deliveries, admitted-product completion, and throughput decline; blocked activities of all causes exceed the geometry-blocked count by about two resource-blocked events per episode, with no robot-blocked events. Utilization stays nearly constant, indicating that the controller keeps using most of the opportunity that remains.
The matched geometry sensitivity values below summarize Greedy over 50 matched base production realizations, generated from suite indices 0--49 of the matched control suite and therefore distinct from the final-test indices 1000--1049; the surrounding analysis reports the corresponding endpoint changes for the other structured references.

\begin{table}[H]
\centering
\caption{Matched geometry sensitivity.}
\label{tab:geometry-sensitivity}
\small
\begin{tabular}{lrrr}
\toprule
Metric & $\rho_{\mathrm{conf}}=0.12$ & $\rho_{\mathrm{conf}}=0.25$ & $\rho_{\mathrm{conf}}=0.40$ \\
\midrule
Delivered products / evaluation horizon & 15.20 & 14.60 & 13.94 \\
Admitted-product completion & 0.792 & 0.784 & 0.776 \\
Throughput & 0.038 & 0.036 & 0.035 \\
Geometry-blocked activities & 267 & 307 & 336 \\
Blocked activities, all causes & 269 & 309 & 338 \\
Mean visited-state concurrency capacity & 0.828 & 0.807 & 0.774 \\
Mean realized productive concurrency & 0.766 & 0.740 & 0.707 \\
Concurrency utilization & 0.927 & 0.923 & 0.916 \\
Robot occupancy & 0.189 & 0.183 & 0.175 \\
\bottomrule
\end{tabular}
\end{table}

The paired endpoint comparison demonstrates the same mechanism across the three structured controllers. From $\rho_{\mathrm{conf}}=0.12$ to $0.40$, delivered products decrease on average by $1.26$, $1.18$, and $1.22$ for Greedy, Parallel-aware, and Centralized, respectively. Realized productive concurrency decreases by $0.060$, $0.037$, and $0.068$, whereas geometry-blocked activities increase by 69, 70, and 75. All paired 95\% intervals over the 50 seeds exclude zero: deliveries $[-1.80,-0.72]$, $[-1.73,-0.63]$ and $[-1.78,-0.67]$; realized productive concurrency $[-0.086,-0.034]$, $[-0.067,-0.007]$ and $[-0.097,-0.039]$; and geometry-blocked activities $[53,85]$, $[52,87]$ and $[60,90]$. The direction is repeated across most matched seed realizations: delivered products decline in 33, 35, and 31 of 50 seed realizations; realized productive concurrency declines in 38, 37, and 39; and geometry-blocked activities increase in 47, 45, and 48. Random masked exhibits no comparable monotonic delivery response ($1.22$, $1.32$, $1.16$) and only a small endpoint increase in geometry-blocked activities (18 to 23), consistent with its limited coordinated parallel activity.

The matched intervention supports a mechanism chain in which stronger workspace interference reduces the productive concurrency permitted by the geometry layer, realized concurrency falls with that opportunity, and fewer products are completed within the fixed horizon. The nearly stable utilization ratio separates this loss of geometric opportunity from a large deterioration in how the structured controllers exploit the remaining opportunity.

\subsection{Reference behavior across the official workload suite}
\label{sec:results}

The deterministic checks above demonstrate that the benchmark mechanisms are executable, after which the controller references provide complementary evidence that the official scenarios admit productive control. Table~\ref{tab:central-solvability} reports the privileged Centralized method separately to establish attainable productive execution under global information. The MARL campaign then tests the decentralized benchmark interface across all nine scenarios under one common protocol, while Greedy and Parallel-aware provide structured decentralized executability references on the same 50 held-out final-test instances. Table~\ref{tab:nonlearning} retains the detailed decentralized non-learning delivery record for \textsc{Coalition}, where temporary multi-robot formation is the primary workload mechanism; Random masked supplies a deliberately weak interface reference.

The privileged centralized reference produced a positive mean output in every official scenario over the 50 held-out final-test instances; Table~\ref{tab:central-solvability} reports the mean delivered products per evaluation horizon and is used only as evidence of centralized productive execution. Its delivery means equal those of Parallel-aware in eight of the nine scenarios. Centralized selects operation candidates globally but uses the Parallel-aware local rule for picks, handoffs, deliveries and idling, causing the two trajectories to coincide whenever the global selection admits the same operations as the local rule. They differ only in \textsc{Coalition}-medium, the one official scenario with a variable-arity operation, where global selection can choose a different coalition size.

\begin{table}[H]
\centering
\caption{Privileged Centralized method.}
\label{tab:central-solvability}
\small
\begin{tabular}{lccc}
\toprule
Family & \textsc{easy} & \textsc{medium} & \textsc{hard} \\
\midrule
\textsc{Flow} & 21.28 & 27.68 & 38.20 \\
\textsc{Coalition} & 17.68 & 15.46 & 3.86 \\
\textsc{Concurrency} & 10.06 & 17.74 & 18.50 \\
\bottomrule
\end{tabular}
\end{table}
\noindent A positive mean delivery in every official scenario demonstrates productive centralized execution under the declared benchmark dynamics.

\begin{table}[H]
\centering
\caption{Non-learning \textsc{Coalition} deliveries.}
\label{tab:nonlearning}
\small
\begin{tabular}{lccc}
\toprule
Reference & \textsc{easy} & \textsc{medium} & \textsc{hard} \\
\midrule
Random masked   & $1.28\pm0.26$ & $1.26\pm0.24$ & $0.00\pm0.00$ \\
Greedy          & $17.34\pm0.64$ & $14.76\pm0.54$ & $3.88\pm0.41$ \\
Parallel-aware  & $17.68\pm0.65$ & $14.86\pm0.39$ & $3.86\pm0.40$ \\
\bottomrule
\end{tabular}
\end{table}
\noindent Values are mean delivered products per evaluation horizon with 95\% normal-approximation interval half-widths $1.96\,s/\sqrt{50}$, where $s$ is the sample standard deviation across the 50 benchmark instances; these instance-level intervals are distinct from the MARL-reference uncertainty computed across ten independently trained policies.

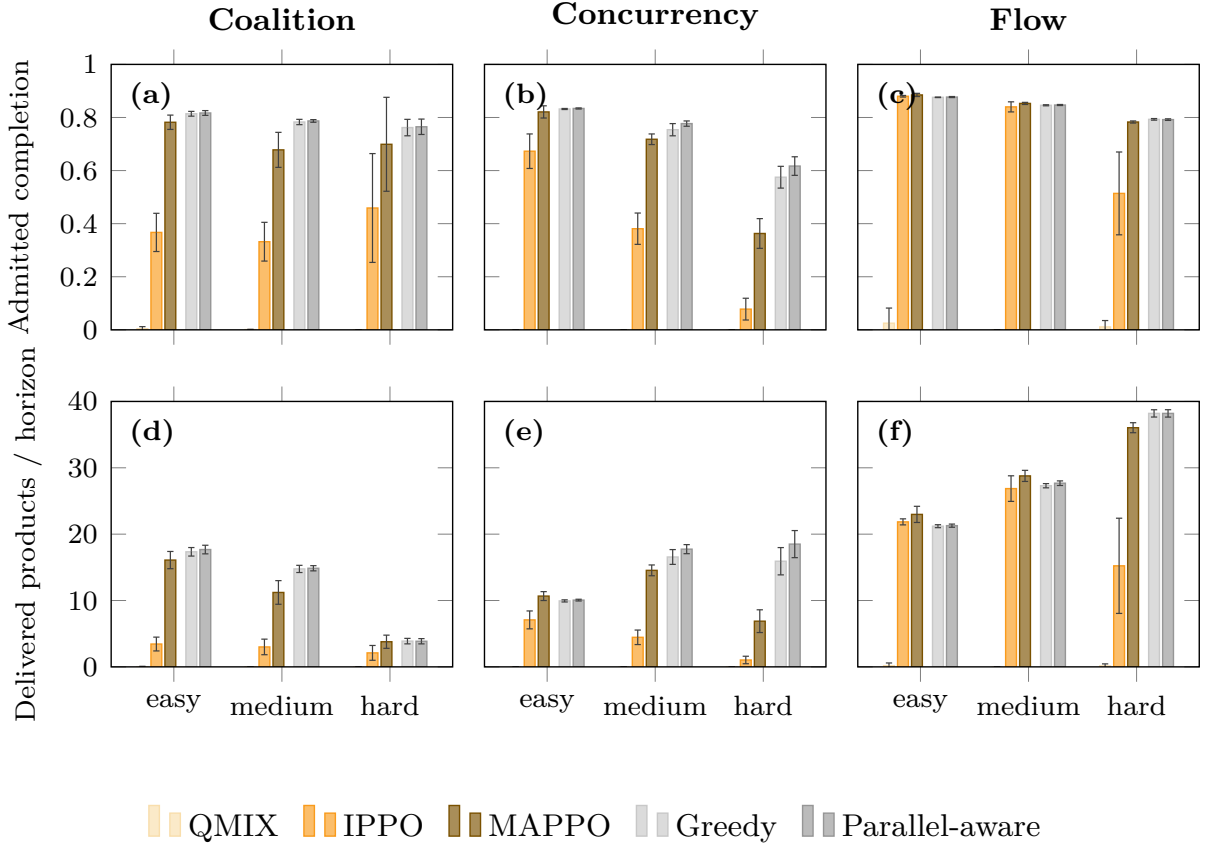
\begin{figure}[H]
\centering
\resizebox{\textwidth}{!}{%
\begin{tikzpicture}
\begin{groupplot}[
  group style={group size=3 by 2, horizontal sep=0.32cm, vertical sep=0.72cm, xticklabels at=edge bottom},
  width=0.315\textwidth, height=4.25cm,
  every axis/.append style={ybar,bar width=3.1pt,clip=true,axis on top},
  symbolic x coords={easy, medium, hard}, xtick=data,
  enlarge x limits=0.29, ymin=0,
  scaled y ticks=false,
  yticklabel style={/pgf/number format/fixed, font=\scriptsize},
  tick label style={font=\scriptsize}, label style={font=\scriptsize}, title style={font=\footnotesize\bfseries},
]
\nextgroupplot[ymax=1, ytick={0,0.2,0.4,0.6,0.8,1.0}, title={\textsc{Coalition}}, ylabel={Admitted completion},
  legend to name=familyoutcomelegend, legend columns=5,
  legend style={draw=none, font=\footnotesize, /tikz/every even column/.append style={column sep=5pt}}]
\addplot+[ag/bar/qmix,bar shift=-9pt, error bars/.cd, y dir=both, y explicit, error bar style={line width=0.35pt, black!75}, error mark options={rotate=90, mark size=0.9pt, line width=0.35pt, black!75}] coordinates {(easy,0.004) +- (0,0.008) (medium,0.001) +- (0,0.001) (hard,0.000) +- (0,0.000)};
\addplot+[ag/bar/ippo,bar shift=-5pt, error bars/.cd, y dir=both, y explicit, error bar style={line width=0.35pt, black!75}, error mark options={rotate=90, mark size=0.9pt, line width=0.35pt, black!75}] coordinates {(easy,0.367) +- (0,0.072) (medium,0.332) +- (0,0.073) (hard,0.459) +- (0,0.205)};
\addplot+[ag/bar/mappo,bar shift=-1pt, error bars/.cd, y dir=both, y explicit, error bar style={line width=0.35pt, black!75}, error mark options={rotate=90, mark size=0.9pt, line width=0.35pt, black!75}] coordinates {(easy,0.782) +- (0,0.027) (medium,0.678) +- (0,0.066) (hard,0.699) +- (0,0.177)};
\addplot+[ag/bar/greedy,bar shift=5pt, error bars/.cd, y dir=both, y explicit, error bar style={line width=0.35pt, black!75}, error mark options={rotate=90, mark size=0.9pt, line width=0.35pt, black!75}] coordinates {(easy,0.814) +- (0,0.009) (medium,0.783) +- (0,0.010) (hard,0.762) +- (0,0.031)};
\addplot+[ag/bar/parallel,bar shift=9pt, error bars/.cd, y dir=both, y explicit, error bar style={line width=0.35pt, black!75}, error mark options={rotate=90, mark size=0.9pt, line width=0.35pt, black!75}] coordinates {(easy,0.817) +- (0,0.009) (medium,0.787) +- (0,0.005) (hard,0.765) +- (0,0.029)};
\legend{QMIX, IPPO, MAPPO, Greedy, Parallel-aware}
\nextgroupplot[ymax=1, ytick={0,0.2,0.4,0.6,0.8,1.0}, yticklabels=\empty, title={\textsc{Concurrency}}]
\addplot+[ag/bar/qmix,bar shift=-9pt, error bars/.cd, y dir=both, y explicit, error bar style={line width=0.35pt, black!75}, error mark options={rotate=90, mark size=0.9pt, line width=0.35pt, black!75}] coordinates {(easy,0.000) +- (0,0.000) (medium,0.000) +- (0,0.000) (hard,0.000) +- (0,0.000)};
\addplot+[ag/bar/ippo,bar shift=-5pt, error bars/.cd, y dir=both, y explicit, error bar style={line width=0.35pt, black!75}, error mark options={rotate=90, mark size=0.9pt, line width=0.35pt, black!75}] coordinates {(easy,0.673) +- (0,0.065) (medium,0.381) +- (0,0.059) (hard,0.078) +- (0,0.041)};
\addplot+[ag/bar/mappo,bar shift=-1pt, error bars/.cd, y dir=both, y explicit, error bar style={line width=0.35pt, black!75}, error mark options={rotate=90, mark size=0.9pt, line width=0.35pt, black!75}] coordinates {(easy,0.821) +- (0,0.023) (medium,0.718) +- (0,0.020) (hard,0.363) +- (0,0.056)};
\addplot+[ag/bar/greedy,bar shift=5pt, error bars/.cd, y dir=both, y explicit, error bar style={line width=0.35pt, black!75}, error mark options={rotate=90, mark size=0.9pt, line width=0.35pt, black!75}] coordinates {(easy,0.832) +- (0,0.002) (medium,0.754) +- (0,0.023) (hard,0.575) +- (0,0.041)};
\addplot+[ag/bar/parallel,bar shift=9pt, error bars/.cd, y dir=both, y explicit, error bar style={line width=0.35pt, black!75}, error mark options={rotate=90, mark size=0.9pt, line width=0.35pt, black!75}] coordinates {(easy,0.834) +- (0,0.002) (medium,0.777) +- (0,0.010) (hard,0.617) +- (0,0.035)};
\nextgroupplot[ymax=1, ytick={0,0.2,0.4,0.6,0.8,1.0}, yticklabels=\empty, title={\textsc{Flow}}]
\addplot+[ag/bar/qmix,bar shift=-9pt, error bars/.cd, y dir=both, y explicit, error bar style={line width=0.35pt, black!75}, error mark options={rotate=90, mark size=0.9pt, line width=0.35pt, black!75}] coordinates {(easy,0.025) +- (0,0.057) (medium,0.000) +- (0,0.000) (hard,0.011) +- (0,0.024)};
\addplot+[ag/bar/ippo,bar shift=-5pt, error bars/.cd, y dir=both, y explicit, error bar style={line width=0.35pt, black!75}, error mark options={rotate=90, mark size=0.9pt, line width=0.35pt, black!75}] coordinates {(easy,0.880) +- (0,0.003) (medium,0.840) +- (0,0.019) (hard,0.514) +- (0,0.156)};
\addplot+[ag/bar/mappo,bar shift=-1pt, error bars/.cd, y dir=both, y explicit, error bar style={line width=0.35pt, black!75}, error mark options={rotate=90, mark size=0.9pt, line width=0.35pt, black!75}] coordinates {(easy,0.885) +- (0,0.006) (medium,0.853) +- (0,0.004) (hard,0.783) +- (0,0.004)};
\addplot+[ag/bar/greedy,bar shift=5pt, error bars/.cd, y dir=both, y explicit, error bar style={line width=0.35pt, black!75}, error mark options={rotate=90, mark size=0.9pt, line width=0.35pt, black!75}] coordinates {(easy,0.876) +- (0,0.001) (medium,0.846) +- (0,0.002) (hard,0.793) +- (0,0.003)};
\addplot+[ag/bar/parallel,bar shift=9pt, error bars/.cd, y dir=both, y explicit, error bar style={line width=0.35pt, black!75}, error mark options={rotate=90, mark size=0.9pt, line width=0.35pt, black!75}] coordinates {(easy,0.877) +- (0,0.002) (medium,0.847) +- (0,0.002) (hard,0.792) +- (0,0.003)};
\nextgroupplot[ymax=40, ytick={0,10,20,30,40}, ylabel={Delivered products / horizon}]
\addplot+[ag/bar/qmix,bar shift=-9pt, error bars/.cd, y dir=both, y explicit, error bar style={line width=0.35pt, black!75}, error mark options={rotate=90, mark size=0.9pt, line width=0.35pt, black!75}] coordinates {(easy,0.026) +- (0,0.059) (medium,0.004) +- (0,0.006) (hard,0.000) +- (0,0.000)};
\addplot+[ag/bar/ippo,bar shift=-5pt, error bars/.cd, y dir=both, y explicit, error bar style={line width=0.35pt, black!75}, error mark options={rotate=90, mark size=0.9pt, line width=0.35pt, black!75}] coordinates {(easy,3.442) +- (0,1.040) (medium,3.004) +- (0,1.173) (hard,2.108) +- (0,1.117)};
\addplot+[ag/bar/mappo,bar shift=-1pt, error bars/.cd, y dir=both, y explicit, error bar style={line width=0.35pt, black!75}, error mark options={rotate=90, mark size=0.9pt, line width=0.35pt, black!75}] coordinates {(easy,16.094) +- (0,1.298) (medium,11.214) +- (0,1.782) (hard,3.786) +- (0,0.994)};
\addplot+[ag/bar/greedy,bar shift=5pt, error bars/.cd, y dir=both, y explicit, error bar style={line width=0.35pt, black!75}, error mark options={rotate=90, mark size=0.9pt, line width=0.35pt, black!75}] coordinates {(easy,17.340) +- (0,0.642) (medium,14.760) +- (0,0.542) (hard,3.880) +- (0,0.414)};
\addplot+[ag/bar/parallel,bar shift=9pt, error bars/.cd, y dir=both, y explicit, error bar style={line width=0.35pt, black!75}, error mark options={rotate=90, mark size=0.9pt, line width=0.35pt, black!75}] coordinates {(easy,17.680) +- (0,0.649) (medium,14.860) +- (0,0.392) (hard,3.860) +- (0,0.396)};
\nextgroupplot[ymax=40, ytick={0,10,20,30,40}, yticklabels=\empty]
\addplot+[ag/bar/qmix,bar shift=-9pt, error bars/.cd, y dir=both, y explicit, error bar style={line width=0.35pt, black!75}, error mark options={rotate=90, mark size=0.9pt, line width=0.35pt, black!75}] coordinates {(easy,0.000) +- (0,0.000) (medium,0.000) +- (0,0.000) (hard,0.002) +- (0,0.005)};
\addplot+[ag/bar/ippo,bar shift=-5pt, error bars/.cd, y dir=both, y explicit, error bar style={line width=0.35pt, black!75}, error mark options={rotate=90, mark size=0.9pt, line width=0.35pt, black!75}] coordinates {(easy,7.078) +- (0,1.349) (medium,4.454) +- (0,1.094) (hard,1.036) +- (0,0.565)};
\addplot+[ag/bar/mappo,bar shift=-1pt, error bars/.cd, y dir=both, y explicit, error bar style={line width=0.35pt, black!75}, error mark options={rotate=90, mark size=0.9pt, line width=0.35pt, black!75}] coordinates {(easy,10.668) +- (0,0.670) (medium,14.544) +- (0,0.813) (hard,6.886) +- (0,1.714)};
\addplot+[ag/bar/greedy,bar shift=5pt, error bars/.cd, y dir=both, y explicit, error bar style={line width=0.35pt, black!75}, error mark options={rotate=90, mark size=0.9pt, line width=0.35pt, black!75}] coordinates {(easy,9.940) +- (0,0.162) (medium,16.560) +- (0,1.115) (hard,15.920) +- (0,2.057)};
\addplot+[ag/bar/parallel,bar shift=9pt, error bars/.cd, y dir=both, y explicit, error bar style={line width=0.35pt, black!75}, error mark options={rotate=90, mark size=0.9pt, line width=0.35pt, black!75}] coordinates {(easy,10.060) +- (0,0.130) (medium,17.740) +- (0,0.683) (hard,18.500) +- (0,2.042)};
\nextgroupplot[ymax=40, ytick={0,10,20,30,40}, yticklabels=\empty]
\addplot+[ag/bar/qmix,bar shift=-9pt, error bars/.cd, y dir=both, y explicit, error bar style={line width=0.35pt, black!75}, error mark options={rotate=90, mark size=0.9pt, line width=0.35pt, black!75}] coordinates {(easy,0.180) +- (0,0.407) (medium,0.000) +- (0,0.000) (hard,0.138) +- (0,0.312)};
\addplot+[ag/bar/ippo,bar shift=-5pt, error bars/.cd, y dir=both, y explicit, error bar style={line width=0.35pt, black!75}, error mark options={rotate=90, mark size=0.9pt, line width=0.35pt, black!75}] coordinates {(easy,21.852) +- (0,0.455) (medium,26.866) +- (0,1.929) (hard,15.228) +- (0,7.174)};
\addplot+[ag/bar/mappo,bar shift=-1pt, error bars/.cd, y dir=both, y explicit, error bar style={line width=0.35pt, black!75}, error mark options={rotate=90, mark size=0.9pt, line width=0.35pt, black!75}] coordinates {(easy,22.978) +- (0,1.219) (medium,28.780) +- (0,0.837) (hard,36.042) +- (0,0.753)};
\addplot+[ag/bar/greedy,bar shift=5pt, error bars/.cd, y dir=both, y explicit, error bar style={line width=0.35pt, black!75}, error mark options={rotate=90, mark size=0.9pt, line width=0.35pt, black!75}] coordinates {(easy,21.200) +- (0,0.224) (medium,27.300) +- (0,0.318) (hard,38.200) +- (0,0.549)};
\addplot+[ag/bar/parallel,bar shift=9pt, error bars/.cd, y dir=both, y explicit, error bar style={line width=0.35pt, black!75}, error mark options={rotate=90, mark size=0.9pt, line width=0.35pt, black!75}] coordinates {(easy,21.280) +- (0,0.238) (medium,27.680) +- (0,0.347) (hard,38.200) +- (0,0.557)};
\end{groupplot}
\foreach \p/\lab in {c1r1/a,c2r1/b,c3r1/c,c1r2/d,c2r2/e,c3r2/f}
  \node[anchor=north west,font=\scriptsize\bfseries] at ($(group \p.north west)+(0.05cm,-0.04cm)$) {(\lab)};
\node[anchor=north, yshift=-0.50cm] at (current bounding box.south) {\pgfplotslegendfromname{familyoutcomelegend}};
\end{tikzpicture}%
}
\caption{Decentralized reference behavior across workload families.}
\label{fig:familyoutcomes}
\end{figure}

Figure~\ref{fig:familyoutcomes} shows the three workload families. Each MARL-reference bar is the mean of ten independently trained policies evaluated on the common 50-instance final-test set, whereas Greedy and Parallel-aware summarize those benchmark instances directly and therefore have a different uncertainty source. The privileged Centralized method is reported separately in Table~\ref{tab:central-solvability}, since its purpose is to demonstrate attainable centralized execution. Tables~\ref{tab:coalcoord-full}--\ref{tab:coalcoord-flow}, the training and optimization figures, and Table~\ref{tab:checkpoint-heldout} provide complementary benchmark and learning diagnostics.

In Figure~\ref{fig:familyoutcomes}, whiskers show 95\% intervals, across the ten training replicates (Student-$t$) for the MARL references and across the 50 final-test instances for Greedy and Parallel-aware. The three workload families produce distinct benchmark signatures under the common interfaces. In \textsc{Coalition}, productive execution requires repeated temporary coalition formation; the easy, variable-arity medium, and fixed four-robot hard configurations yield different formation, blocking, and delivery patterns under the declared support-fraction observation. In \textsc{Flow}, productive MARL execution is obtained across all three levels by the PPO-based methods while the scenario progression jointly changes serial recipe depth, WIP, release pressure, and, at the hard level, grid size. In \textsc{Concurrency}, increasing WIP and release frequency exposes more simultaneously ready work and changes blocked-ready behavior and realized overlap. These workload-family observations are descriptive because the official difficulty levels are complete scenarios with co-varying parameters; causal attribution to any one changed factor requires a matched intervention such as the geometry study above.

These results are interpreted as evidence about the benchmark rather than about the relative merit of the controller algorithms. The structured decentralized methods demonstrate that each workload family supports productive execution under simple task-level rules, while the MARL methods demonstrate that the same local observation and action space can produce nontrivial learned behavior across the official suite. Delivered count and admitted-product completion are read together: delivered count measures absolute production within the fixed horizon, whereas admitted-product completion conditions on the controller-dependent admitted population. The demand-facing definitions in Section~\ref{sec:evaluation} make scheduled, admitted, and lost releases explicit so that a high completion fraction cannot be mistaken for high service of offered demand. Numerical differences among MARL methods remain descriptive consequences of the fixed reference configurations and are not used to infer an algorithm ordering.

Tables~\ref{tab:coalcoord-full}--\ref{tab:coalcoord-flow} complement the outcome figures with a common diagnostic core covering blocking, operational no-progress behavior, production, and robot use. Coalition-formation diagnostics remain central for \textsc{Coalition}. In \textsc{Concurrency}, whose recipe combines two unary branches with a pairwise join, the same counts include singleton admissions and are therefore reported as admission diagnostics (Section~\ref{sec:evaluation}); they are omitted from the \textsc{Flow} table, as the official \textsc{Flow} recipes contain only unary operations. Table~\ref{tab:coalcoord-concurrency} also reports concurrency diagnostics computed as in Section~\ref{sec:geometry-sensitivity}: mean visited-state productive-concurrency capacity and mean realized productive concurrency are per-tick means over the states visited by each controller, whereas utilization is the mean ratio over exact-capacity opportunity states and therefore need not equal the ratio of the two means. Because the visited states differ among controllers, mean capacity is not interpreted as a controller-independent property of a scenario. QMIX utilization is undefined in runs without such a state and averages the runs in which it is defined (7, 7, and 5 of 10 for \textsc{easy}, \textsc{medium}, and \textsc{hard}). Capacity was solved exactly at every sampled state except in IPPO \textsc{hard} and QMIX \textsc{medium}, where 49 and 112 of the 500 final-test episodes contained approximate-capacity states. The ready-operation feasible-coalition fraction is also omitted since Section~\ref{sec:evaluation} defines it exactly as $1-\mathrm{BR}$. MARL-reference uncertainty is computed across independent training replicates, whereas non-learning uncertainty arises from benchmark-instance variation; the two are therefore interpreted separately.

\begin{table}[htbp]
\centering
\caption{\textsc{Coalition} MARL-reference diagnostics.}
\label{tab:coalcoord-full}
\scriptsize
\setlength{\tabcolsep}{5.0pt}
\begin{tabular}{llrrr}
\toprule
Metric & Difficulty & QMIX & IPPO & MAPPO \\
\midrule
Coalition-formation failures / horizon & easy   & 1.45 & 0.59 & 5.93 \\
 & medium & 59.31 & 27.09 & 26.38 \\
 & hard   & 1.14 & 0.75 & 1.69 \\
\addlinespace
Coalition success (resolved) & easy   & 0.015 & 0.750 & 0.774 \\
 & medium & 0.021 & 0.379 & 0.557 \\
 & hard   & 0.027 & 0.816 & 0.741 \\
\addlinespace
Candidate switches / horizon & easy   & 0.00 & 0.00 & 0.00 \\
 & medium & 225.63 & 4.39 & 14.76 \\
 & hard   & 0.00 & 0.19 & 0.00 \\
\addlinespace
Admission latency, ticks & easy   & 0.56$^{\dagger}$ & 0.16 & 0.41 \\
 & medium & 0.71$^{\dagger}$ & 0.30 & 0.63 \\
 & hard   & 0.53$^{\dagger}$ & 0.11 & 0.05$^{\dagger}$ \\
\addlinespace
Blocked-ready-operation rate & easy   & 0.114 & 0.828 & 0.934 \\
 & medium & 0.192 & 0.733 & 0.814 \\
 & hard   & 0.176 & 0.830 & 0.762 \\
\addlinespace
Operational no-progress ticks, of 400 & easy   & 396.1 & 327.5 & 158.8 \\
 & medium & 379.2 & 316.6 & 164.8 \\
 & hard   & 398.3 & 240.2 & 219.2 \\
\addlinespace
Delivered / horizon & easy   & 0.03 & 3.44 & 16.09 \\
 & medium & 0.00 & 3.00 & 11.21 \\
 & hard   & 0.00 & 2.11 & 3.79 \\
\addlinespace
Admitted-product completion & easy   & 0.004 & 0.367 & 0.782 \\
 & medium & 0.001 & 0.332 & 0.678 \\
 & hard   & 0.000 & 0.459 & 0.699 \\
\addlinespace
Robot occupancy & easy   & 0.003 & 0.035 & 0.158 \\
 & medium & 0.009 & 0.036 & 0.128 \\
 & hard   & 0.001 & 0.036 & 0.050 \\
\bottomrule
\end{tabular}
\par\smallskip
\parbox{0.62\textwidth}{\footnotesize $^{\dagger}$Latency is averaged over replicates with at least one admission: QMIX 1, 7 and 1 of 10 in easy, medium and hard, and MAPPO 9 of 10 in hard.}
\end{table}

\begin{table}[htbp]
\centering
\caption{\textsc{Concurrency} MARL-reference diagnostics.}
\label{tab:coalcoord-concurrency}
\scriptsize
\setlength{\tabcolsep}{5.0pt}
\begin{tabular}{llrrr}
\toprule
Metric & Difficulty & QMIX & IPPO & MAPPO \\
\midrule
Admission failures / horizon & easy   & 1.94 & 1.07 & 0.48 \\
 & medium & 0.00 & 5.06 & 10.34 \\
 & hard   & 3.19 & 2.58 & 7.06 \\
\addlinespace
Admission success & easy   & 0.298 & 0.950 & 0.987 \\
 & medium & 0.015 & 0.868 & 0.848 \\
 & hard   & 0.235 & 0.866 & 0.860 \\
\addlinespace
Blocked-ready-operation rate & easy   & 0.100 & 0.612 & 0.483 \\
 & medium & 0.013 & 0.859 & 0.884 \\
 & hard   & 0.156 & 0.904 & 0.973 \\
\addlinespace
Operational no-progress ticks, of 400 & easy   & 389.3 & 211.7 & 139.9 \\
 & medium & 366.7 & 277.7 & 92.8 \\
 & hard   & 384.1 & 319.2 & 190.9 \\
\addlinespace
Delivered / horizon & easy   & 0.00 & 7.08 & 10.67 \\
 & medium & 0.00 & 4.45 & 14.54 \\
 & hard   & 0.00 & 1.04 & 6.89 \\
\addlinespace
Admitted-product completion & easy   & 0.000 & 0.673 & 0.821 \\
 & medium & 0.000 & 0.381 & 0.718 \\
 & hard   & 0.000 & 0.078 & 0.363 \\
\addlinespace
Robot occupancy & easy   & 0.003 & 0.073 & 0.104 \\
 & medium & 0.010 & 0.052 & 0.174 \\
 & hard   & 0.004 & 0.020 & 0.087 \\
\addlinespace
Mean visited-state concurrency capacity & easy   & 0.919 & 0.524 & 0.727 \\
 & medium & 3.155 & 0.495 & 1.114 \\
 & hard   & 0.734 & 0.580 & 0.776 \\
\addlinespace
Mean realized productive concurrency & easy   & 0.020 & 0.472 & 0.699 \\
 & medium & 0.000 & 0.371 & 1.091 \\
 & hard   & 0.057 & 0.284 & 0.766 \\
\addlinespace
Concurrency utilization & easy   & 0.145 & 0.916 & 0.965 \\
 & medium & 0.001 & 0.875 & 0.978 \\
 & hard   & 0.202 & 0.854 & 0.988 \\
\bottomrule
\end{tabular}

{\footnotesize Capacity is exact on 99.6\% or more of the evaluated states in every cell except QMIX \textsc{Concurrency}-medium (81.0\%); on the remaining states the search exceeds its exact limits and the reported capacity is a greedy lower bound. Utilization is averaged over exact states only.\par}
\end{table}

\begin{table}[htbp]
\centering
\caption{\textsc{Flow} MARL-reference diagnostics.}
\label{tab:coalcoord-flow}
\scriptsize
\setlength{\tabcolsep}{5.0pt}
\begin{tabular}{llrrr}
\toprule
Metric & Difficulty & QMIX & IPPO & MAPPO \\
\midrule
Blocked-ready-operation rate & easy   & 0.015 & 0.006 & 0.018 \\
 & medium & 0.003 & 0.055 & 0.049 \\
 & hard   & 0.027 & 0.054 & 0.038 \\
\addlinespace
Operational no-progress ticks, of 400 & easy   & 394.5 & 156.9 & 147.8 \\
 & medium & 392.6 & 72.4 & 56.3 \\
 & hard   & 373.8 & 150.1 & 21.9 \\
\addlinespace
Delivered / horizon & easy   & 0.18 & 21.85 & 22.98 \\
 & medium & 0.00 & 26.87 & 28.78 \\
 & hard   & 0.14 & 15.23 & 36.04 \\
\addlinespace
Admitted-product completion & easy   & 0.025 & 0.880 & 0.885 \\
 & medium & 0.000 & 0.840 & 0.853 \\
 & hard   & 0.011 & 0.514 & 0.783 \\
\addlinespace
Robot occupancy & easy   & 0.005 & 0.193 & 0.189 \\
 & medium & 0.002 & 0.194 & 0.194 \\
 & hard   & 0.009 & 0.108 & 0.236 \\
\bottomrule
\end{tabular}
\end{table}

\subsection{MARL-reference behavior and same-generator held-out evaluation}

\begin{figure}[H]
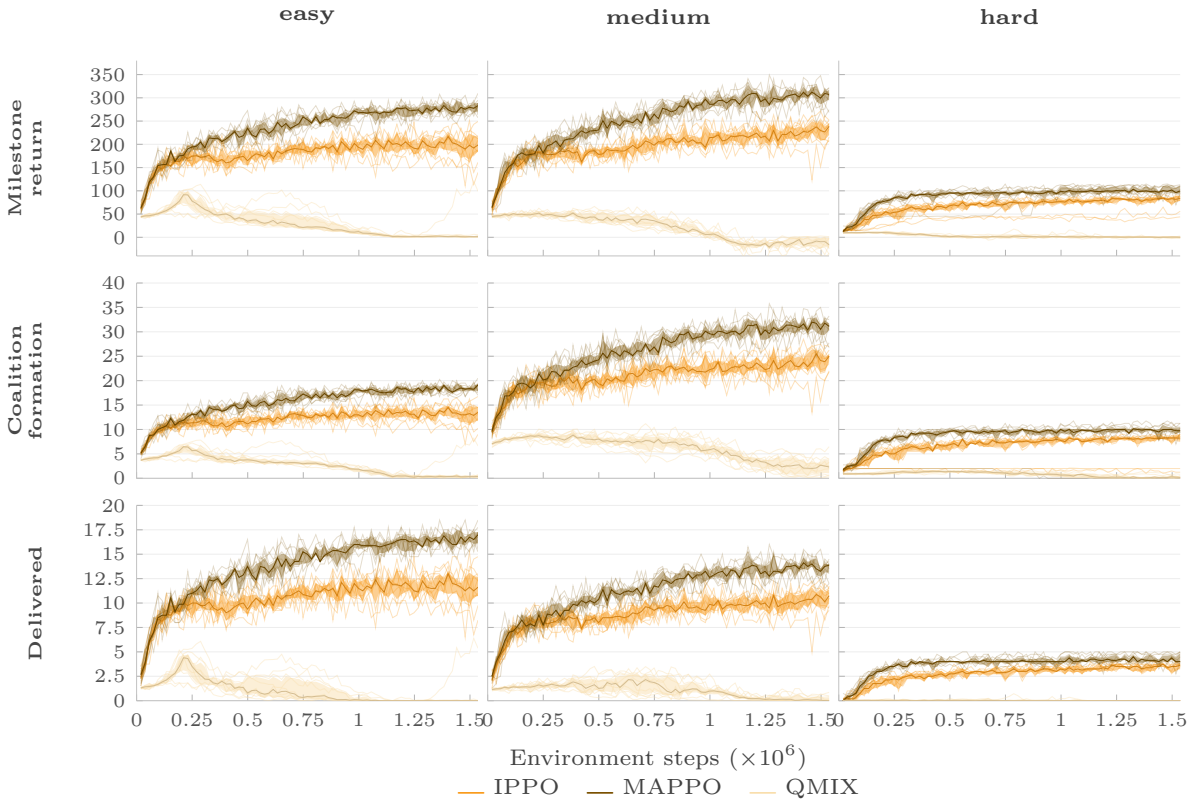

\centering
\resizebox{\textwidth}{!}{%
%
}
\caption{\textsc{Coalition} training curves.}
\label{fig:traincurves}
\end{figure}

\begin{figure}[H]
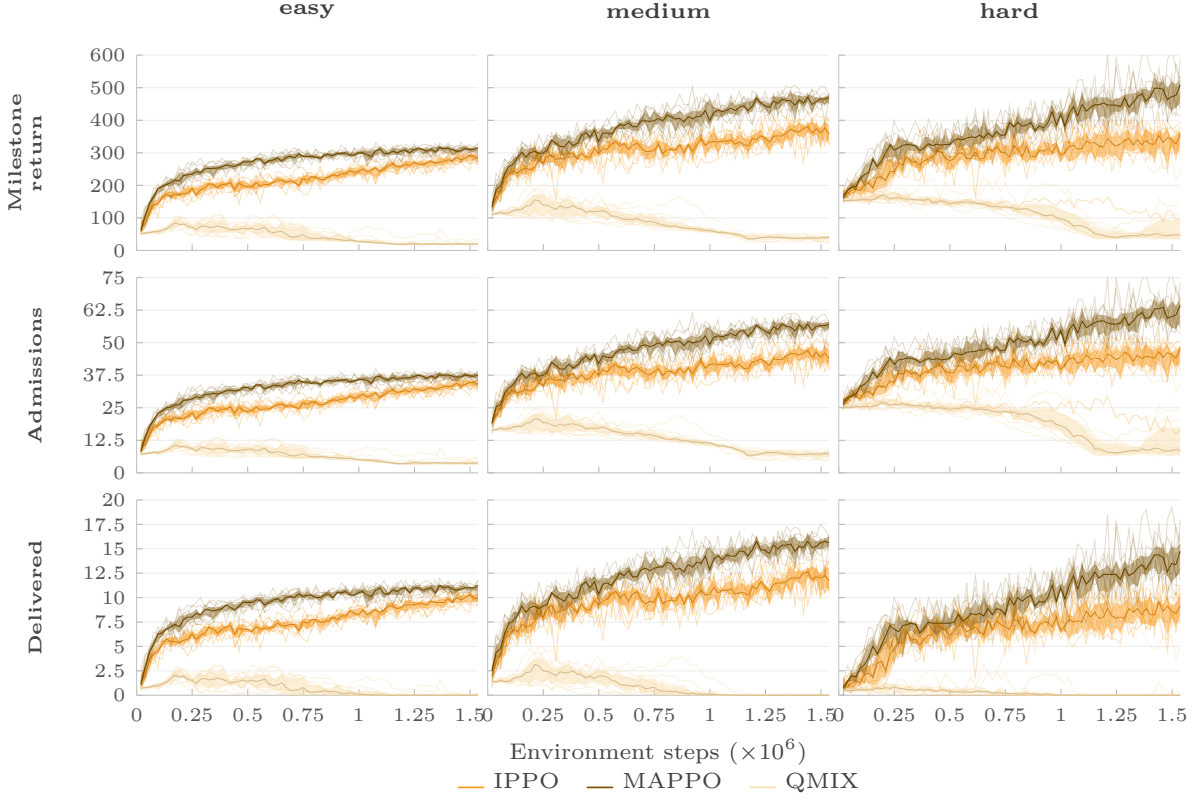

\centering
\resizebox{\textwidth}{!}{%
%
%
}
\caption{\textsc{Concurrency} training curves.}
\label{fig:traincurves-concurrency}
\end{figure}

\begin{figure}[H]
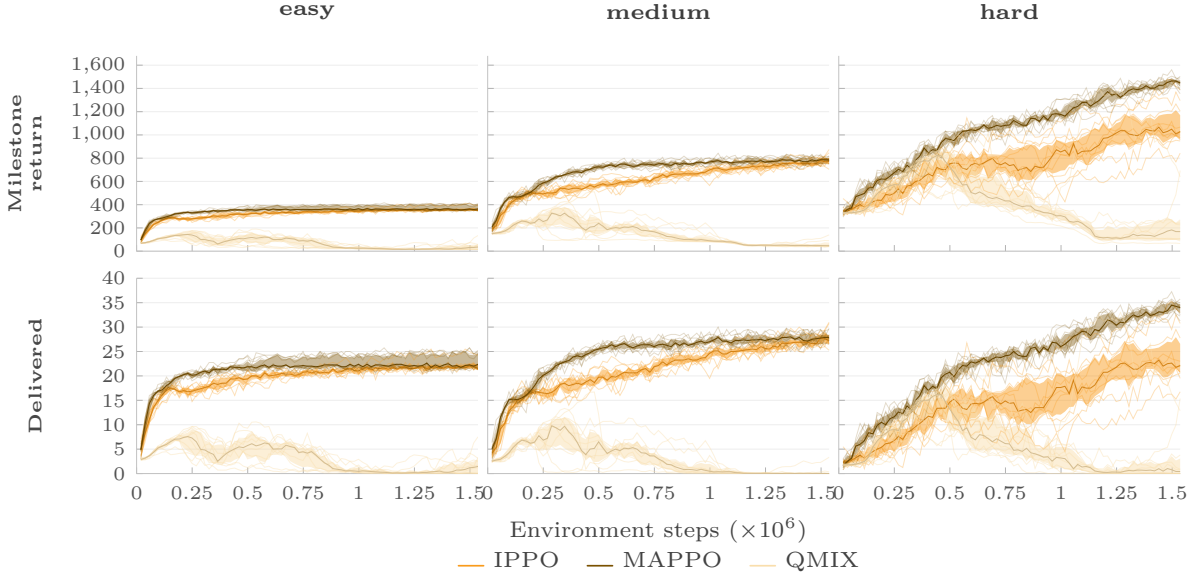

\centering
\resizebox{\textwidth}{!}{%
%
%
}
\caption{\textsc{Flow} training curves.}
\label{fig:traincurves-flow}
\end{figure}

\begin{figure}[H]
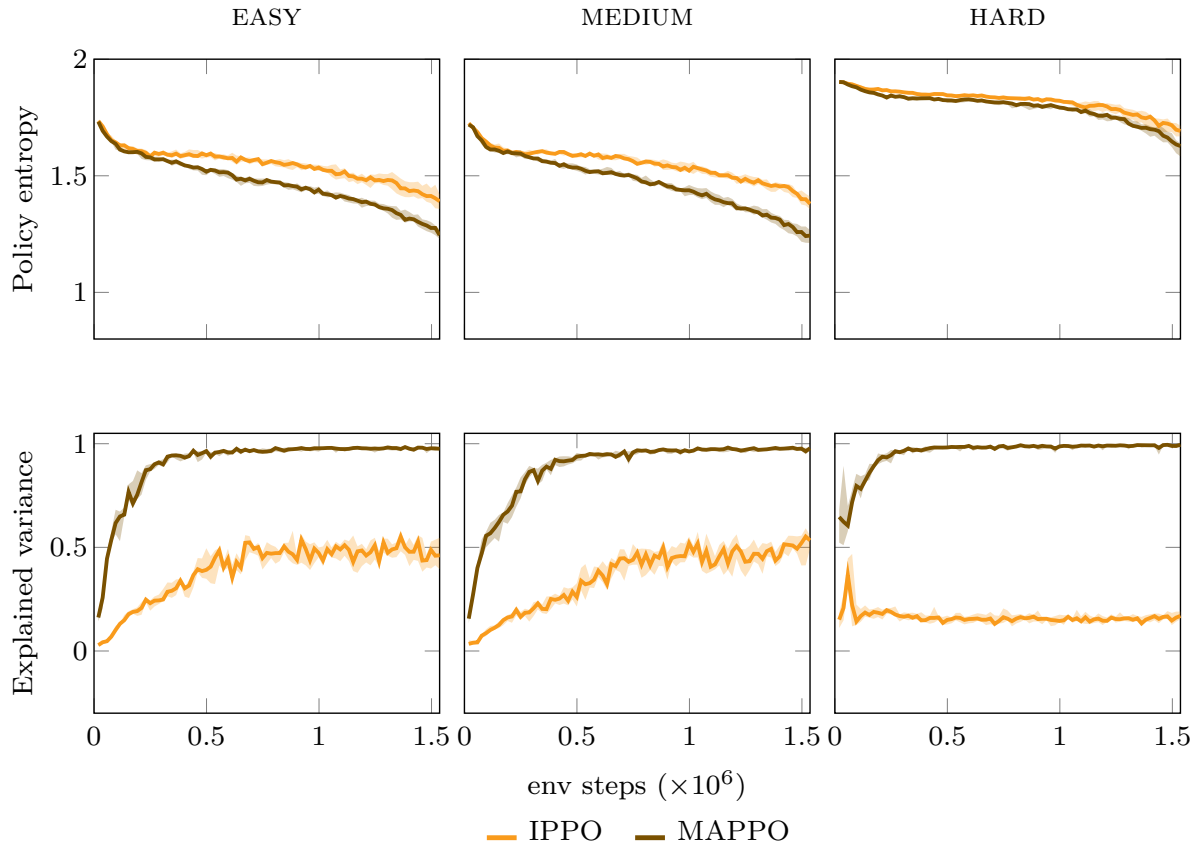

\centering
\resizebox{\textwidth}{!}{%
%
}
\caption{\textsc{Coalition} actor-critic diagnostics.}
\label{fig:trainopt}
\end{figure}

\begin{figure}[H]
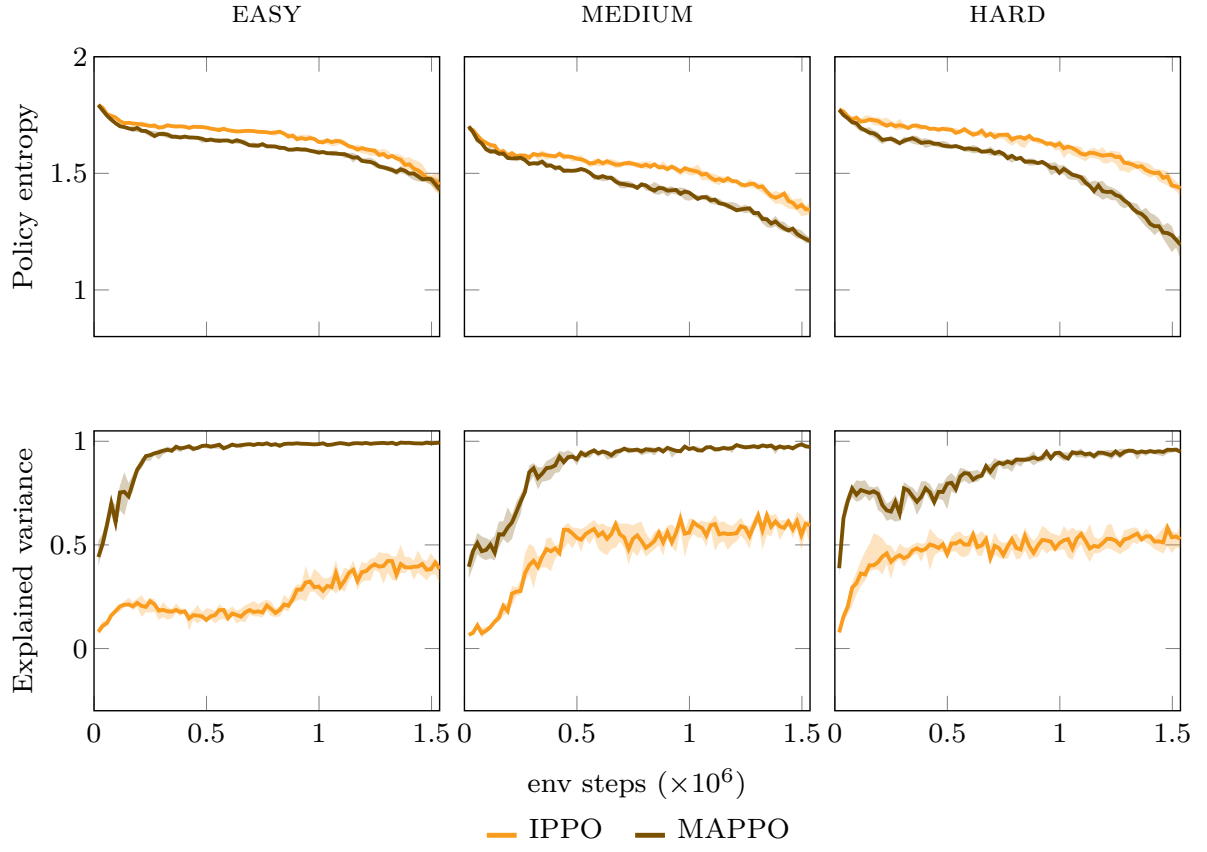

\centering
\resizebox{\textwidth}{!}{%
%
%
}
\caption{\textsc{Concurrency} actor-critic diagnostics.}
\label{fig:trainopt-concurrency}
\end{figure}

\begin{figure}[H]
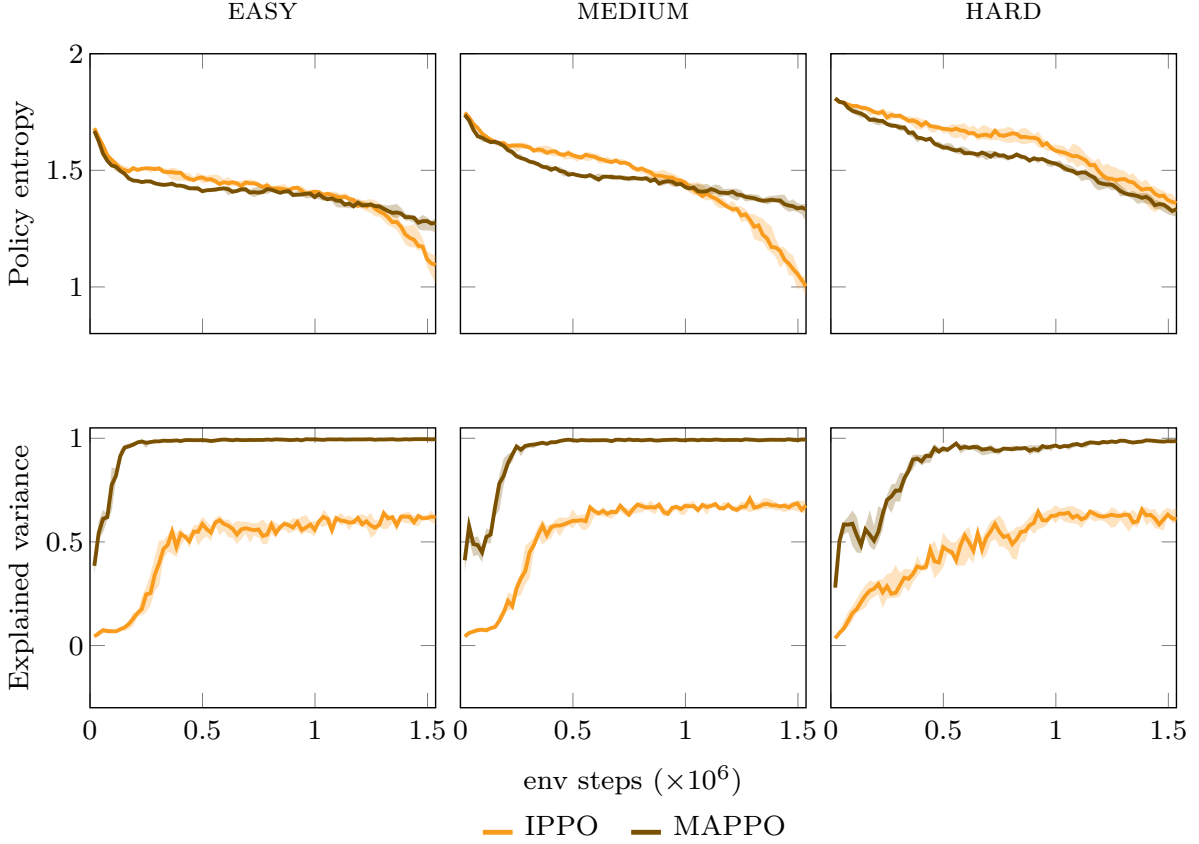

\centering
\resizebox{\textwidth}{!}{%
%
%
}
\caption{\textsc{Flow} actor-critic diagnostics.}
\label{fig:trainopt-flow}
\end{figure}

The training record provides optimization context for the full 270-run MARL campaign over 1.536 million environment steps. Here \emph{milestone return} denotes the logged mean undiscounted milestone return over the rollouts contributing to a recorded training milestone, where each rollout return sums the team terms of Eq.~\ref{eq:trainingreward} with a delivery coefficient of $10$ together with the agent-mean support credit, and excludes the separate \texttt{sparse-delivery-v1} environment reward that supplies the remaining delivery unit; it is an optimization diagnostic not a benchmark score. Figures~\ref{fig:traincurves} and~\ref{fig:traincurves-concurrency} report the ten-replicate median and interquartile range for milestone return, admissions, and delivery in the \textsc{Coalition} and \textsc{Concurrency} families, whose recipes contain multi-robot operations; admissions are coalition formations in \textsc{Coalition} and also include singleton admissions in \textsc{Concurrency}. Figure~\ref{fig:traincurves-flow} reports milestone return and delivery for \textsc{Flow}; coalition-formation curves are intentionally omitted given the unary structure of the official \textsc{Flow} recipes. Figures~\ref{fig:trainopt}, \ref{fig:trainopt-concurrency}, and~\ref{fig:trainopt-flow} report actor-critic optimization diagnostics for IPPO and MAPPO, allowing policy and critic learning dynamics to be inspected independently of final production outcomes. All training figures document behavior under the fixed reference protocol and are used for reproducibility and optimization-health context.

Under the reported implementations, reward specification, exploration settings, and fixed training budget, IPPO and MAPPO develop sustained delivery behavior across all three workload families, although the final operating level remains strongly scenario-dependent. The QMIX reference exhibits transient improvement in several training traces but produces little or no final-test output across the official suite. This pattern characterizes the reported QMIX configuration, including its observation encoding, action representation, exploration schedule, replay settings, and training budget; it does not support a general claim about QMIX as an algorithm class.

Policy entropy measures the stochasticity of the decentralized actor, whereas explained variance describes the fraction of variation in the critic target captured by the corresponding value prediction. Across the nine official scenarios, the entropy trajectories of IPPO and MAPPO decrease progressively without collapsing to zero, which is consistent with increasingly concentrated masked action distributions over the visited states while retaining nonzero stochasticity over the reported training budget; since the legal action set is state dependent, raw entropy is interpreted only as an optimization diagnostic. The critic diagnostics separate more strongly: MAPPO reaches high explained variance across the scenarios, whereas IPPO remains markedly lower under its local observation-based critic. The observed distinction is consistent with the different critic-information regimes defined by the two reference implementations, although the diagnostic alone neither identifies the cause of their production differences nor demonstrates algorithmic superiority. Policy entropy and critic explained variance are reported only for IPPO and MAPPO since these actor-critic diagnostics are directly comparable in the logged training record; QMIX uses a different value-decomposition and exploration structure and is therefore characterized through the common training and outcome measures in Figures~\ref{fig:traincurves}, \ref{fig:traincurves-concurrency}, and~\ref{fig:traincurves-flow}.

Table~\ref{tab:checkpoint-heldout} consolidates the MARL outcomes for the full official suite. For every family and difficulty level, each entry summarizes ten independently trained final checkpoints, each evaluated on the same 50 held-out instances, and reports the mean delivered count with the 95\% Student-$t$ interval half-width $t_{0.975,9}\,s/\sqrt{10}$, where $s$ is the sample standard deviation of the ten per-policy means.

\begin{table}[H]
\centering
\caption{Held-out final-checkpoint deliveries.}
\label{tab:checkpoint-heldout}
\small
\setlength{\tabcolsep}{5.0pt}
\begin{tabular}{llrrr}
\toprule
Family & Difficulty & QMIX & IPPO & MAPPO \\
\midrule
\textsc{Coalition} & easy   & $0.03\pm0.06$ & $3.44\pm1.04$ & $16.09\pm1.30$ \\
 & medium & $0.00\pm0.01$ & $3.00\pm1.17$ & $11.21\pm1.78$ \\
 & hard   & $0.00\pm0.00$ & $2.11\pm1.12$ & $3.79\pm0.99$ \\
\addlinespace
\textsc{Concurrency} & easy   & $0.00\pm0.00$ & $7.08\pm1.35$ & $10.67\pm0.67$ \\
 & medium & $0.00\pm0.00$ & $4.45\pm1.09$ & $14.54\pm0.81$ \\
 & hard   & $0.00\pm0.00$ & $1.04\pm0.57$ & $6.89\pm1.71$ \\
\addlinespace
\textsc{Flow} & easy   & $0.18\pm0.41$ & $21.85\pm0.46$ & $22.98\pm1.22$ \\
 & medium & $0.00\pm0.00$ & $26.87\pm1.93$ & $28.78\pm0.84$ \\
 & hard   & $0.14\pm0.31$ & $15.23\pm7.17$ & $36.04\pm0.75$ \\
\bottomrule
\end{tabular}
\end{table}

Held-out deliveries are lower than the late training-curve values in Figures~\ref{fig:traincurves}, \ref{fig:traincurves-concurrency}, and~\ref{fig:traincurves-flow} for IPPO in \textsc{Coalition} easy and medium, \textsc{Concurrency} hard, and \textsc{Flow} hard, and for MAPPO in \textsc{Concurrency} hard. Training rollouts sample actions from the policy on run-specific training realizations, whereas final-test evaluation selects the highest-scoring legal action on the shared held-out instances. Taken together with the mechanism controls and non-learning methods, these experiments address the four validation questions summarized in Table~\ref{tab:vqsummary}.

\begin{table}[H]
\centering
\caption{Summary of validation evidence.}
\label{tab:vqsummary}
\small
\begin{tabularx}{\textwidth}{p{0.20\textwidth}p{0.30\textwidth}Y}
\toprule
Validation question & Primary evidence & Established result \\
\midrule
Recipe and material semantics & 203/203 conformance checks and deterministic recipe/material traces & The reference implementation realizes the declared recipe, material, handoff and delivery semantics. \\
Temporary coalition coordination & deterministic coalition-size controls and \textsc{Coalition} formation diagnostics & Two- to four-robot coalitions execute; under the declared support-fraction observation, the three \textsc{Coalition} scenarios produce distinct formation and failure behavior. \\
Productive concurrency & matched $C_{\max}=11$ versus $8$ control & Productive concurrent execution is represented separately from robot busyness. \\
Geometric feasibility & deterministic geometry cases and the 50-instance matched conflict-radius intervention & The geometry layer classifies task-level feasibility as specified, and stronger workspace interference reduces productive concurrency and production under matched conditions. \\
\bottomrule
\end{tabularx}
\end{table}

Taken together, the experiments verify consistency with the benchmark task definition, exercise the declared recipe, coalition, productive-concurrency, and geometry mechanisms at the specified abstraction level, and demonstrate through a matched intervention that geometric interference changes productive opportunity and production. Coalition workload results are conditional on the canonical support-fraction observation, and the official family levels are interpreted as complete workload scenarios. The structured non-learning methods demonstrate productive executability across the official scenarios, while the 270-run MARL campaign characterizes scenario-dependent production, coordination, and optimization behavior under one common protocol.

\section{Discussion}
\label{sec:discussion}

This section interprets the results reported in Section~\ref{sec:evalresults}.

\subsection{Benchmark behavior across the official workload suite}

The three workload families emphasize different workload structures; accordingly, difficulty labels are meaningful within a family but not across families. The MARL results show that the benchmark supports decentralized policy learning through its local observation and action space: IPPO and MAPPO yield positive mean final-test delivery in every official scenario without centralized action selection at execution time, although 4 of the 90 IPPO replicates and 1 of the 90 MAPPO replicates deliver nothing on the final test. The reported QMIX configuration yields little or no final-test output, with 84 of 90 replicates delivering nothing, which coincided with divergence recorded in the archived training diagnostics: the temporal-difference loss rose more than tenfold in 70 of the 90 runs, and the median absolute mixed value estimate grew by one to two orders of magnitude in every family. In the experiments, QMIX is only implemented via a single-network temporal-difference target with mean-squared-error loss.

\subsection{Productive concurrency and geometry}

The matched $C_{\max}$ control shows that allowing otherwise identical independent operations to overlap has a concrete effect on completion time. The conflict-radius intervention indicates how geometry reduces productive opportunity for a fixed controller: stronger workspace interference increases geometry-blocked activities, lowers visited-state capacity and realized concurrency, and reduces deliveries for all three structured controllers, while utilization barely changes. These matched controls support the stated mechanism interpretations. By contrast, the easy, medium, and hard \textsc{Concurrency} scenarios are not a one-factor intervention because their load parameters, and at the hard level grid size, co-vary.

\subsection{Scope and limitations}
\label{sec:scope}

The evaluation covers all three official workload families, with the MARL experiments spanning all nine scenarios. AssemblyGrid v1 works at the task level and abstracts lower-level robot details such as inverse kinematics, joint limits, dynamics, and grasp physics. Its geometry model instead captures factors such as reach, clearance, motion time, and workspace conflicts. The main assumptions are listed in Appendix~\ref{app:assumptions}. Coalition results use the standard decentralized observation, including the support-fraction cue, which is kept fixed in the current experiments. Broader settings, such as different layouts, heterogeneous robots, sensing uncertainty, and alternative coordination mechanisms, are left for future extensions.
All three MARL methods are trained from scratch for $1{,}536{,}000$ steps using the reference settings and reward defined in Section~\ref{sec:evalresults}. The reported results therefore reflect these particular training settings and budget, and different architectures, tuning, or credit-assignment methods could lead to different outcomes. Training uses five realizations per scenario, while evaluation uses 50 unseen realizations from the same generator. The results therefore show generalization within the same scenario.

\subsection{Future research enabled by AssemblyGrid v1}
\label{sec:extensions}

Beyond learning, the same environment supports controlled production-system studies in which topology, workspace overlap, routing redundancy, robot density, product mix and coalition-capable regions are varied with controller assumptions fixed. Future controller studies may extend the MARL-reference characterization with method-specific tuning, recurrent or attention-based architectures, and alternative credit assignment, or evaluate how learned MARL policies respond to declared geometry changes using the same matched conflict-radius intervention. Broader extensions should introduce one explicitly defined source of variation at a time. Examples include an explicit communication channel with specified range, bandwidth, latency, loss, and cost; standard generalization splits over recipes, geometry, product mix, teammates, and scale; disturbances such as robot downtime, blocked workspaces, stockouts, and duration noise; higher-detail geometry with robot-model kinematics and collision checking; and richer process models with rework, rollback, or scrap. Treating these as named extensions keeps the AssemblyGrid v1 workload families, information rules, feasibility conditions, and metrics stable for comparison.

\section{Conclusion}

AssemblyGrid v1 provides a reproducible task definition for decentralized multi-robot production in which explicit recipe and material progression, temporary coalitions, bounded local information that includes a declared support-fraction coordination cue, productive concurrency, and geometry-dependent admissibility interact within a common environment. The reference implementation satisfies the benchmark conformance checks and realizes the intended recipe, coalition, concurrency, and geometry mechanisms, while the matched geometry intervention supports that workspace interference reduces productive opportunity and production. Across the official suite, the centralized and structured decentralized references achieve productive execution, and IPPO and MAPPO yield positive mean final-test delivery, under distinct information and decision regimes, demonstrating the validity of the proposed benchmark. Physical robot validity, perception, alternative coordination interfaces, learned communication, robustness, and transfer are beyond the scope of this research.

\section*{Data and code availability}

The AssemblyGrid v1 reference implementation, the nine-scenario official realization catalogue (\texttt{assemblygrid-official-realizations-v4}, with the pre-v1 geometry family kept separately as a historical extension), the JSON schemas, the invariant and conformance test suite, the four non-learning reference controllers, the MARL campaign training code, including the shared actor network and the campaign runners of Table~\ref{tab:marlconfig}, and the frozen MARL campaign \texttt{aag046-official-\allowbreak suite-\allowbreak canonical-\allowbreak v1} (270 runs: IPPO, MAPPO, and QMIX $\times$ the three workload families $\times$ three difficulty levels $\times$ ten seeds, 1{,}536{,}000 environment steps each) are released under version tag \texttt{v1.0.0}. The benchmark repository contains the implementation, the catalogues, the schemas, the conformance suite, the reference controllers and the campaign training code; the frozen 270-run campaign records are deposited separately as an archived dataset whose SHA-256 manifest binds every reported table and figure value to its source record. The code is released under the MIT licence. The repository is available at \url{https://github.com/Fouad-Bahrpeyma/AssemblyGrid_v1}, and the versioned software release is archived at \url{https://doi.org/10.5281/zenodo.22734115}.

\appendix

\section{Reproducibility, versioning, and reference implementation}
\label{app:reproducibility}

Appendix~\ref{app:reproducibility} records release-level information needed to reproduce and interpret the benchmark without changing the AssemblyGrid v1 task definition. The appendix separates the abstraction-level definition, experimental archive, and software-interface details from the formal benchmark definition that follows.

\subsection{Geometry profile and release identity}

The released recipe schema is \texttt{assemblygrid-recipe-v1}, and the default geometry profile is \texttt{abstract-v1}. A dimensional mapping of the normalized reference quantities to a particular manipulator, including a UR10 interpretation, does not change the AssemblyGrid v1 task-level abstraction. Evaluations incorporating robot-model inverse kinematics, joint-limit checks, synchronized link-level collision checking, dynamics, contact, or hardware execution should be reported as higher-detail extensions.

\subsection{Release identifiers used for reproducibility}

The main paper uses semantic names instead of software identifiers. The released implementation uses the following identifiers where exact software traceability is required: candidate-support fraction \texttt{proposal\_support}, admitted-product completion fraction \texttt{completion\_rate}, admission success rate \texttt{coalition\_success\_rate}, admission failures \texttt{team\_formation\_failures}, the legacy software key \texttt{deadlock\_ticks} for the operational no-progress-tick diagnostic, the legacy aggregate blocking keys \texttt{trajectory\_conflicts} and \texttt{geometry\_conflict\_count} together with the cause-specific keys \texttt{geometry\_block\_count}, \texttt{resource\_block\_count} and \texttt{robot\_block\_count}, the archived count-capacity key \texttt{K\_feasible\_star} for the canonical $K_{\mathrm{prod}}^*$ quantity (the implementation also exports \texttt{productive\_concurrency\_capacity\_mean} as the semantic alias), Random masked reference \texttt{random\_feasible}, shared actor class \texttt{StructuredActor}, and the fixed MARL training reward \texttt{coalition\_milestone\_v1}. The semantic reward terms $\Delta\mathrm{coal}_t$ and $c_t^{\mathrm{sup}}$ correspond to the legacy implementation concepts for candidate-admission events and proposal-support credit, respectively. These identifiers are implementation details and do not alter the corresponding benchmark definitions.

The official scenario recipes are implemented under the following identifiers: \textsc{Flow} easy, medium, and hard use \texttt{solo\_inspection}, \texttt{flow\_serial\_3}, and \texttt{flow\_serial\_5}; \textsc{Coalition} easy, medium, and hard use \texttt{pair\_assembly}, \texttt{supported\_insert}, and \texttt{standard\_abc}; and all three \textsc{Concurrency} difficulty levels use \texttt{parallel\_branch\_light}, with the corresponding semantic recipe structures given in Table~\ref{tab:cellparameters}.

\subsection{Experimental archive and evaluation records}

The reported controller campaign uses five training realizations, five disjoint validation instances (indices 200--204), and 50 held-out final-test instances (indices 1000--1049) in every scenario. Within each scenario, the scenario configuration, generator, and active geometry profile are held fixed across these splits. The two evaluation splits are the canonical suite instances, generated once from the released instance index and therefore identical across every run, method, and algorithm seed. The training split is constructed differently: each run takes the scenario parameters of suite indices 0--4 and regenerates five realizations under run-specific generation and execution seeds derived from the algorithm seed, therefore every replicate trains on its own five realizations of the same scenario. The training budget is divided evenly across those five realizations, which are trained as consecutive blocks in a fixed order. A realization changes only the generation and execution seeds: within a scenario the grid, topology, recipe graph, WIP limit, release interval, geometry profile and horizon are fixed, and the generation seed determines only which additional input robots may reach each shelf, which shelf row supplies each raw material, and the product masses and due slack. Due slack does not enter v1 dynamics; product mass enters payload feasibility, which never binds under the official mass and payload ranges, and it appears in the privileged global state. Training and final-test realizations are therefore drawn from the same generator with the same scenario parameters and differ only in their seeds; accordingly, a policy always trains on the same layout and recipe and sees five distinct access and material-source patterns, each revisited over many episodes of the 1{,}536{,}000-step budget. This matched split was chosen to give each run its own reproducible five-realization training set per scenario. The same final-test instances are used for IPPO, MAPPO, QMIX, and the reported non-learning methods. Periodic training checkpoints were not retained; the reported policies are therefore evaluated at their final training checkpoint. The validation split was used only for online monitoring during training, not for early stopping, reward adjustment, post-hoc model selection, or final-test selection. The final-test figures reported in Section~\ref{sec:results} and Table~\ref{tab:checkpoint-heldout} are the per-run final-checkpoint evaluation means recorded in each run's result record, \texttt{aag042\_result.json}.

The MARL campaign contains 270 completed training runs, comprising ten independent statistical replicates for each method-scenario configuration: IPPO, MAPPO, and QMIX $\times$ the three workload families $\times$ three difficulty levels $\times$ ten independent algorithm seeds, each trained from scratch for 1{,}536{,}000 environment steps. The archived campaign identifier is \texttt{aag046-official-suite-canonical-v1}. Each valid run contributes one final checkpoint, its optimization history, and its evaluations on the validation and final-test splits. Algorithmic zero-production outcomes remain valid replicates and are retained in aggregate summaries. Pilot and tuning runs, deterministic mechanism checks, non-learning evaluations, geometry-control evaluations, infrastructure failures, incomplete runs, and reruns are not counted as additional MARL replicates and are tracked separately in the archive. Non-learning reference behavior is taken from the frozen non-learning reference evaluation record of the four reference controllers evaluated on the same instances. Earlier exploratory campaigns, protocol-development runs, and the pre-AssemblyGrid v1 geometry-family records are retained in the archive for traceability only and are excluded from every reported aggregate.

Each released instance records its complete runtime configuration and content identity. The archived evaluation package is accompanied by a SHA-256 checksum manifest that binds the released records to their archived counterparts. The archive retains the complete training histories of all 270 runs.

\subsection{Reference implementation interface}

The reference implementation exposes the decentralized AssemblyGrid v1 observation and candidate-specific action space, deterministic reset and step semantics, action masks, separate generation and execution seeds, benchmark KPI logging, and a distinct privileged global-state interface for centralized references and centralized-training methods. Candidate identity and realization follow the canonical key and realization rule of Section~\ref{sec:environment}, and candidates are generated at most once per key. A candidate identifier is the software reference to that identity, while a candidate encoding is the controller-side representation used by a particular implementation.

For candidate construction, the engine first identifies logically ready operation occurrences and their required material holders, enumerates manipulation-local robot subsets that satisfy declared arity and role capability, constructs the corresponding task-level target assignment, and rejects structurally or task-level-geometrically inadmissible realizations. For decentralized exposure, robot $u$ receives only candidates that contain $u$. The candidate list is deliberately not filtered by hidden remote active-trajectory conflicts or globally contended resources, as such filtering would leak joint feasibility through action availability. Those conflicts are resolved during capacity diagnosis or joint admission. Candidate ordering uses aggregate support and stable canonical identity. The reference implementation provides 32 pick-option slots and 64 operation-option slots per agent and raises an error on overflow. No candidate-addressability errors occurred in the official scenarios or reported experiments; the release addressability audit, which runs Parallel-aware for 180 ticks on each of the 45 official realizations, observed at most ten visible pick options per agent (limit 32), in \textsc{Concurrency}-hard, and 24 operation candidates (limit 64), in \textsc{Coalition}-medium.

Optional Gymnasium and PettingZoo adapters and algorithm-layer reward helpers may be supplied for convenience; they do not alter the underlying AssemblyGrid v1 semantics. Instance records retain the active geometry profile or declared variant and all problem-defining parameters, while AssemblyGrid v1 decentralized execution retains aggregate local candidate-support fraction without adding a free-form or addressable communication channel.

\subsection{Reference MARL configurations}

Table~\ref{tab:marlconfig} lists the MARL-controller configurations used for the reported full-suite campaign and records the method-level settings needed to reproduce the 270 reference training runs. These settings define the reference implementations used in the reported MARL characterization.

\begin{table}[H]
\centering
\caption{MARL methods' configurations.}
\label{tab:marlconfig}
\scriptsize
\begin{tabularx}{\textwidth}{p{0.19\textwidth}YYY}
\toprule
Field & IPPO & MAPPO & QMIX \\
\midrule
Execution network
& shared actor: LayerNorm, linear encoder to width 256, two $\tanh$ residual blocks, and separate action-family and action-slot heads whose logits are summed; orthogonal initialization
& same shared actor as IPPO
& shared $Q$ network: three-layer $\tanh$ multilayer perceptron (MLP) with two hidden layers of width 256 \\
\addlinespace
Training value model
& local critic on the agent's own observation; same residual trunk at width 256
& privileged agent-conditioned global critic on the concatenated global state and agent observation; same trunk
& monotonic QMIX mixer over the privileged global state, mixing width 64, hypernetwork hidden width 128, non-negative first-layer weights \\
\addlinespace
Optimizer
& AdamW ($\epsilon=10^{-5}$)
& AdamW ($\epsilon=10^{-5}$)
& AdamW ($\epsilon=10^{-5}$) \\
\addlinespace
Learning rate
& actor $10^{-4}$, critic $3\times10^{-4}$
& actor $10^{-4}$, critic $3\times10^{-4}$
& $5\times10^{-4}$ \\
\addlinespace
Discount
& $\gamma=0.99$ & $\gamma=0.99$ & $\gamma=0.99$ \\
\addlinespace
On-policy update
& generalized advantage estimation (GAE) $\lambda=0.95$; policy clip $0.10$; value clip $0.20$; 1{,}600 environment steps per update; four epochs; minibatch 4{,}096; gradient-norm clip $0.5$; target Kullback--Leibler (KL) $0.02$
& same as IPPO
& not applicable \\
\addlinespace
Exploration / regularization
& entropy coefficient annealed linearly $0.015\to0.001$ over the training budget
& same as IPPO
& $\epsilon$-greedy, $1.0\to0.05$ linearly over the first 75\% of the budget, then held at $0.05$ \\
\addlinespace
Replay / target
& not applicable & not applicable
& uniform replay buffer $50{,}000$; batch 64; one update every 4 environment steps; hard target sync every 1{,}600 steps; single-network temporal-difference (TD) target with mean-squared-error loss (neither Double $Q$ nor Huber loss was enabled) \\
\addlinespace
Warm-up
& 3{,}200-step critic warm-up (actor frozen)
& 3{,}200-step critic warm-up (actor frozen)
& 3{,}200-transition replay warm-up before the first update \\
\addlinespace
Training curriculum
& \multicolumn{3}{Y}{one declared difficulty per run; the budget is split evenly over five realizations regenerated from the scenario parameters of suite indices 0--4 under run-specific seeds} \\
\addlinespace
Initialization & \multicolumn{3}{c}{random weights; no demonstrations, behavior cloning, or pretraining} \\
\addlinespace
Training budget & \multicolumn{3}{c}{$1{,}536{,}000$ environment steps} \\
\bottomrule
\end{tabularx}
\end{table}

\subsection{Executable conformance mapping}

The benchmark invariants in Appendix~\ref{app:invariants} are linked to executable checks in the released reference implementation. The mapping makes the correspondence between the formal benchmark definition and the simulator explicitly traceable without making Python function names part of the benchmark semantics. Table~\ref{tab:invariant-check-map} gives the principal mapping used by the conformance suite reported in Section~\ref{sec:evalresults}.

\begin{table}[H]
\centering
\caption{Invariant-to-check mapping.}
\label{tab:invariant-check-map}
\scriptsize
\begin{tabularx}{\textwidth}{p{0.13\textwidth}p{0.34\textwidth}Y}
\toprule
Invariant & Benchmark property & Principal executable check \\
\midrule
Conservation & Material conservation and unique ownership & \texttt{check\_holding\_uniqueness}; duplication tests in \texttt{check\_support\_vs\_ownership} \\
Support & Support does not create a second material owner & \texttt{check\_support\_vs\_ownership} \\
Progress & Product progress is monotone across state transitions & \texttt{test\_progress\_is\_monotone}, together with the construction that completed operations are only added \\
Recipe & Recipe-state consistency & \texttt{check\_recipe\_state} \\
Locality & Coalition locality & \texttt{check\_coalitions} \\
Lifetime & Coalition membership and roles remain fixed while committed & \texttt{check\_coalition\_lifetime} \\
Arity & Coalition arity and the canonical role assignment are admissible & \texttt{check\_coalitions} \\
Exclusivity & Robot, reservation, and finite-resource exclusivity & \texttt{check\_parallelism}, \texttt{check\_reservations}, \texttt{check\_shelves} \\
Observation & Decentralized information locality & observation-constructor conformance tests \\
Masking & Action-mask admissibility does not reveal global joint feasibility & \texttt{check\_action\_mask\_respected} and encoding tests \\
Accounting & Released and delivered accounting is monotone & \texttt{check\_monotonic\_counts} \\
\bottomrule
\end{tabularx}
\end{table}

Additional implementation-consistency checks verify WIP accounting, candidate-support table consistency, and agreement between realized concurrency and the declared diagnostic capacity. The exact capacity solver is checked against hand-derived optima on small constructed cases, including a case in which greedy selection is suboptimal. For auditability, the public release record binds the reported 203-check suite to the benchmark version, an immutable source or archive identifier, the exact test command, the Python and dependency environment, and the invariant-to-test mapping in Table~\ref{tab:invariant-check-map}. These checks support reproducibility but do not define additional benchmark objectives.

\section{Formal AssemblyGrid v1 definition}
\label{app:formal-spec}

Appendix~\ref{app:formal-spec} collects the formal elements that define an AssemblyGrid v1 problem independently of any particular controller. The reference implementation distinguishes state, instance parameters, transition rules, hard constraints, structural relations, feasibility conditions, metrics, and invariants. Implementation identifiers and release information are kept in Appendix~\ref{app:reproducibility}, whereas experimental evidence remains in Section~\ref{sec:evalresults}.

\subsection{Core notation}
\label{app:notation}

\begin{longtable}{p{0.27\textwidth}p{0.66\textwidth}}
\toprule
Symbol & Meaning \\
\midrule
$u_{i,j}$ & Robot at fixed grid position $(i,j)$ \\
$\mathcal U$ & Complete robot population \\
$\mathcal N(u)$ & AssemblyGrid v1 local neighborhood of robot $u$ \\
$\mathcal W(u)$ & Abstract reachable workspace of robot $u$ \\
$\mathcal H(u,v)$ & Shared handoff region of neighboring robots $u$ and $v$ \\
$\mathcal G_{\mathrm{obs}},\mathcal G_{\mathrm{handoff}},\mathcal G_{\mathrm{manip}}$ & Local observation, direct handoff, and direct manipulation relations \\
$\Gamma_r=(\mathcal T_r,\mathcal E_r,\mathcal X_r)$ & Recipe structure of product type $r$; $(\mathcal T_r,\mathcal E_r)$ is its precedence DAG \\
$\mathcal X_r$ & Mutually exclusive operation alternatives for XOR recipe structure \\
$z$ & Product unit; $r(z)$ denotes its product type \\
$q$ & Material item, representing a part or subassembly \\
$\tau$ & Recipe operation \\
$\Delta_\tau$ & Atomic material-state transition applied when operation $\tau$ completes successfully \\
$(z,\tau)$ & Operation occurrence for product unit $z$ \\
$c$ & Execution candidate for a ready operation occurrence \\
$e$ & Activity created after an execution candidate is admitted \\
$C(e)$ & Temporary coalition committed to activity $e$ \\
$k_\tau^{\min},k_\tau^{\max}$ & Minimum and maximum coalition sizes allowed by operation $\tau$ \\
$k_{\mathrm{topo}}^{\max}$ & Largest coalition supported by the declared manipulation topology \\
$s_t$ & Global benchmark state at decision epoch $t$ \\
$p_z(t)$ & Monotone product-progress state of product $z$ \\
$G_t^{\mathrm{op}}$ & Conflict graph over ongoing and newly admissible concrete activities at decision epoch $t$ \\
$\mathcal E_t^{\mathrm{run}}$ & Activities already committed and still executing at the start of decision epoch $t$ \\
$\mathfrak I_t$ & Conflict-free activity sets that preserve all activities in $\mathcal E_t^{\mathrm{run}}$ \\
$K_{\mathrm{prod}}^*(s_t)$ &  Canonical maximum number of mutually compatible recipe-operation activities that can productively occupy that interval while preserving existing commitments \\
\bottomrule
\end{longtable}

\subsection{Entities and state variables}
\label{app:entities}

The minimum state-bearing entities are robots, product units, material items, recipe operations, execution candidates, admitted activities, temporary coalitions, and physical or resource reservations. The global state $s_t$ records robot availability and held material; part and subassembly location and ownership; recipe readiness, execution, completion, and resolved XOR choices; execution candidates and pending support records, including the lifecycle timing needed for expiry and switching eligibility; roles, activity commitments, and ongoing activities; fixtures, buffers, abstract workspaces, and other capacity-limited resources; product release, WIP, completion, and output status; and benchmark time. No separate communication-state component is included in AssemblyGrid v1, consistent with the absence of any agent-authored or addressable communication channel; only the benchmark-defined aggregate candidate-support cue is exposed.

A product carries a recipe identity and an evolving product-progress state. A robot likewise has task-level status in addition to its fixed grid position. Richer privileged or extension state may contain joint configuration, detailed robot-model features, skills, due dates, or economic values, but implementation-level storage alone does not make those variables AssemblyGrid v1 observations or instance requirements. Table~\ref{tab:fixed-semantics} lists the semantic elements held fixed across all AssemblyGrid v1 instances.

\subsection{Configuration parameters and instance identity}
\label{app:parameters}

AssemblyGrid v1 separates fixed semantic choices from scenario parameters: the fixed semantics define how observations, handoffs, coalitions, recipes, geometry, and success are interpreted, whereas the official scenarios select problem quantities such as grid size, recipe, WIP limit, and release interval. The nine scenarios in Table~\ref{tab:cellparameters}, together with the constants in Table~\ref{tab:abstractv1} and the recipe parameters in Table~\ref{tab:recipeparams}, are the complete workload definitions; no separate generic grid, WIP, release-interval, or recipe default supersedes them.

\begin{table}[H]
\centering
\caption{Fixed semantics of AssemblyGrid v1.}
\label{tab:fixed-semantics}
\small
\begin{tabularx}{\textwidth}{p{0.28\textwidth}Y}
\toprule
AssemblyGrid v1 element & Fixed interpretation \\
\midrule
Local topology & Moore radius one for the AssemblyGrid v1 observation, handoff, and manipulation relations; the relations remain semantically distinct. \\
Coalition upper bound & $k_{\mathrm{topo}}^{\max}=\omega(\mathcal G_{\mathrm{manip}})$, which equals four for the official AssemblyGrid v1 grids under the Moore-radius-one manipulation topology. \\
Geometry profile & Versioned \texttt{abstract-v1} task-level geometry. Reach, spacing, clearance, timing, and conflict constants are owned by the declared profile. \\
Official conflict setting & The nine official scenarios use normalized workspace-conflict radius $\rho_{\mathrm{conf}}=0.25$; matched geometry studies use explicitly named variants. \\
Handoff semantics & Bilateral sender and receiver commitment is required. \\
Coalition semantics & Required members independently select the same candidate-specific action; the environment does not assign teammates. \\
Support lifecycle & Each robot holds support for at most one execution candidate. Support created at $t_0$ is admission-eligible through $t_0+3$ and first expires at $t_0+4$; repeated support does not refresh the timeout or lock, and switching is permitted from $t_0+1$. \\
Information structure & Local task information plus the aggregate candidate-support fraction; supporter identities, global feasibility, and free-form or addressable messages are not exposed. \\
Objective definition & Task success and benchmark metrics are defined by AssemblyGrid v1; learning reward and discount are method-level quantities. \\
\bottomrule
\end{tabularx}
\end{table}

An instance identity records every problem-defining choice required to reproduce the dynamics: grid and topology; active geometry profile or named variant; recipe and product mix; material, buffer, fixture, and other resource settings; release process and WIP limit; operation and transfer durations; coalition and handoff rules; observation definition; initial condition; and generation seed. Evaluation mode and any finite-batch timeout belong to the evaluation-execution record. Algorithm-specific quantities such as optimizer, network architecture, training reward, learning rate, batch size, entropy coefficient, or exploration schedule are method configuration.

\begin{table}[H]
\centering
\caption{Constants shared by the official configurations.}
\label{tab:abstractv1}
\small
\begin{tabularx}{\textwidth}{p{0.20\textwidth}p{0.44\textwidth}Y}
\toprule
Group & Constant & Value \\
\midrule
Geometry & Cell spacing & 1.0 \\
 & Reference reach and robot reach radius & 1.5 \\
 & Workspace-conflict radius $\rho_{\mathrm{conf}}$ & 0.25 \\
 & Robot clearance & 0.10 \\
 & Role-target offset & 0.12 \\
 & Retreat fraction & 0.25 \\
Timing & Reference time; robot speed; joint speed & 1.0; 1.0; 1.0 \\
 & Pick, delivery, and handoff durations & 2, 2, and 1 ticks \\
Coordination & Support timeout & 3 ticks (admission-eligible through $t_0+3$) \\
 & Candidate-switching lock & 1 tick \\
Supply and access & Shelf capacity per raw material type & 6 \\
 & Shelf replenishment interval & 5 ticks \\
 & Extra shelf-access probability & 0.30 \\
 & Robot payload; product mass range & 10.0; 0.5 to 2.0 \\
Disturbances & Robot, operation, and handoff failure; duration noise; stockout & 0 (disabled) \\
Evaluation & Horizon $T$ (evaluation protocol) & 400 ticks \\
\bottomrule
\end{tabularx}
\end{table}

Table~\ref{tab:abstractv1} lists the constants shared by all nine official configurations; distances are in units of cell spacing and durations in ticks. The extra shelf-access probability determines, once per instance and from the generation seed, which additional input robots may pick from each shelf (Appendix~\ref{app:relations}). Payload feasibility never binds in the official scenarios, since product mass lies between 0.5 and 2.0 while the payload is 10.0, and generated due dates enter only the urgency term of the optional weighted productive value, which is used by optional weighted diagnostics and by the weighted coalition selection of the privileged centralized reference; neither therefore affects AssemblyGrid v1 dynamics or benchmark measures.

\begin{table}[H]
\centering
\caption{Official recipe parameters.}
\label{tab:recipeparams}
\scriptsize
\setlength{\tabcolsep}{3.0pt}
\begin{tabularx}{\textwidth}{p{0.21\textwidth}p{0.15\textwidth}p{0.11\textwidth}p{0.13\textwidth}cYc}
\toprule
Recipe & Operation & Inputs $\to$ output & Predecessors & $[k_\tau^{\min},k_\tau^{\max}]$ & Roles & $d_\tau$ \\
\midrule
\texttt{solo\_inspection} & \texttt{inspect} & A $\to$ FINAL & none & $[1,1]$ & inspector & 2 \\
\addlinespace
\texttt{flow\_serial\_3} & \texttt{prepare} & A $\to$ P1 & none & $[1,1]$ & operator & 2 \\
 & \texttt{process} & P1 $\to$ P2 & \texttt{prepare} & $[1,1]$ & operator & 3 \\
 & \texttt{inspect} & P2 $\to$ FINAL & \texttt{process} & $[1,1]$ & inspector & 2 \\
\addlinespace
\texttt{flow\_serial\_5} & \texttt{prepare} & A $\to$ P1 & none & $[1,1]$ & operator & 2 \\
 & \texttt{process\_1} & P1 $\to$ P2 & \texttt{prepare} & $[1,1]$ & operator & 3 \\
 & \texttt{process\_2} & P2 $\to$ P3 & \texttt{process\_1} & $[1,1]$ & operator & 3 \\
 & \texttt{finish} & P3 $\to$ P4 & \texttt{process\_2} & $[1,1]$ & operator & 2 \\
 & \texttt{inspect} & P4 $\to$ FINAL & \texttt{finish} & $[1,1]$ & inspector & 2 \\
\addlinespace
\texttt{pair\_assembly} & \texttt{pair\_join} & A, B $\to$ FINAL & none & $[2,2]$ & holder\_A, holder\_B & 4 \\
\addlinespace
\texttt{supported\_insert} & \texttt{align\_ab} & A, B $\to$ AB & none & $[2,2]$ & holder\_A, holder\_B & 3 \\
 & \texttt{supported\_insert} & AB $\to$ FINAL & \texttt{align\_ab} & $[2,4]$ & holder, inserter, stabilizer, inspector & 4 \\
\addlinespace
\texttt{standard\_abc} & \texttt{align\_ab} & A, B $\to$ AB & none & $[2,2]$ & holder\_A, holder\_B & 3 \\
 & \texttt{stabilize\_insert} & AB, C $\to$ FINAL & \texttt{align\_ab} & $[4,4]$ & holder\_AB, holder\_C, operator, stabilizer & 5 \\
\addlinespace
\texttt{parallel\_branch\_light} & \texttt{prepare\_a} & A $\to$ PA & none & $[1,1]$ & operator\_a & 3 \\
 & \texttt{prepare\_b} & B $\to$ PB & none & $[1,1]$ & operator\_b & 3 \\
 & \texttt{join\_branches} & PA, PB $\to$ FINAL & \texttt{prepare\_a}, \texttt{prepare\_b} & $[2,2]$ & holder\_PA, holder\_PB & 4 \\
\bottomrule
\end{tabularx}
\end{table}

Table~\ref{tab:recipeparams} lists every operation of the seven official recipes, with the arity interval $[k_\tau^{\min},k_\tau^{\max}]$ and the nominal productive duration $d_\tau$ in ticks; recipe identifiers map to scenarios as listed in Appendix~\ref{app:reproducibility}. In AssemblyGrid v1 an operation takes place at a common target point computed from the base positions of its input holders, and each member's abstract target is offset from that point by the role-target offset. Role skills are also recorded in the recipe files but are inactive as skill semantics are disabled. Productive weights enter only the optional weighted productive value, which is used by optional weighted diagnostics and by the weighted coalition selection of the privileged centralized reference; decentralized agents never observe it, and it affects neither AssemblyGrid v1 dynamics nor benchmark measures.

\subsection{Rules and transition logic}
\label{app:rules}

The following rules define AssemblyGrid v1.
\begin{enumerate}[leftmargin=*]
\item \textbf{Release.} A product enters only when the declared release process and WIP or input-capacity constraints permit it; a scheduled release that is not admitted is lost and not queued.
\item \textbf{Recipe readiness.} An operation is logically ready only when it is neither completed nor already executing, all mandatory predecessor conditions hold, and it remains enabled under the product's resolved XOR choices. Material availability is checked separately. Completing one XOR alternative closes its mutually exclusive siblings as encoded by $\mathcal X_r$ for that product.
\item \textbf{Local coalition structure.} Direct-manipulation coalition members are pairwise adjacent in $\mathcal G_{\mathrm{manip}}$; handoff uses $\mathcal G_{\mathrm{handoff}}$; observation uses $\mathcal G_{\mathrm{obs}}$.
\item \textbf{Decentralized commitment.} A cooperative operation starts only after all required members independently support the same uniquely identified execution candidate and the joint admission check succeeds. The environment does not assign coalition partners. After admission, the participating robots become committed to the resulting activity.
\item \textbf{Bilateral handoff.} Transfer requires neighboring robots $u$ and $v$ to select matching sender and receiver actions, with $v\in\mathcal N(u)\setminus\{u\}$. Specifically,
\[
a_t^u=\mathrm{handoff}(v,q),\qquad a_t^v=\mathrm{receive}(u,q).
\]
The transfer point $p\in\mathcal H(u,v)$ must satisfy the operation geometry, local transfer feasibility, and the absence of a conflicting committed resource.
\item \textbf{Execution.} A committed activity occupies its actual robots, material inputs, exclusive resources, and abstract workspace or motion reservation for its declared duration.
\item \textbf{Completion.} Successful recipe-operation completion atomically applies the declared transition $\Delta_\tau$ and strictly extends the product-progress state. Rejected attempts do not fabricate progress.
\item \textbf{Delivery.} A product succeeds only when its terminal recipe output has been produced and successfully delivered through an admissible output interface.
\end{enumerate}

\subsection{Hard constraints}
\label{app:constraints}

Hard constraints govern both local candidate construction and later joint admission; they are not reward penalties. For an execution candidate $c$ associated with operation occurrence $(z,\tau)$ and proposed robot set $U_c$, the structural conditions are
\begin{align}
k_\tau^{\min}\le |U_c| &\le \min\!\left(k_\tau^{\max},k_{\mathrm{topo}}^{\max}\right),\\
\mathrm{ready}_t(z,\tau)&=1,\\
\mathrm{InputsPresent}(z,\tau,s_t)&=1,\\
x_u(c)&\in\mathcal W(u)\quad \forall u\in U_c.
\end{align}
Here $k_\tau^{\min}$ and $k_\tau^{\max}$ are the operation's coalition-size bounds, $\mathrm{ready}_t(z,\tau)$ is logical recipe readiness, $\mathrm{InputsPresent}(z,\tau,s_t)$ requires the declared material inputs, and $x_u(c)$ is the role-specific abstract target for robot $u$. For variable-arity operations with explicit roles, the ordered AssemblyGrid v1 role declaration covers $k_\tau^{\max}$ members, while the canonical assignment rule instantiates exactly $|U_c|$ roles for the realized candidate. Together with the declared role and manipulation-locality requirements, these conditions determine whether a candidate can be exposed to an eligible robot when they are available from that robot's permitted local information.

A fully supported candidate is admitted only if the joint transition additionally satisfies
\begin{equation}
\mathrm{ResourceOK}(c,s_t)=1,
\qquad
\mathrm{AbstractMotionOK}(c,s_t)=1,
\end{equation}
where the first predicate resolves finite and exclusive auxiliary resources and the second resolves the active task-level geometry, including conflicts with already committed or simultaneously considered activities. A robot, material item, fixture, buffer slot, or other exclusive resource cannot be simultaneously committed to incompatible activities, and material conservation and declared capacity limits must hold at every transition. Thus hidden nonlocal resource or concurrency conflicts are checked at joint admission. If $c$ is admitted, the resulting activity $e(c)$ inherits the product unit, recipe operation, robot set, roles, operation site, auxiliary resources, and geometric assignment specified by the candidate. Table~\ref{tab:bindings} summarizes the principal bindings these constraints create.

\paragraph{Feasibility.}
A generated benchmark instance is feasible only if a policy-independent successful execution exists: raw material can reach every operation that consumes it, a feasible handoff chain connects required sources to an output interface, every required operation satisfies $k_\tau^{\min}\le k_{\mathrm{topo}}^{\max}$ and admits at least one admissible execution candidate under the active geometry profile, and declared resource capacities do not make a required operation permanently infeasible. An instance that violates these conditions is invalid.

\subsection{Bindings and dependencies}
\label{app:bindings}

\begin{table}[H]
\centering
\caption{AssemblyGrid v1 bindings and dependencies.}
\label{tab:bindings}
\small
\begin{tabularx}{\textwidth}{p{0.24\textwidth}p{0.30\textwidth}Y}
\toprule
Binding & Formal or operational form & Consequence \\
\midrule
Product to recipe & $z\mapsto\Gamma_{r(z)}$ & Product achievement and legal future operations are recipe-specific. \\
Operation to material & $\tau\leftrightarrow(\mathrm{inputs}_\tau,\Delta_\tau)$ & Logical readiness alone is insufficient unless the required material inputs are present; successful completion applies the operation's declared atomic material transition. \\
Operation to coalition & $\tau\leftrightarrow[k_\tau^{\min},k_\tau^{\max}]$ & Required simultaneous participation is operation-specific and limited by the topology. \\
Coalition to role & candidate member-to-role assignment & A candidate is admissible only if its proposed members can fill the roles declared by the operation. \\
Robot to capability & $u\leftrightarrow$ declared task-level capabilities & Interchangeability is conditional under the active geometry profile. \\
Robot to workspace & $u\leftrightarrow\mathcal W(u)$ & Base location and geometry profile determine abstract reach and direct transfer or manipulation opportunities. \\
Operation to resource & $\tau\leftrightarrow\mathrm{res}_\tau$ & Fixtures, buffers, tools, or shared zones can couple otherwise independent products. \\
Coalition to time & $C\leftrightarrow[t,t+D)$ & Committed robots are unavailable to incompatible concurrent work. \\
\bottomrule
\end{tabularx}
\end{table}

\subsection{Structural relations and interaction graphs}
\label{app:relations}

Several graphs coexist and are not interchangeable:
\begin{itemize}[leftmargin=*]
\item $\mathcal G_{\mathrm{obs}}$: which neighbors' task-level dynamic state is locally observable;
\item $\mathcal G_{\mathrm{handoff}}$: which pairs may perform direct bilateral material transfer;
\item $\mathcal G_{\mathrm{manip}}$: which pairs may jointly participate in one direct-manipulation coalition;
\item $\Gamma_r=(\mathcal T_r,\mathcal E_r,\mathcal X_r)$: the recipe structure for product type $r$, with mandatory predecessor relations in $\mathcal E_r$ and explicit XOR alternatives in $\mathcal X_r$;
\item $G_t^{\mathrm{op}}$: the state-dependent conflict graph over concrete ongoing and prospective activities.
\end{itemize}
The three robot relations coincide geometrically in the default Moore-radius-one configuration but remain semantically distinct. The maximum coalition size is derived specifically from $\mathcal G_{\mathrm{manip}}$ through $k_{\mathrm{topo}}^{\max}=\omega(\mathcal G_{\mathrm{manip}})$.

\paragraph{Grid and boundary sets.}
For an $M\times N$ grid, $\mathcal U=\{u_{i,j}:1\le i\le M,\ 1\le j\le N\}$ with $|\mathcal U|=MN$. The closed Moore neighborhood is
\begin{equation}
\mathcal N(u_{i,j})=\{u_{i',j'}\in\mathcal U:\max(|i'-i|,|j'-j|)\le1\},
\end{equation}
and it induces $\mathcal G_{\mathrm{obs}}$, $\mathcal G_{\mathrm{handoff}}$, and $\mathcal G_{\mathrm{manip}}$ in the default configuration. Input shelves $\mathcal S_{\mathrm{in}}$ with pick regions $c_s$ feed the left column, and the boundary robot sets are
\begin{equation}
\begin{gathered}
\mathcal U_{\mathrm{in}}=\{u_{i,1}\}_{i=1}^{M},\qquad
\mathcal U_{\mathrm{out}}=\{u_{i,N}\}_{i=1}^{M},\\
\mathcal U(s)=\{u\in\mathcal U_{\mathrm{in}}:c_s\cap\mathcal W(u)\neq\varnothing,\ \mathrm{row}(u)=\mathrm{row}(s)\lor\beta_{u,s}=1\},
\end{gathered}
\end{equation}
where $\mathcal U_{\mathrm{in}}$ and $\mathcal U_{\mathrm{out}}$ are the left-boundary input and right-boundary output robot sets, respectively, and $\mathcal U(s)$ is the subset of input robots with pick access to $c_s$. An input robot in the shelf's own row has access whenever its workspace reaches $c_s$; any other input robot whose workspace reaches $c_s$ has access when the instance-generation indicator $\beta_{u,s}\sim\mathrm{Bernoulli}(0.30)$, drawn once from the generation seed, equals one. Internal routing need not be monotone. Each robot $u$ has an abstract reachable workspace $\mathcal W(u)=\{x:\lVert x-b(u)\rVert_2\le R_u\}$ about its base $b(u)$ with declared reach radius $R_u$, and a bilateral handoff between neighbors $u,v$ is possible only in the shared region $\mathcal H(u,v)=\mathcal W(u)\cap\mathcal W(v)$.

\subsection{State-dependent feasibility}
\label{app:feasibility}

Let $c$ denote an execution candidate and let $s_t$ denote the state at decision epoch $t$. The candidate-level structural predicate is
\begin{equation}
\begin{aligned}
\mathrm{CandidateBase}(c,s_t)={}&
\mathrm{Ready}(c,s_t)\land
\mathrm{Material}(c,s_t)\land
\mathrm{Arity}(c)\land
\mathrm{Local}(c)\\
&\land\mathrm{Capability}(c)\land
\mathrm{Reach}(c).
\end{aligned}
\end{equation}
The predicates require, respectively, a logically ready operation, present material inputs, a coalition size within recipe and topology bounds, a valid manipulation-local robot set, role-compatible robots, and role-specific targets within the corresponding abstract workspaces. Robot $u$ can be exposed to candidate $c$ only if $u\in U_c$ and these candidate-level conditions are satisfied using information permitted by the local interface.

For a fully supported candidate, joint admission additionally requires
\begin{equation}
\mathrm{JointAdmissible}(c,s_t)=
\mathrm{CandidateBase}(c,s_t)
\land\mathrm{Resource}(c,s_t)
\land\mathrm{AbstractMotion}(c,s_t),
\end{equation}
where $\mathrm{Resource}(c,s_t)$ resolves finite and exclusive resources and $\mathrm{AbstractMotion}(c,s_t)$ resolves the active approach, retreat, clearance, and concurrency-conflict conditions against activities already executing and other candidates considered for the same interval. These nonlocal conditions are evaluated by the joint transition and are not used to leak hidden global state through the decentralized candidate list or action mask.

\paragraph{Joint feasibility and concurrency capacity.} At decision epoch $t$, let $\mathcal E^{\mathrm{run}}_t$ be the committed recipe-operation activities already executing and let $\mathcal C^{\mathrm{new}}_t$ be the individually admissible execution candidates available in the current state, independent of whether they are fully supported by the controller. Running nonproductive logistics reservations remain fixed feasibility constraints through the resource and workspace checks but are not counted as capacity activities. For capacity analysis, each $c\in\mathcal C^{\mathrm{new}}_t$ is represented by the prospective activity it would create if admitted; let $\mathcal E_t$ contain those prospective activities together with $\mathcal E^{\mathrm{run}}_t$. Each $e\in\mathcal E_t$ therefore fixes a product unit, recipe operation, coalition, roles, operation site, and abstract-motion assignment. Actual joint admission is evaluated separately and is restricted to the fully supported subset $\mathcal C^{\mathrm{sup}}_t\subseteq\mathcal C^{\mathrm{new}}_t$. The conflict graph used for capacity analysis is
\begin{equation}
G_t^{\mathrm{op}}=(\mathcal E_t,E_t^{\mathrm{conf}}),\qquad
(e,e')\in E_t^{\mathrm{conf}}\ \Leftrightarrow\ e,e'\ \text{cannot share the execution interval,}
\end{equation}
where $E_t^{\mathrm{conf}}$ is the set of pairwise execution-conflict edges. The feasible concurrent activity sets that preserve existing commitments are
\begin{equation}
\mathfrak I_t=\Bigl\{I\subseteq\mathcal E_t:\ I\ \text{independent in}\ G_t^{\mathrm{op}},\ \mathcal E_t^{\mathrm{run}}\subseteq I,\ \bigl|\{e\in I:\rho\in\mathrm{res}(e)\}\bigr|+n_\rho^{\mathrm{ext}}(t)\le\mathrm{cap}(\rho)\ \ \forall\rho\Bigr\}.
\end{equation}
 In canonical AssemblyGrid v1, every activity in $\mathcal E_t$ is a recipe-operation activity whose successful completion advances product progress, thus $\mathcal E_t^{\mathrm{prod}}=\mathcal E_t$. The feasibility calculation preserves all committed operation teams and enforces conflicts from nonproductive logistics reservations through the resource and workspace checks used to construct admissible prospective activities; those logistics reservations constrain capacity but are not counted as capacity activities. The resource condition is stated separately from the conflict graph because a pairwise relation cannot express it: three activities sharing a resource of capacity two are pairwise compatible yet cannot all execute. Here $\mathrm{res}(e)$ are the exclusive resources of activity $e$, $\mathrm{cap}(\rho)$ is the declared capacity of resource $\rho$, and $n_\rho^{\mathrm{ext}}(t)$ counts only committed users of $\rho$ outside $I$, such as ongoing logistics reservations, thus committed productive activities are counted once through their membership in $I$. The single canonical count capacity is therefore $K_{\mathrm{prod}}^*(s_t)=\max_{I\in\mathfrak I_t}|I|$. Alternative realizations of the same product-operation occurrence are mutually exclusive in the conflict/selection construction, and nonproductive logistics reservations constrain this maximum without contributing to it. In the reference implementation, the group-aware search is exact when the candidate set contains at most 64 candidates, at most 6 product-operation groups, and at most 100{,}000 grouped include-or-skip combinations. States outside these limits use deterministic greedy selection as an explicitly flagged lower bound; exactness is recorded for each sample. Canonical utilization averages are computed only on exact-capacity opportunity states, while approximate-state quantities are reported separately as bounds. For cooperative-only analysis the counted activities may be restricted to operations with $k_\tau^{\min}\ge2$. The product-progress state $p_z(t)$ is the triple of completed operations, resolved XOR choices, and delivery status; it is monotone in AssemblyGrid v1 (no rework or scrap), and an activity is productive iff its completion strictly extends $p_z(t)$ for some product.

\subsection{Assumptions and abstractions}
\label{app:assumptions}

AssemblyGrid v1 uses synchronized discrete time; fixed robot bases; perfect local task-level sensing under $\mathcal G_{\mathrm{obs}}$; aggregate local candidate-support fraction with no free-form or addressable messages; discrete material items with unique ownership; abstracted part pose, orientation, grasp, and contact; deterministic operation durations together with the default abstract geometric approach and retreat model; monotone product progress with no rework or scrap; a homogeneous robot population under the default \texttt{abstract-v1} profile; and task-level cooperation with no adversarial agent-specific objectives. The geometry model uses nominal task-level reach, clearance, timing, and abstract motion segments.

Kinematic or task-and-motion planning, physics and perception, disturbances, heterogeneous robot-capability studies, skills, due-date or value semantics, and standardized out-of-distribution generalization splits over recipes, layouts, geometry profiles, or scale are explicit extensions. If used, they must be named and reported as extensions rather than silently changing AssemblyGrid v1.

\subsection{Structural instance descriptors and experimental factors}
\label{app:factors}

The following descriptors summarize structural properties of an instance or a matched intervention. They are not official workload-family labels and do not define a universal scalar difficulty:
\begin{align}
D_{\mathrm{size}} &=(MN,\mathrm{WIP}_{\max},|\mathcal T_r|,|\mathcal R|),\\
D_{\mathrm{coalition}} &=(f_{\mathrm{coal}},\mathcal K_r),\\
D_{\mathrm{geometry}} &=(r_{\mathrm{space}},\rho_{\mathrm{conf}},\nu_{\mathrm{geom}}),\\
D_{\mathrm{process}} &=(L_r,w_r,n_{\mathrm{join}},n_{\mathrm{xor}}),\\
D_{\mathrm{topology}} &=(\mathcal G_{\mathrm{obs}},\mathcal G_{\mathrm{handoff}},\mathcal G_{\mathrm{manip}}),
\end{align}
where $MN$ is the robot count, $\mathrm{WIP}_{\max}$ is the configured WIP cap, $|\mathcal T_r|$ is the number of operations in recipe $r$, and $|\mathcal R|$ is the number of active product types. The coalition fraction is $f_{\mathrm{coal}}=|\{\tau\in\mathcal T_r:k_\tau^{\min}>1\}|/|\mathcal T_r|$, and $\mathcal K_r$ is the multiset of minimum coalition requirements $k_\tau^{\min}$ in recipe $r$. The spacing ratio is $r_{\mathrm{space}}=d_{\mathrm{base}}/R_{\mathrm{ref}}$, where $d_{\mathrm{base}}$ is nominal center-to-center robot-base spacing and $R_{\mathrm{ref}}$ is reference reach. The quantity $\rho_{\mathrm{conf}}$ is the declared normalized workspace-conflict radius, and $\nu_{\mathrm{geom}}$ is the active geometry-profile or variant identifier. Finally, $L_r$ is recipe DAG depth, $w_r$ is maximum antichain width, and $n_{\mathrm{join}}$ and $n_{\mathrm{xor}}$ count join and XOR structures respectively.

Topology and required coalition size are coupled, since every required operation must satisfy $k_\tau^{\min}\le k_{\mathrm{topo}}^{\max}$; consequently, scenario design cannot vary coalition demand independently of the active manipulation topology. Communication noise, disturbances, skill learning, and held-out transfer splits remain extension factors.

\subsection{Performance and diagnostic metrics}
\label{app:metric-register}

Metrics do not alter environment dynamics and are independent of any training reward. Throughput, admitted-product completion, release acceptance, offered-demand fulfillment, lost releases, WIP, restricted mean flow time, admission and coalition success, occupancy, blocked-ready rates, blocked-activity events with their cause-specific subsets, and the operational no-progress-tick diagnostic are defined in Section~\ref{sec:evaluation}; statewise productive-concurrency capacity, realized productive concurrency, and utilization are defined in Section~\ref{sec:productive}. Execution-averaged capacity is a visited-state diagnostic and can differ across controllers because their trajectories visit different states.

For a completed finite batch $\mathcal Z$, the makespan is $C_{\max}=\max_{z\in\mathcal Z}C_z$; if the declared finite-batch timeout $T_B$ is reached first, the execution is reported as timed out and no $C_{\max}$ is assigned. Candidate switching counts robot switches from pending candidate support, and geometry reporting additionally uses normalized approach or retreat delay ticks. The legacy implementation key for the operational no-progress-tick diagnostic is \texttt{deadlock\_ticks}.
Further optional diagnostics include handoffs per product and role imbalance; abstract activity duration, path length, and reach or clearance margin are optional geometry diagnostics. Joint-space path length, inverse-kinematic failure, link collision, dynamics, contact, and grasp metrics require a declared higher-detail extension. Wall-clock speed, memory use, and diagnostic-solver cost are implementation instrumentation.

\paragraph{Reference bounds.}
Conservative algorithm-independent quantities give validation checks and small-instance references. If $n_k(t)$ operations requiring exactly $k$ robots run concurrently, robot exclusivity gives $\sum_{k=1}^{4}k\,n_k(t)\le MN$, hence $n_k(t)\le\lfloor MN/k\rfloor$ for a batch of operations with one coalition size before locality, reachability, precedence, and motion constraints. For product $z$ released at $a_z$,
\begin{equation}
C_z-a_z\ge\max\{L_{\mathrm{crit}}(z),L_{\mathrm{transport}}(z),L_{\mathrm{motion}}(z),L_{\mathrm{resource}}(z)\},
\end{equation}
where $L_{\mathrm{crit}}$ is a critical-path bound, $L_{\mathrm{transport}}$ is a required-transfer bound, $L_{\mathrm{motion}}$ a necessary-abstract-motion bound, and $L_{\mathrm{resource}}$ an exclusive-resource bound; the terms need not be independent. The robot-time workload below sums the declared operations of a recipe and is therefore a valid lower bound for recipes whose operations all execute, which covers every official recipe; for a recipe with exclusive alternative routes the sum must be restricted to the operations of the realized route. For a finite batch $\mathcal Z$, let $D_{\min}(z,\tau)$ denote the minimum admissible realized duration of operation $\tau$ for product $z$ under the instance. The lower-bound robot-time workload is $W_R^{\mathrm{LB}}=\sum_{z\in\mathcal Z}\sum_{\tau\in\mathcal T_{r(z)}}k_\tau^{\min}D_{\min}(z,\tau)$,
\begin{equation}
C_{\max}\ge\max\{\textstyle\max_z L_{\mathrm{crit}}(z),\ W_R^{\mathrm{LB}}/MN,\ L_{\mathrm{resource,batch}}\}.
\end{equation}
where $L_{\mathrm{resource,batch}}$ is the batch-level lower bound induced by exclusive-resource workloads. These bounds hold across changing coalition sizes and do not assume independent $2\times2$ machines.

\subsection{Formal invariants}
\label{app:invariants}

A conforming simulator must preserve the following properties. These are semantic consistency conditions not performance objectives, and each is referred to by its name in the executable mapping of Appendix~\ref{app:reproducibility}.
\begin{description}[leftmargin=*,itemsep=1pt,font=\normalfont\bfseries]
\item[Conservation.] Every material item has a single consistent location and ownership state; no transition creates a duplicate material item except through an explicitly declared recipe output transition.
\item[Support.] Cooperative support for a held item does not create additional material ownership or duplicate the material item.
\item[Progress.] AssemblyGrid v1 completion never removes an already completed recipe operation from a product's progress state.
\item[Recipe.] Completed and executing operation sets are disjoint, predecessor and XOR constraints are respected, and delivery occurs only after terminal recipe completion.
\item[Locality.] Every pair of members of a committed direct-manipulation coalition is connected in $\mathcal G_{\mathrm{manip}}$.
\item[Lifetime.] Coalition membership, roles, and operation identity remain fixed while the activity is committed or executing and change only at a terminal event.
\item[Arity.] Realized coalition size lies within the operation's declared interval and does not exceed $k_{\mathrm{topo}}^{\max}$; assigned roles cover exactly the participating members.
\item[Exclusivity.] A robot or other exclusive resource cannot be committed to incompatible overlapping activities, and finite-capacity resources cannot exceed their declared capacity.
\item[Observation.] A decentralized observation is a function only of the robot's own state, declared local dynamic state, and explicitly public benchmark information.
\item[Masking.] Local action masking removes declared local impossibilities but does not expose a global joint-feasibility oracle; at least one legal action remains available under every reachable decision state.
\item[Accounting.] Cumulative released and delivered counts never decrease, and reported WIP remains consistent with released but undelivered products.
\end{description}

Appendix~\ref{app:reproducibility} provides the executable mapping for these invariants and links each declared consistency condition to the corresponding implementation-level check. Exact diagnostic solvers must additionally agree with independently derived reference optima, obtained by hand or by exhaustive search, on the small validation instances for which exactness is claimed, providing a direct consistency check between the optimized diagnostic procedures and their reference calculations.

\section{Benchmark scope and extensions}
\label{app:capability}

Table~\ref{tab:capabilityregister} consolidates the scope of AssemblyGrid v1 by distinguishing mechanisms represented by the benchmark from properties introduced only through declared extensions.

\subsection{Reading the scope summary}

Each row names one mechanism, interface, or evaluation control and answers three independent questions about the named property: \emph{Modeled} asks whether AssemblyGrid v1 represents the property at all, that is, whether it appears in the state, the transition model, the information structure, or the evaluation protocol. \emph{Varied} asks whether the official suite or a declared matched control changes the property across evaluated instances, which is what makes it an experimental variable. \emph{Measured} asks whether the benchmark reports a metric for the property, thereby making its effect visible in results.

The three questions are kept separate, since they represent distinct properties of the benchmark. Action masking is modeled and constrains every step, yet it is held fixed across the official suite and is not itself an outcome measure. Temporary coalition formation is modeled, varied across the three \textsc{Coalition} scenarios, and reported through formation success and failure counts. The privileged global state is modeled and used by centralized-training methods, but it is never a reported outcome, since the global state serves as a training-time interface not a production result. Reading the three columns together therefore shows not only what the benchmark contains, but which parts of it are currently exercised as experimental variables and which are observable in the reported measures.

A filled mark records that the property holds directly, and an open mark records that it holds partially, at a coarser abstraction, or only inside a matched control. A dash records that the property is part of AssemblyGrid v1 but is not varied or measured in that column; ``n/a'' is reserved for properties outside AssemblyGrid v1.

\begin{table}[htbp]
\centering
\caption{AssemblyGrid v1 scope and extensions.}
\label{tab:capabilityregister}
\scriptsize
\setlength{\tabcolsep}{4pt}
\renewcommand{\arraystretch}{1.05}
\begin{tabular}{@{}>{\raggedright\arraybackslash}p{\dimexpr0.525\textwidth-32pt\relax}>{\centering\arraybackslash}p{0.08\textwidth}>{\centering\arraybackslash}p{0.08\textwidth}>{\centering\arraybackslash}p{0.08\textwidth}>{\raggedright\arraybackslash}p{0.235\textwidth}@{}}
\toprule
Aspect & Modeled & Varied & Measured & Specified in \\
\midrule
\multicolumn{5}{@{}l}{\itshape Product and material progression}\\[1pt]
Recipe graph with multi-stage transformation & $\checkmark$ & $\checkmark$ & $\checkmark$ & Sec.~\ref{sec:recipes}, App.~\ref{app:rules} \\
Sequence, AND, and XOR precedence & $\checkmark$ & $\circ$ & $\circ$ & Sec.~\ref{sec:recipes} \\
Material conservation and unique ownership & $\checkmark$ & -- & $\checkmark$ & App.~\ref{app:constraints}, App.~\ref{app:invariants} \\
Bilateral synchronized handoff & $\checkmark$ & $\checkmark$ & $\checkmark$ & Sec.~\ref{sec:teamformation}, App.~\ref{app:rules} \\
Terminal delivery and task success & $\checkmark$ & $\checkmark$ & $\checkmark$ & Sec.~\ref{sec:tasksuccess} \\
Rework, rollback, or scrap semantics & n/a & n/a & n/a & App.~\ref{app:assumptions} \\
\addlinespace
\multicolumn{5}{@{}l}{\itshape Temporary cooperation}\\[1pt]
Single-robot operations or temporary coalitions of two to four robots & $\checkmark$ & $\checkmark$ & $\checkmark$ & Sec.~\ref{sec:teamformation}, Sec.~\ref{sec:difficultylevels} \\
Declared manipulation roles per member & $\checkmark$ & $\circ$ & $\circ$ & Sec.~\ref{sec:recipes} \\
Membership fixed while an operation is committed & $\checkmark$ & -- & $\checkmark$ & Sec.~\ref{sec:teamformation}, App.~\ref{app:invariants} \\
Coalition formation success, failure, and latency & $\checkmark$ & $\checkmark$ & $\checkmark$ & Sec.~\ref{sec:mechanismevidence} \\
Aggregate support-fraction coordination cue & $\checkmark$ & $\circ$ & $\circ$ & Sec.~\ref{sec:obscontract} \\
Persistent teams or standing alliances & n/a & n/a & n/a & Sec.~\ref{sec:teamformation} \\
Free-form or addressable communication & n/a & n/a & n/a & Sec.~\ref{sec:scope} \\
\addlinespace
\multicolumn{5}{@{}l}{\itshape Decentralized information}\\[1pt]
Bounded local observation on the Moore neighborhood & $\checkmark$ & -- & $\circ$ & Sec.~\ref{sec:obscontract} \\
Action masking of hard impossibilities detectable from permitted local information & $\checkmark$ & -- & $\circ$ & Sec.~\ref{sec:obscontract} \\
Privileged global-state access (training for MAPPO and QMIX; evaluation-time action selection for Centralized) & $\checkmark$ & $\circ$ & -- & Sec.~\ref{sec:obscontract}; Sec.~\ref{sec:evalresults} \\
Privileged global joint-feasibility oracle exposed to the controller & n/a & n/a & n/a & Sec.~\ref{sec:obscontract} \\
Sensing noise or perception layer & n/a & n/a & n/a & App.~\ref{app:assumptions} \\
\addlinespace
\multicolumn{5}{@{}l}{\itshape Geometry and motion}\\[1pt]
Abstract reach and workspace overlap & $\checkmark$ & $\checkmark$ & $\checkmark$ & Sec.~\ref{sec:motionfeas}, App.~\ref{app:relations} \\
Clearance and workspace-conflict predicates & $\checkmark$ & $\circ$ & $\checkmark$ & Sec.~\ref{sec:motionfeas}, Sec.~\ref{sec:mechanismevidence} \\
Approach and retreat motion delay & $\checkmark$ & $\circ$ & $\checkmark$ & Sec.~\ref{sec:motionexecution} \\
Kinematics, link collision, contact, and grasp dynamics & n/a & n/a & n/a & Sec.~\ref{sec:scope} \\
\addlinespace
\multicolumn{5}{@{}l}{\itshape Concurrency and resources}\\[1pt]
Productive concurrency distinguished from occupancy & $\checkmark$ & $\checkmark$ & $\checkmark$ & Sec.~\ref{sec:productive} \\
Productive-concurrency capacity & $\checkmark$ & $\circ$ & $\checkmark$ & Sec.~\ref{sec:productive}, App.~\ref{app:feasibility} \\
Exclusive resources, fixtures, and buffers & $\checkmark$ & $\checkmark$ & $\circ$ & App.~\ref{app:constraints}, App.~\ref{app:bindings} \\
WIP cap and product release process & $\checkmark$ & $\checkmark$ & $\circ$ & Sec.~\ref{sec:difficultylevels} \\
Offered, admitted, and lost releases / demand fulfillment (derivable, not exported) & $\checkmark$ & $\checkmark$ & $\circ$ & Sec.~\ref{sec:evaluation} \\
\addlinespace
\multicolumn{5}{@{}l}{\itshape Robots and topology}\\[1pt]
Fixed-base homogeneous manipulator lattice & $\checkmark$ & $\circ$ & -- & Sec.~\ref{sec:environment} \\
Grid size & $\checkmark$ & $\checkmark$ & -- & Sec.~\ref{sec:difficultylevels} \\
Moore-radius-one interaction topology & $\checkmark$ & -- & -- & Sec.~\ref{sec:environment}, App.~\ref{app:relations} \\
Heterogeneous robot capability & n/a & n/a & n/a & App.~\ref{app:assumptions} \\
Mobile robots or conveyor transport & n/a & n/a & n/a & Sec.~\ref{sec:scope} \\
Robot downtime and failure recovery & n/a & n/a & n/a & App.~\ref{app:assumptions} \\
\addlinespace
\multicolumn{5}{@{}l}{\itshape Evaluation protocol}\\[1pt]
Task success scored independently of learning reward & $\checkmark$ & -- & $\checkmark$ & Sec.~\ref{sec:obscontract}, Sec.~\ref{sec:tasksuccess} \\
Instance identity, seed separation, and versioning & $\checkmark$ & $\checkmark$ & $\checkmark$ & Sec.~\ref{sec:instances}, App.~\ref{app:reproducibility} \\
Fixed validation and final-test instance splits & $\checkmark$ & $\checkmark$ & $\checkmark$ & Sec.~\ref{sec:evalresults} \\
Held-out generalization split over recipes or scale & n/a & n/a & n/a & App.~\ref{app:assumptions} \\
Standardized robustness or disturbance protocol & n/a & n/a & n/a & Sec.~\ref{sec:scope} \\
\bottomrule
\multicolumn{5}{@{}p{\textwidth}@{}}{\vspace{0.4mm}$\checkmark$ property holds directly; $\circ$ holds partially, at a coarser abstraction, or only in a matched control; -- means not varied or measured in that column; n/a means outside AssemblyGrid v1 and reserved for a named extension.}
\end{tabular}
\end{table}

\subsection{Scope entries and extensions}

A row carrying ``n/a'' in all three columns identifies a property requiring a named extension with its own specification and evaluation protocol. Incorporating such a property into the canonical task definition constitutes a benchmark-version change, whereas varying an already modeled property extends the experimental design without changing the underlying task.

Two limits of this summary should be kept in mind. The summary records the state of the released benchmark; consequently, a controller that ignores a modeled mechanism is still evaluated against that mechanism. The summary also describes AssemblyGrid v1 only and does not compare it with other benchmarks; Table~\ref{tab:litfamilies} provides that comparison.

\end{document}